\documentclass[x11names]{article}
\usepackage[a4paper, total={7.2in, 10.5in}]{geometry}

\usepackage{amsmath}
\usepackage{amsfonts}
\usepackage{amssymb}
\usepackage{siunitx}
\usepackage{bbding}
\usepackage{bm}
\usepackage[export]{adjustbox}
\usepackage{enumitem}
\usepackage{mathtools}
\usepackage{booktabs}

\usepackage[dvipsnames, table]{xcolor}
\usepackage{subcaption}
\usepackage{floatrow}
\usepackage[framemethod=tikz]{mdframed}

\usepackage{array}
\usepackage{calc} 
\usepackage{makecell} 
\usepackage{highlight}
\usepackage{colortbl}

\usepackage{pgfplots}
\pgfplotsset{compat=1.18}

\usepackage{nomencl}
\makenomenclature
\usepackage{framed}
\usepackage{multicol}
\renewcommand{\nompreamble}{\begin{multicols}{3}}
\renewcommand{\nompostamble}{\end{multicols}}

\renewcommand*\nompreamble{\begin{multicols}{3}}
\renewcommand*\nompostamble{\end{multicols}}

\usepackage[numbers, compress]{natbib}
\usepackage[colorlinks = true, pdfstartview = FitV, linkcolor = blue, citecolor = blue, urlcolor = blue, pdfencoding=auto, psdextra]{hyperref}
\usepackage{cleveref}

\usepackage{caption}
\usepackage{soul}
\soulregister\cite7
\soulregister\Cref7
\soulregister\texttt7
\sethlcolor{ForestGreen!20}

\crefname{section}{Section}{Sections}
\crefname{appendix}{Appendix}{Appendices}

\newcommand*\justify{%
  \fontdimen2\font=0.4em
  \fontdimen3\font=0.2em
  \fontdimen4\font=0.1em
  \fontdimen7\font=0.1em
  \hyphenchar\font=`\-
}

\renewcommand{\texttt}[1]{%
  \begingroup
  \ttfamily
  \begingroup\lccode`~=`/\lowercase{\endgroup\def~}{/\discretionary{}{}{}}%
  \begingroup\lccode`~=`[\lowercase{\endgroup\def~}{[\discretionary{}{}{}}%
  \begingroup\lccode`~=`.\lowercase{\endgroup\def~}{.\discretionary{}{}{}}%
  \catcode`/=\active\catcode`[=\active\catcode`.=\active
  \justify\scantokens{#1\noexpand}%
  \endgroup
}

\usepackage{tikz}
\usetikzlibrary{matrix,calc,arrows.meta,shapes,arrows,positioning,shadows,trees,mindmap,backgrounds,fit}

\definecolor{tikz_Blue}{HTML}{1f77b4}
\definecolor{tikz_Red}{HTML}{d62728}
\definecolor{tikz_Green}{HTML}{2ca02c}
\definecolor{tikz_Orange}{HTML}{ff7f0e}

\definecolor{v1}{HTML}{fde725}
\definecolor{v2}{HTML}{f8e621}
\definecolor{v3}{HTML}{f1e51d}
\definecolor{v4}{HTML}{ece51b}
\definecolor{v5}{HTML}{e5e419}
\definecolor{v6}{HTML}{dfe318}
\definecolor{v7}{HTML}{d8e219}
\definecolor{v8}{HTML}{d0e11c}
\definecolor{v9}{HTML}{cae11f}
\definecolor{v10}{HTML}{c2df23}
\definecolor{v11}{HTML}{bddf26}
\definecolor{v12}{HTML}{b5de2b}
\definecolor{v13}{HTML}{addc30}
\definecolor{v14}{HTML}{a8db34}
\definecolor{v15}{HTML}{a0da39}
\definecolor{v16}{HTML}{9bd93c}
\definecolor{v17}{HTML}{93d741}
\definecolor{v18}{HTML}{8ed645}
\definecolor{v19}{HTML}{86d549}
\definecolor{v20}{HTML}{7fd34e}
\definecolor{v21}{HTML}{7ad151}
\definecolor{v22}{HTML}{73d056}
\definecolor{v23}{HTML}{6ece58}
\definecolor{v24}{HTML}{67cc5c}
\definecolor{v25}{HTML}{60ca60}
\definecolor{v26}{HTML}{5cc863}
\definecolor{v27}{HTML}{56c667}
\definecolor{v28}{HTML}{52c569}
\definecolor{v29}{HTML}{4cc26c}
\definecolor{v30}{HTML}{48c16e}
\definecolor{v31}{HTML}{42be71}
\definecolor{v32}{HTML}{3dbc74}
\definecolor{v33}{HTML}{3aba76}
\definecolor{v34}{HTML}{35b779}
\definecolor{v35}{HTML}{32b67a}
\definecolor{v36}{HTML}{2eb37c}
\definecolor{v37}{HTML}{2ab07f}
\definecolor{v38}{HTML}{28ae80}
\definecolor{v39}{HTML}{25ac82}
\definecolor{v40}{HTML}{24aa83}
\definecolor{v41}{HTML}{22a785}
\definecolor{v42}{HTML}{20a486}
\definecolor{v43}{HTML}{1fa287}
\definecolor{v44}{HTML}{1fa088}
\definecolor{v45}{HTML}{1f9e89}
\definecolor{v46}{HTML}{1e9b8a}
\definecolor{v47}{HTML}{1f998a}
\definecolor{v48}{HTML}{1f968b}
\definecolor{v49}{HTML}{20938c}
\definecolor{v50}{HTML}{20928c}
\definecolor{v51}{HTML}{218f8d}
\definecolor{v52}{HTML}{228d8d}
\definecolor{v53}{HTML}{238a8d}
\definecolor{v54}{HTML}{24878e}
\definecolor{v55}{HTML}{25858e}
\definecolor{v56}{HTML}{26828e}
\definecolor{v57}{HTML}{26818e}
\definecolor{v58}{HTML}{277e8e}
\definecolor{v59}{HTML}{287c8e}
\definecolor{v60}{HTML}{29798e}
\definecolor{v61}{HTML}{2a768e}
\definecolor{v62}{HTML}{2b748e}
\definecolor{v63}{HTML}{2c718e}
\definecolor{v64}{HTML}{2d708e}
\definecolor{v65}{HTML}{2e6d8e}
\definecolor{v66}{HTML}{306a8e}
\definecolor{v67}{HTML}{31688e}
\definecolor{v68}{HTML}{32658e}
\definecolor{v69}{HTML}{33638d}
\definecolor{v70}{HTML}{34608d}
\definecolor{v71}{HTML}{365d8d}
\definecolor{v72}{HTML}{375b8d}
\definecolor{v73}{HTML}{38588c}
\definecolor{v74}{HTML}{39558c}
\definecolor{v75}{HTML}{3b528b}
\definecolor{v76}{HTML}{3c508b}
\definecolor{v77}{HTML}{3d4d8a}
\definecolor{v78}{HTML}{3e4989}
\definecolor{v79}{HTML}{3f4788}
\definecolor{v80}{HTML}{414487}
\definecolor{v81}{HTML}{424186}
\definecolor{v82}{HTML}{433e85}
\definecolor{v83}{HTML}{443a83}
\definecolor{v84}{HTML}{453882}
\definecolor{v85}{HTML}{463480}
\definecolor{v86}{HTML}{46327e}
\definecolor{v87}{HTML}{472e7c}
\definecolor{v88}{HTML}{472c7a}
\definecolor{v89}{HTML}{482878}
\definecolor{v90}{HTML}{482475}
\definecolor{v91}{HTML}{482173}
\definecolor{v92}{HTML}{481d6f}
\definecolor{v93}{HTML}{481b6d}
\definecolor{v94}{HTML}{481769}
\definecolor{v95}{HTML}{471365}
\definecolor{v96}{HTML}{471063}
\definecolor{v97}{HTML}{460b5e}
\definecolor{v98}{HTML}{46085c}
\definecolor{v99}{HTML}{450457}
\definecolor{v100}{HTML}{440154}

\newcommand{\gc}[1]{%
    \ifdim#1 pt < 0.01 pt v100%
    \else\ifdim#1 pt < 0.02 pt v99%
    \else\ifdim#1 pt < 0.03 pt v98%
    \else\ifdim#1 pt < 0.04 pt v97%
    \else\ifdim#1 pt < 0.05 pt v96%
    \else\ifdim#1 pt < 0.06 pt v95%
    \else\ifdim#1 pt < 0.07 pt v94%
    \else\ifdim#1 pt < 0.08 pt v93%
    \else\ifdim#1 pt < 0.09 pt v92%
    \else\ifdim#1 pt < 0.10 pt v91%
    \else\ifdim#1 pt < 0.11 pt v90%
    \else\ifdim#1 pt < 0.12 pt v89%
    \else\ifdim#1 pt < 0.13 pt v88%
    \else\ifdim#1 pt < 0.14 pt v87%
    \else\ifdim#1 pt < 0.15 pt v86%
    \else\ifdim#1 pt < 0.16 pt v85%
    \else\ifdim#1 pt < 0.17 pt v84%
    \else\ifdim#1 pt < 0.18 pt v83%
    \else\ifdim#1 pt < 0.19 pt v82%
    \else\ifdim#1 pt < 0.20 pt v81%
    \else\ifdim#1 pt < 0.21 pt v80%
    \else\ifdim#1 pt < 0.22 pt v79%
    \else\ifdim#1 pt < 0.23 pt v78%
    \else\ifdim#1 pt < 0.24 pt v77%
    \else\ifdim#1 pt < 0.25 pt v76%
    \else\ifdim#1 pt < 0.26 pt v75%
    \else\ifdim#1 pt < 0.27 pt v74%
    \else\ifdim#1 pt < 0.28 pt v73%
    \else\ifdim#1 pt < 0.29 pt v72%
    \else\ifdim#1 pt < 0.30 pt v71%
    \else\ifdim#1 pt < 0.31 pt v70%
    \else\ifdim#1 pt < 0.32 pt v69%
    \else\ifdim#1 pt < 0.33 pt v68%
    \else\ifdim#1 pt < 0.34 pt v67%
    \else\ifdim#1 pt < 0.35 pt v66%
    \else\ifdim#1 pt < 0.36 pt v65%
    \else\ifdim#1 pt < 0.37 pt v64%
    \else\ifdim#1 pt < 0.38 pt v63%
    \else\ifdim#1 pt < 0.39 pt v62%
    \else\ifdim#1 pt < 0.40 pt v61%
    \else\ifdim#1 pt < 0.41 pt v60%
    \else\ifdim#1 pt < 0.42 pt v59%
    \else\ifdim#1 pt < 0.43 pt v58%
    \else\ifdim#1 pt < 0.44 pt v57%
    \else\ifdim#1 pt < 0.45 pt v56%
    \else\ifdim#1 pt < 0.46 pt v55%
    \else\ifdim#1 pt < 0.47 pt v54%
    \else\ifdim#1 pt < 0.48 pt v53%
    \else\ifdim#1 pt < 0.49 pt v52%
    \else\ifdim#1 pt < 0.50 pt v51%
    \else\ifdim#1 pt < 0.51 pt v50%
    \else\ifdim#1 pt < 0.52 pt v49%
    \else\ifdim#1 pt < 0.53 pt v48%
    \else\ifdim#1 pt < 0.54 pt v47%
    \else\ifdim#1 pt < 0.55 pt v46%
    \else\ifdim#1 pt < 0.56 pt v45%
    \else\ifdim#1 pt < 0.57 pt v44%
    \else\ifdim#1 pt < 0.58 pt v43%
    \else\ifdim#1 pt < 0.59 pt v42%
    \else\ifdim#1 pt < 0.60 pt v41%
    \else\ifdim#1 pt < 0.61 pt v40%
    \else\ifdim#1 pt < 0.62 pt v39%
    \else\ifdim#1 pt < 0.63 pt v38%
    \else\ifdim#1 pt < 0.64 pt v37%
    \else\ifdim#1 pt < 0.65 pt v36%
    \else\ifdim#1 pt < 0.66 pt v35%
    \else\ifdim#1 pt < 0.67 pt v34%
    \else\ifdim#1 pt < 0.68 pt v33%
    \else\ifdim#1 pt < 0.69 pt v32%
    \else\ifdim#1 pt < 0.70 pt v31%
    \else\ifdim#1 pt < 0.71 pt v30%
    \else\ifdim#1 pt < 0.72 pt v29%
    \else\ifdim#1 pt < 0.73 pt v28%
    \else\ifdim#1 pt < 0.74 pt v27%
    \else\ifdim#1 pt < 0.75 pt v26%
    \else\ifdim#1 pt < 0.76 pt v25%
    \else\ifdim#1 pt < 0.77 pt v24%
    \else\ifdim#1 pt < 0.78 pt v23%
    \else\ifdim#1 pt < 0.79 pt v22%
    \else\ifdim#1 pt < 0.80 pt v21%
    \else\ifdim#1 pt < 0.81 pt v20%
    \else\ifdim#1 pt < 0.82 pt v19%
    \else\ifdim#1 pt < 0.83 pt v18%
    \else\ifdim#1 pt < 0.84 pt v17%
    \else\ifdim#1 pt < 0.85 pt v16%
    \else\ifdim#1 pt < 0.86 pt v15%
    \else\ifdim#1 pt < 0.87 pt v14%
    \else\ifdim#1 pt < 0.88 pt v13%
    \else\ifdim#1 pt < 0.89 pt v12%
    \else\ifdim#1 pt < 0.90 pt v11%
    \else\ifdim#1 pt < 0.91 pt v10%
    \else\ifdim#1 pt < 0.92 pt v9%
    \else\ifdim#1 pt < 0.93 pt v8%
    \else\ifdim#1 pt < 0.94 pt v7%
    \else\ifdim#1 pt < 0.95 pt v6%
    \else\ifdim#1 pt < 0.96 pt v5%
    \else\ifdim#1 pt < 0.97 pt v4%
    \else\ifdim#1 pt < 0.98 pt v3%
    \else\ifdim#1 pt < 0.99 pt v2%
    \else v1%
    \fi\fi\fi\fi\fi\fi\fi\fi\fi\fi\fi\fi\fi\fi\fi\fi\fi\fi\fi\fi\fi\fi\fi\fi\fi\fi\fi\fi\fi\fi\fi\fi\fi\fi\fi\fi\fi\fi\fi\fi\fi\fi\fi\fi\fi\fi\fi\fi\fi\fi\fi\fi\fi\fi\fi\fi\fi\fi\fi\fi\fi\fi\fi\fi\fi\fi\fi\fi\fi\fi\fi\fi\fi\fi\fi\fi\fi\fi\fi\fi\fi\fi\fi\fi\fi\fi\fi\fi\fi\fi\fi\fi\fi\fi\fi\fi\fi\fi\fi
}

\newcommand{\mc}[1]{
    \cellcolor{\gc{#1}!50} #1
}

\newdimen\figrasterwd
\figrasterwd\textwidth

\newcommand{\Rom}[1]{\uppercase\expandafter{\romannumeral #1\relax}}

\DeclareMathOperator{\Tr}{Tr}

\definecolor{MyForestGreen}{RGB}{34, 139, 34}
\definecolor{MyRedOrange}{RGB}{255, 69, 0}
\newcommand{\y}{\textcolor{MyForestGreen}{\Checkmark}}
\newcommand{\n}{\textemdash}

\definecolor{TableGray}{RGB}{240, 240, 240}
\definecolor{TableWhite}{RGB}{253, 253, 253}

\newcommand{\Rlogo}{\protect\includegraphics[height=1.8ex,keepaspectratio]{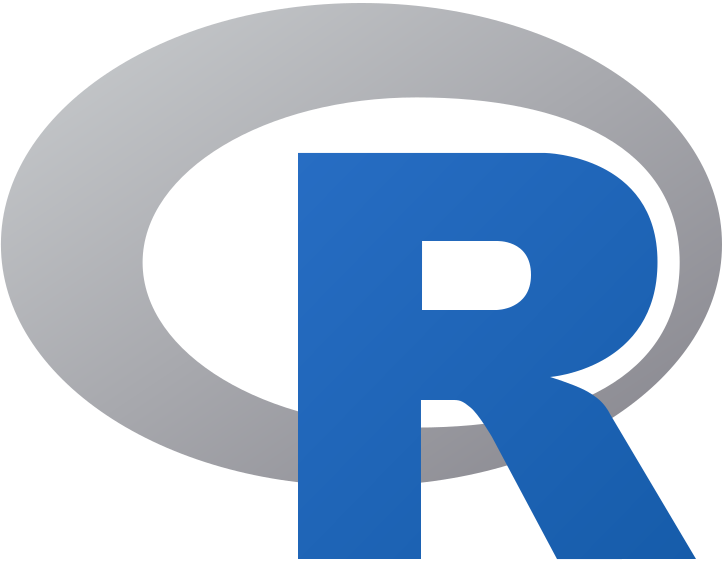}}

\usepackage{listings}
\begin{document}
	\title{On the Use of Relative Validity Indices for Comparing Clustering Approaches}

	\author{Luke W. Yerbury, Ricardo J.G.B. Campello, G. C. Livingston Jr,\\Mark Goldsworthy, Lachlan O'Neil}
	\date{}

        \maketitle
	\begin{abstract}
\noindent Relative Validity Indices (RVIs) such as the Silhouette Width Criterion, Calinski-Harabasz and Davies Bouldin indices are the most widely used tools for evaluating and optimising clustering outcomes. Traditionally, their ability to rank collections of candidate dataset partitions has been used to guide the selection of the number of clusters, and to compare partitions from different clustering algorithms. However, there is a growing trend in the literature to use RVIs when selecting a Similarity Paradigm (SP) for clustering --- the combination of normalisation procedure, representation method, and distance measure which affects the computation of object dissimilarities used in clustering. Despite the growing prevalence of this practice, there has been no empirical or theoretical investigation into the suitability of RVIs for this purpose. Moreover, since RVIs are computed using object dissimilarities, it remains unclear how they would need to be implemented for fair comparisons of different SPs.

This study presents the first comprehensive investigation into the reliability of RVIs for SP selection. We conducted extensive experiments with seven popular RVIs on over 2.7 million clustering partitions of synthetic and real-world datasets, encompassing feature-vector and time-series data. We identified fundamental conceptual limitations undermining the use of RVIs for SP selection, and our empirical findings confirmed this predicted unsuitability. Among our recommendations, we suggest instead that practitioners select SPs by using external validation on high quality labelled
datasets or carefully designed outcome-oriented objective criteria, both of which should be informed by careful consideration of dataset characteristics, and domain requirements. Our findings have important implications for clustering methodology and evaluation, suggesting the need for more rigorous approaches to SP selection in clustering applications.
\end{abstract}


        \nomenclature[Vp]{SP}{Similarity Paradigm}
\nomenclature[Vp]{RVI}{Relative Validity Index}
\nomenclature[Vp]{SWC}{Silhouette Width Criterion}
\nomenclature[Vp]{DBI}{Davies-Bouldin Index}
\nomenclature[Vp]{DI}{Dunn’s Index}
\nomenclature[Vp]{CHI}{Calinski-Harabasz Index}
\nomenclature[Vp]{EVI}{External Validity Index}
\nomenclature[Vp]{ARI}{Adjusted Rand Index}
\nomenclature[Vp]{HMM}{Hidden Markov Model}
\nomenclature[Vp]{PCA}{Principal Component Analysis}
\nomenclature[Vp]{ED}{Euclidean Distance}
\nomenclature[Vp]{MD}{Manhattan Distance}
\nomenclature[Vp]{DTW}{Dynamic Time Warping}
\nomenclature[Vp]{EM}{Expectation-Maximisation}
\nomenclature[Vp]{CI}{C-Index}
\nomenclature[Vp]{AUCC}{Area Under Curve for Clustering}
\nomenclature[Vp]{CD/CoD}{Cosine Distance}
\nomenclature[Vp]{CaD}{Canberra Distance}
\nomenclature[Vp]{BD}{Braycurtis Distance}
\nomenclature[Vp]{ChD}{Chebyshev Distance}
\nomenclature[Vp]{SBD}{Shape-Based Distance}
\nomenclature[Vp]{MSM}{Move-Split-Merge}
\nomenclature[Vp]{TWED}{Time Warping Edit Distance}
\nomenclature[Vp]{MDS}{Multidimensional Scaling}
\nomenclature[Vp]{AMI}{Adjusted Mutual Information}

        \renewcommand{\nomname}{\normalsize Nomenclature}
        \begin{footnotesize}
        \begin{framed}
            \vspace{-0.5cm}
            \printnomenclature
            \vspace{-0.4cm}
        \end{framed}
        \end{footnotesize}
 
	
	\section{Introduction}
\label{Sec:Introduction}

Clustering is a ubiquitous data mining technique with the aim of assembling the objects from a dataset into groups \cite{Aggarwal2014, Gan2007DataApplications}. An effective clustering partitions a dataset such that objects within groups will share a greater notion of similarity or homogeneity than objects in different groups. Such partitions are sought for a variety of purposes across a breadth of domains \cite{Oyewole2023}, which has resulted in a vast array of clustering approaches and components ranging from the generic to the highly task-specific. Broadly speaking, a majority of \textit{clustering approaches} can be disassembled into five fundamental \textit{components}: a data normalisation procedure, a data representation method, a distance measure, a clustering algorithm and a prototype definition \cite{jain1988algorithms,10.1145/331499.331504,Aghabozorgi2015,Aggarwal2012,Zhang2023,Bian2018}. A prototype definition may be enforced within a subroutine of a clustering algorithm, or as an optional data abstraction post-clustering to compactly represent clusters. These five components should be selected carefully when designing candidate clustering approaches for particular domains and applications, as it is widely accepted that all five can bear significant influence on final clustering outcomes. As the normalisation procedure, representation method and distance measure all directly influence the pairwise similarities between objects in the dataset, one combination of these components in this paper is collectively referred to as a \textit{Similarity Paradigm} (SP).

Due to the unsupervised nature of clustering, there is no way of knowing a priori which combination of clustering components from the extensive literature will most aptly capture the unknown categories in a dataset, and no single combination could be expected to prove superior across many different datasets or domains \cite{Wolpert1997}. Thus the resulting component selection problem is a daunting task, and has been approached in various ways throughout the literature. An understandable reaction to the difficulty of this problem is to select components based on precedents and defaults in the domain and wider clustering literature. This is a common and self-perpetuating system which is unlikely to be optimal in general. At the very least, consideration of domain knowledge and application aims is recommended for reducing the pool of candidate components \cite{Hennig2015}. Whilst concepts such as cluster stability \cite{Liu2022} and various domain-specific, outcome-targeted tools (such as forecast error \cite{Li2022}) have been applied sporadically, it seems that cluster components tend to be selected by optimising Relative Validity Indices (RVIs).

Whilst the concept of a cluster may differ between distinct contexts, most RVIs attempt to quantify and juxtapose generally desirable notions of clusters, such as \textit{compactness} or ``high similarity" within groups, and \textit{separation} or ``high dissimilarity" between groups. Popular RVIs include the Silhouette Width Criterion (SWC) \cite{Rousseeuw1987Silhouettes:Analysis}, the Davies-Bouldin Index (DBI) \cite{Davies1979AMeasure}, Dunn's Index (DI) \cite{Dunn1974Well-separatedPartitions} and the Calinski-Harabasz Index (CHI) \cite{Calinski1974AAnalysis}. RVIs, according to their individual notions of quality, can be used to rank a set of candidate partitions, and subsequently select the best partition.
These capabilities have frequently been used to infer the ideal number of clusters ($k$) for a dataset, referred to hereon as the \textit{$k$-selection} task. This application aligns with the original purpose of the vast majority of RVIs. It is a small leap to consider that other model selection problems could also be approached using RVIs, such as selecting the ideal SP for clustering a dataset, which we refer to as the \textit{SP-selection} task. However, as far as the authors are aware, no RVIs have been proposed, or shown to be suitable for SP-selection, despite widespread use for this purpose.

The computation of RVIs typically requires a pairwise distance matrix at the very least. Both classic and more contemporary RVIs have overwhelmingly been introduced within the context of feature-vector data embedded in a Euclidean space, using the Euclidean distance (or $L_2$-norm) as the default distance measure. Examples in the literature suggest it is reasonable to change this default measure for RVI computations where the data are incompatible, such as when mining communities in networks \cite{Rabbany2012RelativeAlgorithms}. Indeed, the authors of the CHI acknowledge that there may be good reasons to use other distance functions, even if they are not defined in
terms of the inner products of Euclidean spaces\footnote{However, it should not be expected that the mathematical or statistical properties of an RVI will still hold if it is instead computed with an SP or prototype definition other than that with which the properties were derived.}. This precedent suggests it is likely reasonable to rank and select the best partition out of a collection of candidates
by computing an RVI using the same non-Euclidean distance or SP that was employed for clustering.

Now supposing it were reasonable to use RVIs for SP-selection, it is unclear whether each partition should be evaluated using the same SP applied to obtain the partition (hereafter referred to as the ``matching-SP" evaluation scheme), or whether a single SP should be maintained for all of the partitions (hereafter referred to as the ``fixed-SP" evaluation scheme). Where papers have compared partitions obtained via non-Euclidean paradigms, it is exceedingly rare for them to indicate which of these two options was enforced when computing the RVIs. In \cite{Hamalainen2017}, \cite{Moayedi2019AnTrajectories} and \cite{Trittenbach2019} it was suggested that the same distance measure should be used for evaluation that was used for clustering. This approach is supported by a conjecture in \cite{Li2022} that a fixed evaluation paradigm would likely show a bias towards partitions that were produced with the same or similar paradigms.
In an attempt to avoid this theorised bias, the authors of \cite{Li2022} proposed a combinatorial scheme inspired by \cite{Iglesias2013}, where all combinations of RVI and clustering SP were considered simultaneously for evaluation. All combinations were analysed via a heatmap in \cite{Li2022}, whereas the unique RVIs in \cite{Iglesias2013} cast votes for the best SP. These approaches are highly sensitive to the set of candidate SPs, and are more complicated to interpret. Nonetheless, inspired by this cross-test concept, in this paper we considered a third combinatorial evaluation scheme: the arithmetic mean over all candidate fixed schemes (hereafter referred to as the ``mean-SP'' evaluation scheme).

\paragraph{}
The main goal of the current paper was to determine whether the continued use of RVIs for selecting an optimal SP should be recommended. This was achieved by investigating the following three fundamental questions:
\begin{enumerate}
    \item[(i)] Do RVIs equipped with a fixed-SP evaluation scheme demonstrate an observable bias towards partitions generated from clustering approaches employing the same SP?
    \item[(ii)] Is there a difference in reliability between the fixed-SP, matching-SP and mean-SP evaluation schemes for either of the SP- or $k$-selection tasks?
    \item[(iii)] Are RVIs as reliable for SP-selection as they are for their conventional application of $k$-selection?
\end{enumerate}
Our primary contribution addresses these three questions via extensive experimental investigations, involving two complementary methodologies applied to more than 2.7 million clustering partitions across three diverse batteries of benchmark datasets. This empirical work is complemented by several additional contributions:
\begin{itemize}[label={--}]
    \setlength\itemsep{0em}
    \item a detailed examination of theoretical issues surrounding RVI use for SP-selection;
    \item introducing a comprehensive nomenclature for uniquely describing clustering approaches by their components;
    \item a thorough review of recent literature where RVIs have been applied for SP-selection;
    \item perspectives on the use of RVIs for the SP-selection task from recent novel RVIs, as well as popular software packages with RVI implementations;
    \item practical recommendations for SP-selection in clustering applications, including recommendations for alternatives.
\end{itemize}

The remainder of this paper is organised as follows. \Cref{Sec:LiteratureReview} introduces critical background information and reviews the relevant literature. \Cref{Sec:Whats_the_problem} discusses in more depth the issues associated with using RVIs for SP-selection. \Cref{Sec:Methodology-Main} describes our experimental methodology and \Cref{Sec:Experimental_Results} presents our results. These findings will be discussed in \Cref{Sec:Discussion}, including a review of alternative approaches to the problem of SP-selection. Finally, \Cref{Sec:Conclusion} will conclude the paper. 
        
	
	\section{Background and Literature Review} \label{Sec:LiteratureReview}

Understanding the suitability of RVIs for SP-selection first requires examining three key areas: their current usage in research literature, their implementations in software packages, and their theoretical development through both traditional and novel indices. To begin \Cref{Sec:LiteratureReview}, \Cref{Subsec:FiveComponents} establishes a comprehensive nomenclature for discussing clustering approaches, disambiguating key terminology used throughout the remainder of the paper. \Cref{Subsec:LiteratureReview} then reviews recent literature applying RVIs for SP-selection, and \Cref{Subsec:SoftwareImplementations} discusses RVI implementation details from popular software packages. \Cref{Subsec:RVIs} introduces the popular RVIs chosen for our experiments and examines their intended scope and purpose. Finally, \Cref{Subsec:Novel_RVIs} provides a similar analysis of more recent novel RVIs.

\subsection{The Components of a Clustering Approach} \label{Subsec:FiveComponents}
As mentioned previously, surveys suggest that a considerable number of clustering procedures, techniques or methods can be decomposed into five core \textit{clustering components}, namely a data normalisation procedure, a data representation method, a distance measure, a clustering algorithm and a prototype definition. A single combination of these is collectively referred to within this paper as a \textit{clustering approach} (see \Cref{Fig:ConceptSchematic}).

\textbf{Normalisation} of data prior to clustering has long been recognised to have an impact on clustering outcomes, and even deemed essential for generating meaningful clusters \cite{Keogh2002, Rakthanmanon2012}. The way that normalisation procedures are applied largely depends on the data type. For feature vector clustering, normalisation is usually performed independently to each feature across all objects to produce comparable values. This ensures that the clustering isn't dominated by larger features. For time series, normalisation is often applied independently to the individual objects, as this serves to direct attention toward the shapes of the time series, rather than their amplitudes when clustering. Several normalisation procedures have been suggested \cite{Walesiak2020TheAnalysis, Paparrizos2020}, and the most common methods are $z$-normalisation, which results in normalised values having zero mean and unit variance; min-max normalisation, which scales values to range between zero and one; and unit-norm normalisation, which ensures the $L_p$ vector norm of the normalised values evaluates to one.

\textbf{Representation method} is a catch-all term for any feature-extraction method or transformation of the \textit{raw} data objects which is intended to improve clustering outcomes. An effective representation should emphasise characteristics that are most relevant or informative for grouping. Furthermore, representations are critical for dimensionality reduction in the context of large-scale data, where they moderate the influence of noise and improve the computational efficiency of a clustering approach \cite{Aghabozorgi2015, Wang2013}. Representations are the most domain-specific clustering component, and along with specialist distance measures are often utilised specifically to make complicated data types compatible with the wealth of existing feature-vector clustering tools. Some examples of representation methods include Bidirectional Encoder Representation from Transformers
for text data \cite{Subakti2022}, 
Discrete Wavelet Transform
for time series \cite{Morchen2003}, autoencoder features for images \cite{Yang2019} or Principal Component Analysis for gene micro-array data \cite{Jin2016}. 
Depending on the method, normalisation can be performed before and/or after the representation has been applied to the data.

A \textbf{distance measure} is a function of two data objects which quantifies their ``sameness". They are required for the vast majority of clustering algorithms, and representation methods typically specify a subset of compatible measures. Popular distance measures such as the Euclidean Distance (ED) and Manhattan Distance (MD) \cite{Aggarwal2014} can be applied across a swathe of different data types and representations, while others are more specific, such as MINDIST for the Symbolic Aggregate Approximation (SAX) \cite{Lin2007ExperiencingSeries} or Dynamic Time Warping (DTW) for raw time series \cite{Berndt1994}. Some measures satisfy all of the properties of a mathematical metric including non-negativity, identity of indiscernibles, symmetry, and triangle inequality. These properties can be leveraged to increase computational efficiency \cite{CHEN2004}, but partial violations of them can introduce fruitful flexibility, allowing measures to feature invariances which have significant utility for more complex data types \cite{Dove2023, Paparrizos2015}. A distinction should be recognised between similarity and dissimilarity measures, with each term conveying how large values of the measure are to be interpreted. Similarity measures yield larger values for more similar objects (0 for minimally similar objects), while dissimilarity measures yield larger values for less similar objects (0 for maximally similar objects). All distance measures used in this paper are formulated as dissimilarity measures; thus from hereon the terms will be used interchangeably.

Together, a normalisation procedure, representation method and distance measure determine a unique spatial embedding of a set of data objects, dictating how the objects are positioned relative to one another. Changing any of these components would fundamentally alter the pairwise relationships between objects. One combination of these three components is referred to in this paper as a \textit{similarity paradigm} or SP. It should also be recognised that different parameter settings of one distance measure or representation also demarcate unique SPs. A well-designed SP will be capable of identifying groups in a dataset which na\"ive SPs may be oblivious to (see \Cref{Sec:Whats_the_problem}). These three components of an SP are collected with the remaining two in \Cref{Fig:ConceptSchematic} for a generic clustering approach.

\begin{figure}
    \centering
    \begin{tikzpicture}[
        optionNode/.style = {rounded rectangle, draw=tikz_Blue!90, fill=tikz_Blue!6, very thick, minimum height=30pt, align=center},
        optionNode2/.style = {rounded rectangle, dashed, draw=tikz_Blue!90, fill=tikz_Blue!6, very thick, minimum height=30pt, align=center},
        inputOutputNode/.style = {rounded rectangle, draw=tikz_Green!90, fill=tikz_Green!6, very thick,minimum height=30pt, align=center},
        inputOutputNode2/.style = {rounded rectangle, dashed, draw=tikz_Green!90, fill=tikz_Green!6, very thick,minimum height=30pt, align=center}
        ]

        \node[optionNode]   (Rep)                    {\textcolor{tikz_Blue!90}{Representation}\\\textcolor{tikz_Blue!90}{Method}};
        \node[optionNode]   (Dist)   [right = 0.2cm of Rep]                    {\textcolor{tikz_Blue!90}{Distance}\\\textcolor{tikz_Blue!90}{Measure}};
        \node[optionNode]   (Norm)   [above = 0.2cm of $(Rep.north west)!0.5!(Dist.north east)$]                                         {\textcolor{tikz_Blue!90}{Normalisation}\\\textcolor{tikz_Blue!90}{Procedure}};
        \node[optionNode]   (Algo)   [right = 1.6cm of Norm]  {\textcolor{tikz_Blue!90}{Clustering}\\\textcolor{tikz_Blue!90}{Algorithm}};
        \node[optionNode2]   (Prot)   [below = 0.2cm of Algo]  {\textcolor{tikz_Blue!90}{Prototype}\\\textcolor{tikz_Blue!90}{Definition}};

        \node[below = 0.2cm of $(Rep.south west)!0.5!(Dist.south east)$] (SP) {}; 
        \node[rectangle] [draw=black!45,inner sep=2mm,fit=(Norm) (Rep) (Dist) (SP), very thick] (SP-box) {};
        \node[below right = 0.65cm of SP.south east] (CA) {}; 
        
        \node[rectangle]  [draw=black!45,inner sep=2mm,fit=(Prot) (Algo) (SP-box) (CA), very thick] (CA-box) {};

        \node[above left] at (SP-box.south east){\textcolor{black!45}{Similarity Paradigm}};
        \node[above left] at (CA-box.south east){\textcolor{black!45}{Clustering Approach}};

        \node[inputOutputNode] (Data)  [left = 0.9cm of CA-box.west] {\textcolor{tikz_Green!90}{Raw}\\\textcolor{tikz_Green!90}{Data}};
        
        \node[inputOutputNode] (Part) [right = 0.9cm of $(CA-box.south east)!0.5!(CA-box.north east)$] {\textcolor{tikz_Green!90}{Partition}};
        \node[inputOutputNode2] (ClPr) [right = 0.9cm of $(CA-box.south east)!0.15!(CA-box.north east)$] {\textcolor{tikz_Green!90}{Cluster}\\\textcolor{tikz_Green!90}{Prototypes}};

        \node[optionNode]   (Eval) [right = 0.9cm of $(CA-box.south east)!0.85!(CA-box.north east)$] {\textcolor{tikz_Blue!90}{Evaluation}\\\textcolor{tikz_Blue!90}{Scheme}};
        \node[optionNode]   (RVI)  [right = 1cm of Part]  {\textcolor{tikz_Blue!90}{RVI}};

        \draw[-{Stealth[]},  thick] (Data) to (CA-box.west);
        \draw[-{Stealth[]},  thick] (CA-box.east|-Part.west) -- (Part.west);
        \draw[-{Stealth[]}, dashed,  thick] (CA-box.east|-ClPr.west) -- (ClPr.west);

        \draw[-{Stealth[]},  thick] (Eval.east) to [out=0, in=90] (RVI.north);
        \draw[-{Stealth[]},  dashed, thick] (ClPr.east) to [out=0, in=270] (RVI.south);
        \draw[-{Stealth[]},  thick] (Part.east) to (RVI.west);

        \path (current bounding box.south west) -- (current bounding box.north east) coordinate[midway] (center);
        \matrix [draw, inner sep = 1mm, matrix of nodes, column sep=3mm, row sep = 0.1mm] (m) [below = 2.6cm of center] {
            \textbf{Legend:} &
            \node[rounded rectangle, dashed, draw=black!60, very thick, minimum height=30pt, align=center] {Optional}; & 
            \node[rounded rectangle, fill=tikz_Green!6, very thick,minimum height=30pt] {\textcolor{tikz_Green!90}{Input/Output}}; &
            \node[rounded rectangle, fill=tikz_Blue!6, very thick, minimum height=30pt, align=center] {\textcolor{tikz_Blue!90}{User Controlled}};\\
         };

    \end{tikzpicture}
    \caption{Diagram of a clustering approach with its five constituent components. A prototype definition is applied within some clustering algorithms, and is also required for producing cluster prototypes for some RVIs or various downstream purposes. Evaluation with an RVI requires the user to select whether the evaluation SP is independent of, or matches the SP used to produce the partition. These are referred to as \textit{fixed} and \textit{matching} evaluation schemes respectively.}
    \label{Fig:ConceptSchematic}
\end{figure}

A \textbf{clustering algorithm} is a procedure which utilises a given SP to produce a partition between a set of objects, where objects placed in the same group are more similar than objects separated into different groups. The obtained partitions are typically hard or crisp, indicating that each object belongs to a single cluster, though partial cluster membership can be obtained by calling upon fuzzy clustering algorithms \cite{Aggarwal2014} or probabilistic clustering algorithms such as Expectation Maximisation (EM) \cite{Duda1974PatternCA} with Gaussian Mixture Models (GMMs) \cite{Mclachlan2014}.
Different algorithms typically optimise some local or global objective function which enforces or is informed by a particular concept or model of a cluster. For instance, the perennially popular $k$-means \cite{Aggarwal2014} with ED optimises the Sum of Squared Errors (SSE), which enforces a spherical or globular cluster model. Common threads of cluster models have been used to categorise clustering algorithms into connectivity-based (e.g. agglomerative or divisive hierarchical clustering \cite{Roux2018}), prototype-based (e.g. $k$-means or $k$-medoids \cite{Aggarwal2014}), model-based (e.g. EM), density-based (e.g. DBSCAN \cite{10.1145/2723372.2737792} or OPTICS \cite{10.1145/304182.304187}), graph-based (e.g. Louvain or Leiden \cite{Traag2019})
or grid-based (e.g. STING \cite{10.5555/645923.758369}  or WaveCluster \cite{10.5555/645924.671342}) algorithms \cite{Aghabozorgi2015}, though some algorithms defy singular categorisation (e.g. HDBSCAN* \cite{Campello2015}, which is both density- and connectivity-based). Note that with this nomenclature, clustering with GMMs or Hidden Markov Models (HMMs) \cite{Aggarwal2014} employs the same EM clustering ``algorithm", but applies two different representations.

\textbf{Prototype definitions} serve to optimally represent or summarise the constituents of a group of objects. They are commonly utilised post-clustering for visualisations, summary purposes or as exemplars for downstream applications. They are not explicitly required for most clustering approaches, but are involved within subroutines of some clustering algorithms. Commonly encountered prototypes are revealed in the set of ``$k$-prototypes" algorithms for continuous feature vectors: $k$-means, $k$-medians, and $k$-medoids. As the clustering steps are virtually identical in these three examples, they can effectively be viewed as variants of a single $k$-prototypes clustering ``algorithm" \cite{Rajabi2020}. Many applications benefit from the prototype being an instance from the dataset, as is enforced by the medoid definition. This prevents pathological cases such as prototypes located ``outside" of the cluster, or excessive smoothing. In some situations, the choice of prototype hinges upon the selected representation or distance measure, for which the commonly employed mean and median may not be appropriate. For example, the standard mean is insufficient to represent a set of time series that have been clustered together with DTW since points are not consistently aligned during distance computation, necessitating specialised approaches like DTW barycenter averaging \cite{Niennattrakul2007InaccuraciesData,Ketterlin2011}. Though sometimes used interchangeably with ``prototype", we have reserved the term ``centroid" to refer to some sort of average value according to a norm (e.g. mean for $L_2$, and median for $L_1$ norms).

Certainly there are some clustering approaches which do not atomise completely into all five of these components. As mentioned previously, clustering with HMMs can be viewed as an application of the EM clustering algorithm to a HMM representation of sequences. Any normalisation or prototype definition can be incorporated into this clustering approach, but no distance measure is required or explicitly specified. Regardless, adoption of this nomenclature within the literature is recommended for accurate and coherent communication in clustering applications or proposals for new clustering tools. For instance, the six ``distance measures" compared in \cite{Kalpakis2001} are more accurately described as six representation methods and a single distance measure, namely ED. In \cite{Gere2023}, hierarchical clustering algorithms equipped with different distance measures are described as ``different clustering algorithms", which is used to justify the comparison of different distance measures with RVIs.
This nomenclature also makes the modular nature of clustering approaches apparent, suggesting that components can be combined in potentially useful ways, or that particular components could be isolated and applied in other machine learning tasks such as indexing, classification or anomaly detection.

\subsection{Literature Review: Comparison of Similarity Paradigms with RVIs}\label{Subsec:LiteratureReview}

\begin{table}[ht!]
    \scriptsize
    \rowcolors{1}{TableGray}{TableWhite}
    \begin{adjustbox}{center}
    \begin{tabular}{p{0.62cm}p{0.7cm}p{0.95cm}p{2.27cm}p{6cm}p{3.9cm}} \toprule
        \hiderowcolors \textbf{Ref.} & \textbf{Year} & \textbf{Comp.}\textsuperscript{a} & \textbf{RVIs}\textsuperscript{b} & \textbf{Use} & \textbf{Data Description} \\ \midrule \showrowcolors

        \scriptsize{\cite{Gere2023}} & \scriptsize{2023} & \scriptsize{D} & \scriptsize{SWC} & \scriptsize{Comparison of distances} & \scriptsize{Consumer Preference Data} \\[0.15cm]

        \scriptsize{\cite{Lala2023}} & \scriptsize{2023} & \scriptsize{N} & \scriptsize{SWC, DBI, CHI} & \scriptsize{Comparison of normalisations} & \scriptsize{Music Feature Data} \\[0.15cm]

        \scriptsize{\cite{Abbasimehr2022}} & \scriptsize{2022} & \scriptsize{R} & \scriptsize{SWC} & \scriptsize{Comparison of representations} & \scriptsize{Transaction Data} \\[0.15cm]
        
        \scriptsize{\cite{Eskandarnia2022AnFramework}} & \scriptsize{2022} & \scriptsize{R} & \scriptsize{SWC} & \scriptsize{Comparison of representations} & \scriptsize{Energy Time Series} \\[0.15cm]
        
        \scriptsize{\cite{Zhang2021TimeDistance}} & \scriptsize{2021} & \scriptsize{D} & \scriptsize{SWC} & \scriptsize{Tuning novel distance parameters} & \scriptsize{General Time Series} \\[0.15cm]
        
        \scriptsize{\cite{Renjith2021AData}} & \scriptsize{2021} & \scriptsize{R} & \makecell[tl]{SWC, DBI, CHI, \\DI} & \makecell[tl]{Comparison of representations \\(Values from 42 RVIs presented in appendix)} & \scriptsize{Recommender System Data} \\[0.15cm] 

        \scriptsize{\cite{Wang2021VehicleVehicles}} & \scriptsize{2021} & \scriptsize{R} & \scriptsize{SWC, DBI, CHI} & \makecell[tl]{Tuning novel representation parameters and \\comparison of representations} & \scriptsize{Spatiotemporal Trajectories} \\[0.15cm]
        
        \scriptsize{\cite{Franco2021PerformanceData}} & \scriptsize{2021} & \scriptsize{R, D} & \scriptsize{SWC} & \scriptsize{Comparison of representations} & \scriptsize{Multi-omics Data} \\[0.15cm]

        \scriptsize{\cite{Abbasimehr2021}} & \scriptsize{2021} & \scriptsize{D} & \scriptsize{SWC} & \scriptsize{Comparison of distances} & \scriptsize{Transaction Data} \\[0.15cm]

        \scriptsize{\cite{Jose-Garcia2021}} & \scriptsize{2021} & \scriptsize{D} & \scriptsize{SWC} & \scriptsize{Comparison of distances} & \scriptsize{General Feature Data} \\[0.15cm]
        
        \scriptsize{\cite{Ruiz2020AData}} & \scriptsize{2020} & \scriptsize{D} & \scriptsize{SWC} & \scriptsize{Comparison of distances} & \scriptsize{Energy Time Series} \\[0.15cm]
        
        \scriptsize{\cite{Biabiany2020DesignAntilles}} & \scriptsize{2020} & \scriptsize{D} & \scriptsize{SWC} & \scriptsize{Comparison supporting proposed distance} & \scriptsize{Spatial Climate Data} \\[0.15cm]
        
        \scriptsize{\cite{Kim2020a}} & \scriptsize{2020} & \scriptsize{R, D} & \scriptsize{SWC, DI} & \makecell[tl]{Comparison of distances based on raw and \\transformed time series} & \scriptsize{Energy Time Series} \\[0.15cm] 
        
        \scriptsize{\cite{Ullah2020DeepData}} & \scriptsize{2020} & \scriptsize{R} & \scriptsize{SWC, DBI, CHI} & \scriptsize{Comparison of representations} & \scriptsize{Energy Time Series} \\[0.15cm]

        \scriptsize{\cite{Shamim2020Multi-DomainData}} & \scriptsize{2020} & \scriptsize{R} & \scriptsize{SWC} & \scriptsize{Comparison of representations} & \scriptsize{Energy Time Series} \\[0.15cm]

        \scriptsize{\cite{Kramer2020AnalysisNetworks}} & \scriptsize{2020} & \scriptsize{R} & \makecell[tl]{SWC, DBI, CHI, \\BHI, HLI} & \scriptsize{Comparison of representations} & \scriptsize{Medical Feature Data} \\[0.15cm]

        \scriptsize{\cite{Toussaint2020}, \cite{Toussaint2019}} & \scriptsize{2020, 2019} & \scriptsize{N} & \scriptsize{SWC, DBI, MIA} & \scriptsize{Comparison of clustering approach combinations via a combined index} & \scriptsize{Energy Time Series} \\[0.15cm]
        
        \scriptsize{\cite{Yilmaz2019}} & \scriptsize{2019} & \scriptsize{N, R} & \scriptsize{SWC} & \scriptsize{Comparison of representations} & \scriptsize{Energy Time Series} \\[0.15cm]
        
        \scriptsize{\cite{Aggarwal2019}} & \scriptsize{2019} & \scriptsize{D} & \scriptsize{DBI} & \scriptsize{Comparison of distances} & \scriptsize{Medical Feature Data} \\[0.15cm] 
        
        \scriptsize{\cite{Moayedi2019AnTrajectories}} & \scriptsize{2019} & \scriptsize{D} & \scriptsize{CSI} & \scriptsize{Comparison of distances} & \scriptsize{Spatiotemporal Trajectories} \\[0.15cm]

        \scriptsize{\cite{Trittenbach2019}} & \scriptsize{2019} & \scriptsize{R} & \scriptsize{SWC} & \makecell[tl]{Comparison of aggregation functions and \\aggregation levels} & \scriptsize{Energy Time Series} \\[0.15cm]

        \scriptsize{\cite{Arunachalam2018}} & \scriptsize{2018} & \scriptsize{D} & \makecell[tl]{SWC, DBI, DI, \\XBI} & \scriptsize{Comparison of distances} & \scriptsize{Consumer Questionnaire Data} \\[0.15cm]
        
        \scriptsize{\cite{Gu2017AClustering}} & \scriptsize{2017} & \scriptsize{D} & \scriptsize{CHI} & \scriptsize{Comparison supporting proposed distance} & \scriptsize{High-Dimensional Feature Data} \\[0.15cm]

        \scriptsize{\cite{Singh2017}} & \scriptsize{2017} & \scriptsize{D} & \scriptsize{SWC} & \scriptsize{Comparison of distances} & \scriptsize{General Proportional Data} \\[0.15cm]
        
        \scriptsize{\cite{Acharya2016}} & \scriptsize{2016} & \scriptsize{D} & \makecell[tl]{SWC, DBI, XBI, \\PBM, FCM} & \makecell[tl]{Multiobjective optimisation criterion for \\automatic selection and comparison of distances} & \scriptsize{Gene Expression Data} \\[0.15cm] 
        
        \scriptsize{\cite{Kim2016AnPathways}} & \scriptsize{2016} & \scriptsize{D} & \scriptsize{SWC, DI} & \scriptsize{Comparison of distances} & \scriptsize{Metabolic Pathways Data} \\[0.15cm] 
        
        \scriptsize{\cite{uczak2016}} & \scriptsize{2016} & \scriptsize{D} & \scriptsize{CHI, MHG, IGV} & \makecell[tl]{Tuning proposed distance parameter with \\transformed indices} & \scriptsize{General Time Series} \\[0.15cm]

        \scriptsize{\cite{Tasdemir2015}} & \scriptsize{2015} & \scriptsize{D} & \makecell[tl]{SWC, DBI, CHI, \\GDI, PBM, CNI} & \scriptsize{Comparison of distances} & \scriptsize{General Feature Data} \\[0.15cm]

        \scriptsize{\cite{Williams2013ClusteringProfiles}} & \scriptsize{2013} & \scriptsize{N, R, D} & \makecell[tl]{SWC, DBI, CHI, \\DI, MIA} & \makecell[tl]{Comparison of 6,992 combinations of distances, \\normalisations and representations} & \scriptsize{Energy Time Series} \\[0.15cm]

        \scriptsize{\cite{Williams-DeVane2013DecisionEndotypes}} & \scriptsize{2013} & \scriptsize{N, D} & \scriptsize{SWC, BHI, HLI} & \scriptsize{Comparison of clustering approach combinations} & \scriptsize{Gene and Feature Data} \\[0.15cm]

        \scriptsize{\cite{Anzanello2011}} & \scriptsize{2011} & \scriptsize{R} & \scriptsize{SWC} & \scriptsize{Variable Selection} & \scriptsize{Production Feature Data} \\[0.15cm]
        
        \scriptsize{\cite{Al-Mubaid2008ComparisonClustering}} & \scriptsize{2008} & \scriptsize{D} & \scriptsize{SWC} & \scriptsize{Comparison of distances} & \scriptsize{Bioinformatics} \\[0.15cm] 
        
        \scriptsize{\cite{Kalpakis2001}} & \scriptsize{2001} & \scriptsize{N, R} & \scriptsize{SWC} & \scriptsize{Comparison of representations} & \scriptsize{ARIMA Time Series} \\[0.15cm] \bottomrule \hiderowcolors

        \multicolumn{6}{p{\dimexpr\textwidth-2\tabcolsep-2\arrayrulewidth-1.5cm}}{ \textsuperscript{a} \scriptsize Distance Measure (D), Normalisation Procedure (N), Representation Method (R)} \\
        \multicolumn{6}{p{\dimexpr\textwidth-2\tabcolsep-2\arrayrulewidth-1.5cm}}{ \textsuperscript{b} \scriptsize Baker-Hubert Index (BHI), Calinski-Harabasz Index (CHI), Chou-Su Index (CSI), Connectivity Index (CNI), Davies-Bouldin Index (DBI), Dunn Index (DI), Fuzzy C-Means Objective Function (FCM), Generalised Dunn Index (GDI), Hubert-Levine Index (HLI), Intra-Group Variance (IGV), Mean Index Adequacy (MIA), Modified Hubert's Gamma (MHG), Pakhira-Bandyopadhyay-Maulik Index (PBM), Silhouette Width Criterion (SWC), Xie-Beni Index (XBI)} \\
    \end{tabular}
    \end{adjustbox}
    \caption{This table summarises a selection of instances where RVIs have been used to compare partitions obtained from different SPs. The columns include the reference, year, compared clustering component, RVI used for the comparison, additional comparison details, and a description of the involved data.}
    \label{Tab:LiteratureSurvey}
\end{table}


\Cref{Tab:LiteratureSurvey} collects instances from the literature where different SPs have been selected or compared using RVIs. Our intent is not to criticise these studies, but rather to document the widespread use of this intuitively appealing practice. The collected papers were primarily case-studies of clustering applications or comparative studies. As evidenced in \Cref{Tab:LiteratureSurvey}, this comparative philosophy has been applied across a broad range of disciplines and seems to have become accepted as commonplace. This broad adoption of RVIs for SP comparison has occurred despite the lack of any theoretical justification for such use. The current study provides a much-needed rigorous evaluation.

It should also be pointed out that while some studies relied upon a single RVI to compare different SPs (with the SWC index being particularly prevalent), better studies used multiple RVIs, or incorporated other comparative methods (such as those discussed in \Cref{Sec:Discussion}). This use of multiple RVIs follows recommendations established in numerous comparative studies \cite{Vendramin2010RelativeOverview,Arbelaitz2013AnIndices,Jain2022WhichProfiles,Jaskowiak2016OnCriteria}. This practice acknowledges the ``no free lunch'' principle \cite{Wolpert1997}, where each RVI's strengths in certain scenarios are balanced by weaknesses in others.

\subsection{Popular Software Implementations}\label{Subsec:SoftwareImplementations}

\begin{table}[ht!]
    \scriptsize
    \rowcolors{1}{TableGray}{TableWhite}
    \begin{adjustbox}{center}
    \begin{tabular}{p{0.35cm} p{1.37cm} >{\centering\arraybackslash}p{1.1cm} >{\centering\arraybackslash}p{0.47cm} >{\centering\arraybackslash}p{1.55cm} >{\centering\arraybackslash}p{0.12cm} >{\centering\arraybackslash}p{0.12cm} >{\centering\arraybackslash}p{0.12cm} >{\centering\arraybackslash}p{0.12cm} >{\centering\arraybackslash}p{0.12cm} >{\centering\arraybackslash}p{1.82cm} p{5.75cm}} \toprule \hiderowcolors
        \multicolumn{1}{l}{} & \multicolumn{1}{l}{} & \multicolumn{1}{l}{} & \multicolumn{1}{l}{} & \multicolumn{6}{c}{\textbf{Automatic Comparison}\textsuperscript{a}} & \multicolumn{1}{l}{}\\ \cmidrule(l){5-10}
        \textbf{Ref.} & \multicolumn{1}{>{\centering\arraybackslash}p{1.37cm}}{\textbf{Toolkit}} & \multicolumn{1}{>{\centering\arraybackslash}p{1.1cm}}{\textbf{Language}} & \multicolumn{1}{>{\centering\arraybackslash}p{0.47cm}}{\textbf{RVIs}} & \textbf{Function} & \textbf{$k$} & \textbf{A} & \textbf{N} & \textbf{R} & \textbf{D} & \multicolumn{1}{>{\centering\arraybackslash}p{1.82cm}}{\textbf{Evaluation Paradigm}} & \multicolumn{1}{>{\centering\arraybackslash}p{5.75cm}}{\textbf{Notes}} \\ \midrule \showrowcolors

        \scriptsize{\cite{Gagolewski2021a}} & \scriptsize{\texttt{genieclust}} & \scriptsize{Python} & \scriptsize{23} & \scriptsize{\n} & \scriptsize{\n} & \scriptsize{\n} & \scriptsize{\n} & \scriptsize{\n} & \scriptsize{\n} & \scriptsize{Default} & \scriptsize{Partitions produced externally, doesn't accept a distance matrix.} \\[0.15cm] 

        \scriptsize{\cite{B.2018}} & \scriptsize{\texttt{clusterCrit}} & \scriptsize{R} & \scriptsize{28} & \scriptsize{\n} & \scriptsize{\n} & \scriptsize{\n} & \scriptsize{\n} & \scriptsize{\n} & \scriptsize{\n} & \scriptsize{Default} & \scriptsize{Partitions produced externally, doesn't accept a distance matrix.} \\[0.15cm] 

        \scriptsize{\cite{Cedex2023}} & \scriptsize{\texttt{Clusters-Features}} & \scriptsize{Python} & \scriptsize{40} & \scriptsize{\n} & \scriptsize{\n} & \scriptsize{\n} & \scriptsize{\n} & \scriptsize{\n} & \scriptsize{\n} & \scriptsize{Default} & \scriptsize{Partitions produced externally, doesn't accept a distance matrix.} \\[0.15cm] 

        \scriptsize{\cite{Nieweglowski2023}} & \scriptsize{\texttt{clv}} & \scriptsize{R} & \scriptsize{8} & \scriptsize{\n} & \scriptsize{\n} & \scriptsize{\n} & \scriptsize{\n} & \scriptsize{\n} & \scriptsize{\n} & \scriptsize{Flexible (S)} & \scriptsize{Partitions produced externally and can either specify one of 3 distances or provide a distance matrix for evaluation.} \\[0.15cm] 
        
        \scriptsize{\cite{CEBECI2020}} & \scriptsize{\texttt{fcvalid}} & \scriptsize{R} & \scriptsize{18} & \scriptsize{\n} & \scriptsize{\n} & \scriptsize{\n} & \scriptsize{\n} & \scriptsize{\n} & \scriptsize{\n} & \scriptsize{Default (NS)} & \scriptsize{Partitions produced externally by \texttt{ppclust} package with one of 18 distances.} \\[0.15cm] 

        \scriptsize{\cite{Dimitriadou2020}} & \scriptsize{\texttt{cclust}} & \scriptsize{R} & \scriptsize{14} & \scriptsize{\n} & \scriptsize{\n} & \scriptsize{\n} & \scriptsize{\n} & \scriptsize{\n} & \scriptsize{\n} & \scriptsize{Default (NS)} & \scriptsize{Partitions produced externally by ``cclust" function (from \texttt{flexclust} package) with one of 2 distance measures.} \\[0.15cm] 
    
        \scriptsize{\cite{Jose-Garcia2023}} & \scriptsize{\texttt{CVIK}} & \scriptsize{\textsc{Matlab}} & \scriptsize{28} & \scriptsize{\textit{evalcvi()}} & \scriptsize{\y} & \scriptsize{\n} & \scriptsize{\n} & \scriptsize{\n} & \scriptsize{\n} & \scriptsize{Default (NS)} & \scriptsize{Partitions produced internally or externally. Features 2 evolutionary algorithms optimising a single requested RVI. User can select one combination from 3 algorithms (5 hierarchical linkages) and 6 distances.} \\[0.15cm] 

        \scriptsize{\cite{Mathworks2013}} & \scriptsize{\texttt{evalclusters}} & \scriptsize{\textsc{Matlab}} & \scriptsize{4} & \scriptsize{\textit{evalclusters()}} & \scriptsize{\y}  & \scriptsize{\n} & \scriptsize{\n} & \scriptsize{\n} & \scriptsize{\n} & \scriptsize{Matching (S)} & \scriptsize{Partitions produced internally. User can select one combination from 3 algorithms (2 hierarchical linkages) and 7 distances.} \\[0.15cm] 

        \scriptsize{\cite{Wang2009}} & \scriptsize{\texttt{CVAP}} & \scriptsize{\textsc{Matlab}} & \scriptsize{14} & \scriptsize{Graphical Interface} & \scriptsize{\y} & \scriptsize{\n} & \scriptsize{\n} & \scriptsize{\n} & \scriptsize{\n} & \scriptsize{Matching (NS)} & \scriptsize{Partitions produced internally or externally. User can select one combination from 5 algorithms (3 hierarchical linkages) and 2 distances.} \\[0.15cm] 

        \scriptsize{\cite{Taskesen_clusteval_is_a_2020}} & \scriptsize{\texttt{clusteval}} & \scriptsize{Python} & \scriptsize{3} & \scriptsize{\textit{clusteval()}} & \scriptsize{\y} & \scriptsize{\n} & \scriptsize{\n} & \scriptsize{\n} & \scriptsize{\n} & \scriptsize{Default (S)} & \scriptsize{Partitions produced internally. User can select one combination from 4 algorithms (7 hierarchical linkages) and 22 scikit-learn distances.} \\[0.15cm] 

        \scriptsize{\cite{Charrad2014Nbclust:Set}} & \scriptsize{\texttt{NbClust}} & \scriptsize{R} & \scriptsize{30} & \scriptsize{\textit{NbClust()}} & \scriptsize{\y} & \scriptsize{\n} & \scriptsize{\n} & \scriptsize{\n} & \scriptsize{\n}& \scriptsize{Matching (NS)} & \scriptsize{Partitions produced internally. User can select one combination from 2 algorithms (8 hierarchical linkages) and 6 built-in distances (or can provide distance matrix).} \\[0.15cm] 
        
        \scriptsize{\cite{Brock2008ClValid:Validation}} & \scriptsize{\texttt{clValid}} & \scriptsize{R} & \scriptsize{3} & \scriptsize{\textit{clValid()}} & \scriptsize{\y} & \scriptsize{\y} & \scriptsize{\n} & \scriptsize{\n} & \scriptsize{\n} & \scriptsize{Matching (NS)} & \scriptsize{Partitions produced internally or externally. User can select combinations of 10 algorithms (4 hierarchical linkages) and one of 3 distances.} \\[0.15cm] 

        \scriptsize{\cite{Baker2019}} & \scriptsize{\texttt{validclust}} & \scriptsize{Python} & \scriptsize{5} & \scriptsize{\textit{ValidClust()}} & \scriptsize{\y} & \scriptsize{\y} & \scriptsize{\n} & \scriptsize{\n} & \scriptsize{\n} & \scriptsize{Default (NS)} & \scriptsize{Partitions produced internally. User can select one combination from 1 algorithm (4 hierarchical linkages) and 5 distances.} \\[0.15cm] 
        
        \scriptsize{\cite{Walesiak2008}, \cite{Walesiak2020TheAnalysis}} & \scriptsize{\texttt{clusterSim}} & \scriptsize{R} & \scriptsize{6} & \scriptsize{\textit{cluster.Sim()}} & \scriptsize{\y} & \scriptsize{\y} & \scriptsize{\y} & \scriptsize{\n} & \scriptsize{\y} & \scriptsize{Matching (NS)} & \scriptsize{Partitions produced internally or externally. Automatic selection between combinations of 11 normalisations, 10 distances and 3 algorithms (7 hierarchical linkages).} \\[0.15cm] 

        \scriptsize{\cite{Wong2019}} & \scriptsize{\texttt{autocluster}} & \scriptsize{Python} & \scriptsize{3} & \scriptsize{\textit{AutoCluster()}} & \scriptsize{\y} & \scriptsize{\y} & \scriptsize{\n} & \scriptsize{\y} & \scriptsize{\n} & \scriptsize{Matching (NS)} & \scriptsize{Partitions are produced internally. Automatic selection between combinations of 10 algorithms (4 hierarchical linkages) and 7 representations.} \\[0.15cm]
        
        \bottomrule 
        
        \hiderowcolors

        \multicolumn{12}{p{\dimexpr\textwidth-2\tabcolsep-2\arrayrulewidth}}{ \textsuperscript{a} \scriptsize Number of Clusters ($k$), Clustering Algorithm (A), Normalisation Procedure (N), Representation Method (R) and Distance Measure (D)} \\

    \end{tabular}
    \end{adjustbox}
    \caption{Packages and toolboxes from three popular data analysis languages targeting clustering evaluation. The first four columns provide the reference, toolkit name, coding language and the number of RVIs implemented in the toolkit. The automatic comparison column indicates which clustering components the toolkit facilitates the automatic selection, recommendation or comparison of where such a function is offered. The next column indicates whether the package implements a matching evaluation scheme or if the default Euclidean distance is used with the provided data (which may be transformed/normalised). The brackets in this column indicate whether this was specified (S) or not specified (NS) in the documentation. In case of the latter, the choice has been inferred by investigating the source code. The notes column provides further details about the package, such as the number of built-in clustering algorithms, distance measures, etc. Additionally, it is noted whether the package offers functionality to evaluate partitions produced ``internally" within its own functions, and/or partitions produced ``externally" by other means.}
    \label{Tab:PackageSurvey}
\end{table}


\Cref{Tab:PackageSurvey} collects the many packages and toolkits targeting clustering evaluation and RVI computation across three popular data analysis languages - \Rlogo, Python and \textsc{Matlab}.
Among these, only \textsc{Matlab}'s \texttt{evalclusters} function explicitly addresses the choice of evaluation scheme in its documentation, stating that the evaluation distance measure ``must match the distance metric used in the clustering algorithm to obtain meaningful results." For the majority of other packages, determining whether RVIs were computed using matching-SP or fixed-SP schemes required meticulous examination of source code, as this crucial detail was often omitted from documentation. 

The surveyed packages fall into two main categories. Six packages focus solely on RVI computation without built-in clustering capabilities. Of these, most default to a fixed Euclidean evaluation scheme, with only \texttt{clv} offering flexibility in distance measure choice. The remaining nine packages provide both clustering and evaluation capabilities, with a majority (6) using matching evaluation schemes.

While all surveyed packages with clustering capabilities allow optimisation over different numbers of clusters, only two extend this comparison to clustering components other than the clustering algorithm. The titular function of the \texttt{clusterSim} package performs a comparison of all combinations of requested normalisations, distance measures, clustering algorithms and numbers of clusters according to one of six user selected RVIs with a matching evaluation scheme enforced. Meanwhile the titular function from the \textit{AutoCluster} package also enforces matching, with RVIs computed using the same dimensionally reduced versions of the input data and Euclidean distance used during clustering.

\subsection{Selected Relative Validity Indices} \label{Subsec:RVIs}

In this section we introduce the seven RVIs selected to feature in our experiments. These indices have not been chosen to represent the state of the art, rather to capture a range of RVIs that have commonly been used for comparing partitions from different SPs in the literature --- which overwhelmingly describes traditional and well-established RVIs. Thus we have even included the Dunn and Davies-Bouldin indices, despite their notoriety for being amongst the worst performers in multiple benchmark studies \cite{Vendramin2010RelativeOverview, Arbelaitz2013AnIndices}. 

\begin{table}[ht!]
   \scriptsize
   \rowcolors{1}{TableGray}{TableWhite}
   \begin{adjustbox}{center}
        \begin{tabular}{p{0.9cm} p{2.1cm} p{0.7cm} >{\centering\arraybackslash}p{0.5cm} >{\centering\arraybackslash}p{0.5cm} p{7cm} >{\centering\arraybackslash}p{0.8cm} p{1.9cm} } \toprule \hiderowcolors

       \textbf{Ref.} & \textbf{Name} & \textbf{Abbr.} & \textbf{Opt.} & \textbf{Inv.} & \textbf{Description} & \textbf{SP}\textsuperscript{a} & \makecell[tl]{\textbf{Advocated}\\ \textbf{Comparisons}\textsuperscript{b}} \\ \midrule 
       \multicolumn{8}{c}{\textcolor{gray}{\textbf{Prototype-Insensitive}}} \\ \cmidrule(l){1-8} \showrowcolors
       
       \cite{Rousseeuw1987Silhouettes:Analysis} & \makecell[tl]{Silhouette Width \\Criterion} & \scriptsize{SWC} & \scriptsize{Max} & \scriptsize{\y} & \scriptsize{Mean normalised difference between average distance to objects in same cluster and average distance to objects in nearest other cluster} & \scriptsize{Any} & \scriptsize{$k$, A} \\[0.15cm]
       
       \cite{Dunn1974Well-separatedPartitions},\cite{Bezdek1995} & \scriptsize{Dunn Index} & \scriptsize{DI} & \scriptsize{Max} & \scriptsize{\y} & \scriptsize{Ratio of minimum set distance between clusters to maximum cluster diameter} & \scriptsize{Any} & \scriptsize{$k$, A} \\[0.15cm]
       
       \makecell[tl]{\cite{Dalrymple-Alford1970MeasurementRecall},\cite{Hubert1976ARecall},\\ \cite{Hubert1985},\cite{Bezdek2016TheValidity}}& \scriptsize{C-Index} & \scriptsize{CI} & \scriptsize{Min} & \scriptsize{\y} & \scriptsize{Compares the normalised sum of within-cluster distances against theoretical best and worst cases} & \scriptsize{Any} & \scriptsize{$k$} \\[0.15cm]
       
       \cite{Jaskowiak2022} & \makecell[tl]{Area Under Curve \\for Clustering} & \scriptsize{AUCC} & \scriptsize{Max} & \scriptsize{\y} & \scriptsize{Based on ROC analysis of clustering viewed as binary classification using normalised pairwise distances} & \scriptsize{Match} & \scriptsize{$k$, A} \\[0.15cm]
       
       \cmidrule(l){1-8} \hiderowcolors
       \multicolumn{8}{c}{\textcolor{gray}{\textbf{Prototype-Sensitive}}} \\ \cmidrule(l){1-8} \showrowcolors
       
       \cite{Calinski1974AAnalysis} & \makecell[tl]{Calinski-Harabasz \\Index} & \scriptsize{CHI} & \scriptsize{Max} & \scriptsize{\y} & \scriptsize{Ratio of between-cluster to within-cluster dispersion with normalisation for number of clusters} & \scriptsize{ED} & \scriptsize{$k$} \\[0.15cm]
       
       \cite{Davies1979AMeasure} & \makecell[tl]{Davies-Bouldin \\Index} & \scriptsize{DBI} & \scriptsize{Min} & \scriptsize{\y} & \scriptsize{Average of worst within-to-between cluster spread ratios for each cluster} & \scriptsize{ED} & \scriptsize{$k$, A} \\[0.15cm]
       
       \cite{Pakhira2004ValidityClusters} & \makecell[tl]{Pakhira-\\Bandyopadhyay-\\Mauli Index} & \scriptsize{PBM} & \scriptsize{Max} & \scriptsize{\n} & \scriptsize{Quotient of cluster separation and compactness ratios with cluster number penalty} & \scriptsize{ED} & \scriptsize{$k$} \\[0.15cm]
       
       \bottomrule
       \hiderowcolors
       
       \multicolumn{8}{p{14cm}}{\textsuperscript{a} \scriptsize Euclidean Distance (ED) \, \textsuperscript{b} \scriptsize Number of Clusters ($k$), Clustering Algorithm (A)} \\
   \end{tabular}
   \end{adjustbox}
   \caption{This table summarises key details of the seven RVIs selected for our experiments. RVIs are grouped by their sensitivity to prototype definitions. The columns include key references, name and abbreviation, optimization direction, invariance to multiplicative scaling, a brief description of their computation, the SP applied when formulating/computing the RVI, and the comparative context explicitly supported by the authors.}
   \label{Tab:SelectedRVIsSummary}
\end{table}

The seven RVIs are presented in \Cref{Tab:SelectedRVIsSummary}, grouped by their sensitivity to cluster prototypes. This grouping reflects how the RVIs interact with prototype definitions. Take for example, a hierarchical clustering of Euclidean feature vectors --- while prototype definitions do not affect the clustering process, they can be applied to the resulting partitions. Values subsequently calculated for prototype-sensitive RVIs would change if the prototype definition were changed, while prototype-insensitive RVIs would remain unchanged. For $k$-prototype algorithms, the distinction is more complicated as the prototype is actively involved in the clustering process. Changing the prototype definition in this case would likely change the partition, as well as its evaluation with prototype-sensitive RVIs.

\Cref{Tab:SelectedRVIsSummary} provides other important details for the seven RVIs including: the original reference and other key references involved in the development of the index, whether the index should be maximised or minimised, whether the index is invariant to multiplicative scaling of the pairwise distances, a description of their formulation, the SP the index was formulated in or intended to be used with, and the kinds of comparisons that the authors explicitly advocated using their index for. Further discussion of these details and specifics regarding the computation of each RVI can be found in Appendix \hyperref[Appendix:RVIs]{A}.

It should be noticed that while these RVIs have been widely used to compare different SPs, they were initially proposed for ranking partitions with different numbers of clusters obtained under the same SP, with some explicitly suggesting the partitions could also be derived from different clustering algorithms. Although this doesn't necessarily preclude these indices from effectively comparing different SPs, such usage requires theoretical or empirical support. Our study is therefore particularly important given the widespread adoption of this practice, as discussed in \Cref{Subsec:LiteratureReview}.

Interestingly, there is a clear distinction regarding the pairwise distances accepted by the two groups of cluster prototypes: the prototype-insensitive RVIs were all recommended for use with any SP, while prototype-sensitive RVIs were specifically formulated within a Euclidean data space. This is likely because a Euclidean space is required to make the typical average centroid meaningful. Nearly all of the RVIs were scale-invariant by design, with PBM being the only exception.

\subsection{Recent Novel Relative Validity Indices} \label{Subsec:Novel_RVIs}
To investigate whether any more recently proposed RVIs were specifically designed for the task of SP-selection, we reviewed a sample of recent papers proposing novel RVIs --- collected in \Cref{Tab:ModernRVIs}. This table records: the name of each index, the year it was proposed, the SP the RVIs were formulated within or intended to be used with, and the kinds of comparisons that the authors explicitly suggested their index was intended for. 

\begin{table}[!htb]
    \scriptsize
    \rowcolors{1}{TableGray}{TableWhite}
    \begin{adjustbox}{center}
        \begin{tabular}{p{0.35cm} p{0.42cm} p{3.2cm} p{9.2cm} >{\centering\arraybackslash}p{0.5cm} p{1.9cm}} \toprule \hiderowcolors

        \textbf{Ref.} & \textbf{Year} & \textbf{Name} & \textbf{Description}\textsuperscript{a} & \textbf{SP}\textsuperscript{b} & \makecell[tl]{\textbf{Advocated}\\ \textbf{Comparisons}\textsuperscript{c}}\\ \midrule \showrowcolors

        \scriptsize{\cite{Wiroonsri2024}} & \scriptsize{2024} & \scriptsize{New Correlation Index} & \scriptsize{Based on Pearson correlation between all pairwise distances and the corresponding distances between their prototypes to identify many potential $k$.} & \scriptsize{Any} & \scriptsize{$k$} \\[0.15cm]
        
        \scriptsize{\cite{Scitovski2023}} & \scriptsize{2023} & \scriptsize{Minimal Distance Index}  & \scriptsize{Dunn-type ratio with separation measured using characteristic balls containing a percentage of points, and compactness measured using WCSS} & \scriptsize{ED} & \scriptsize{$k$} \\[0.15cm]

        \scriptsize{\cite{Duan2023}} & \scriptsize{2023} & \makecell[tl]{Augmented Non-Shared \\ Nearest Neighbors \\Cluster Validity Index}  & \scriptsize{Difference of separation and compactness, both defined using representative points which are the non-shared NN of point pairs with fewer shared NN.} & \scriptsize{ED} & \scriptsize{$k$} \\[0.15cm]

        \scriptsize{\cite{DouglasGuimaraesdeAquino2023ADescription}} & \scriptsize{2023} & \makecell[tl]{Support, Length, \\Exclusivity and Difference \\for Group Evaluation Index}  & \scriptsize{The average of four semantic descriptors for binary categorical datasets.} & \scriptsize{NS} & \scriptsize{$k$} \\[0.15cm]

        \scriptsize{\cite{Modak2023}} & \scriptsize{2023} & \scriptsize{$M_{clus}$}  & \scriptsize{Silhouette-type index utilising Gaussian KDE mode estimates in place of the average inter- and intra-cluster distances.} & \scriptsize{Any}  & \scriptsize{$k$, A} \\[0.15cm]

        \scriptsize{\cite{Senol2022}} & \scriptsize{2022} & \makecell[tl]{Validity Index for \\Arbitrary-Shaped Clusters \\Based on the Kernel \\Density Estimation}  &  \scriptsize{KDE weighted normalised difference between compactness and separation for each point, combined in the form of the weighted average Silhouette} & \scriptsize{ED} & \scriptsize{$k$, A} \\[0.15cm]

        \scriptsize{\cite{CarlosRojasThomas2021}} & \scriptsize{2021} & \makecell[tl]{Contiguous Density \\Region Index}  & \makecell[tl]{A weighted sum of normalised differences between local densities and \\average densities within each cluster.} & \scriptsize{ED} & \scriptsize{$k$} \\[0.15cm]
        
        \scriptsize{\cite{Gagolewski2021}} & \scriptsize{2021} & \scriptsize{Generalisd DuNN}  & \scriptsize{Dunn-type ratio utilising ordered weighted averaging operators on the NN graph.} & \scriptsize{ED} & \scriptsize{A} \\[0.15cm]

        \scriptsize{\cite{Ribeiro2021}} & \scriptsize{2021} & \scriptsize{Temporal Gap Statistic} & \scriptsize{Gap statistic adapted for time series by changing distance measure and producing the random reference data via a phase space transformation.} & \scriptsize{DTW} & \scriptsize{$k$} \\[0.15cm]


        \scriptsize{\cite{Li2020a}} & \scriptsize{2020} & \makecell[tl]{Volume and Area based \\Index}  & \makecell[tl]{Area and perimeter based index relying on a 2D MDS embedding and \\spherical transformation.} & \scriptsize{ED} & \scriptsize{$k$} \\[0.15cm]

        \scriptsize{\cite{Guan2020}} & \scriptsize{2020} & \makecell[tl]{Distance-based \\Separability Index}  & \scriptsize{The Kolmogorov-Smirnov similarity between the distributions of intra-cluster distances and inter-cluster distances, averaged across all clusters.} & \scriptsize{ED}  & \scriptsize{$k$} \\[0.15cm]

        \scriptsize{\cite{Xie2020}} & \scriptsize{2020} & \makecell[tl]{Density-Core-based \\Clustering Validation Index}  & \scriptsize{Dunn-type ratio with compactness measured using noise-robust MST based on local density peak set, and minimum single-linkage style separation.} & \scriptsize{NS} & \scriptsize{$k$} \\[0.15cm]

        \scriptsize{\cite{Flexa2019}} & \scriptsize{2019} & \makecell[tl]{Mutual Equidistant-\\Scattering Criterion}  & \scriptsize{Compactness assessed with absolute differences of all pairs of intra-cluster distances. Maximum inter-prototype distance for separation.} & \scriptsize{ED} & \scriptsize{$k$} \\[0.15cm]

        \scriptsize{\cite{Hu2019}} & \scriptsize{2019} & \makecell[tl]{Cluster Validity Index \\based on Density-Involved \\Distance}  & \scriptsize{Dunn-type ratio with density-based measure of separation and MST-based measure of compactness} & \scriptsize{ED} & \scriptsize{$k$, A} \\[0.15cm] 

        \scriptsize{\cite{Cheng2019}} & \scriptsize{2019} & \makecell[tl]{Local Cores-based Cluster \\Validity}  & \scriptsize{Silhouette-type index utilising a graph-based distance between points of local maximum density identified using the concept of natural NN.} & \scriptsize{ED} & \scriptsize{$k$} \\[0.15cm]

        \scriptsize{\cite{Thomas2019}} & \scriptsize{2019} & \scriptsize{Segment Index}  & \scriptsize{Dunn-type index with cluster dispersion measured as the average distance of objects to a line segment across the dimension of maximum variation.} & \scriptsize{ED}  & \scriptsize{$k$} \\[0.15cm]

        \scriptsize{\cite{Lee2018}} & \scriptsize{2018} & \makecell[tl]{Support Vector Data \\Description based RVIs}  & \scriptsize{Versions of DBI, CHI, DI and XBI with compactness measured using a kernel representation.}  & \scriptsize{ED} & \scriptsize{$k$} \\[0.15cm]

        
        \bottomrule 
        
        \hiderowcolors

        \multicolumn{6}{p{15.7cm}}{\textsuperscript{a} \scriptsize Kernel Density Estimate (KDE), Multi-Dimensional Scaling (MDS), Minimum Spanning Tree (MST), Nearest Neighbours (NN), Within-Cluster Sum of Squares (WCSS)} \\
        \multicolumn{6}{p{15.7cm}}{\textsuperscript{b} \scriptsize Dynamic Time Warping (DTW), Euclidean Distance (ED), Not Specified (NS)} \\
        \multicolumn{6}{p{15.7cm}}{\textsuperscript{c} \scriptsize Number of Clusters ($k$), Clustering Algorithm (A)} \\

    \end{tabular}
    \end{adjustbox}
    \caption{This table summarises a selection of recent novel RVIs. The columns include the reference, year, name of the proposed RVI, a brief description of how the RVI is computed, the SP applied when formulating/computing the RVI, and the comparative context explicitly supported by the authors.}
    \label{Tab:ModernRVIs}
\end{table}

It is noteworthy that none of the surveyed RVIs promoted the use of their index for comparing different SPs. Instead, most of these novel RVIs were proposed in an effort to improve upon the $k$-selection performance of traditional RVIs in one or more specific clustering scenarios. Traditional RVIs suffer from well-known limitations, particularly when dealing with noisy data or clusters that exhibit variable density, complex shapes, or sub-structures \cite{Liu2010, Xiong2013, Moulavi2014Density-basedValidation}. They also tend to inadequately penalise large, noisy clusters \cite{Dang-Ha2017ClusteringHvaler-Norway}, or preference partitions dominated by one major cluster or small outlier clusters \cite{Chicco2012OverviewGrouping,Rajabi2020,Dang-Ha2017ClusteringHvaler-Norway}. While such imbalanced partitions can sometimes offer valuable insights, they are rarely intended outcomes. Though these newer indices have shown clear improvements in some of these contexts, the continual influx of new RVIs each year hinders their adoption, as practitioners tend to fall back on popular indices. This topic is ripe for a broad comparative study in the vein of \cite{Vendramin2010RelativeOverview}.

	
	\section{What are the issues?} \label{Sec:Whats_the_problem}

Our literature review revealed that RVIs have been widely adopted for a purpose that: a) was never advocated for by the authors of these indices, and b) they have not been established, either theoretically or experimentally, to be suitable for. In this study, we investigated both the conceptual suitability of RVIs to SP-selection, as well as their empirical performance. We now proceed by examining fundamental conceptual issues concerning the use of RVIs for SP-selection. 

As discussed in \Cref{Sec:Introduction,Sec:LiteratureReview}, there are two basic options when computing RVIs for evaluating and comparing candidate partitions produced by different SPs. The first option is to compute the RVI using a fixed SP, which is most commonly chosen to be the ED applied to raw data. The second option is to compute the RVI using the same SP that was used to produce the partition. These are the two main options for the evaluation scheme in \Cref{Fig:ConceptSchematic}. Whilst the matching-SP scheme is completely prescriptive, any number of fixed-SP schemes are possible beyond the default fixed-ED scheme. We also consider an alternative (mean-SP) evaluation scheme in our experiments, the following discussion focuses on the foundational matching and fixed schemes.

\begin{figure}[p]
    \centering
    \includegraphics[width=0.98\linewidth]{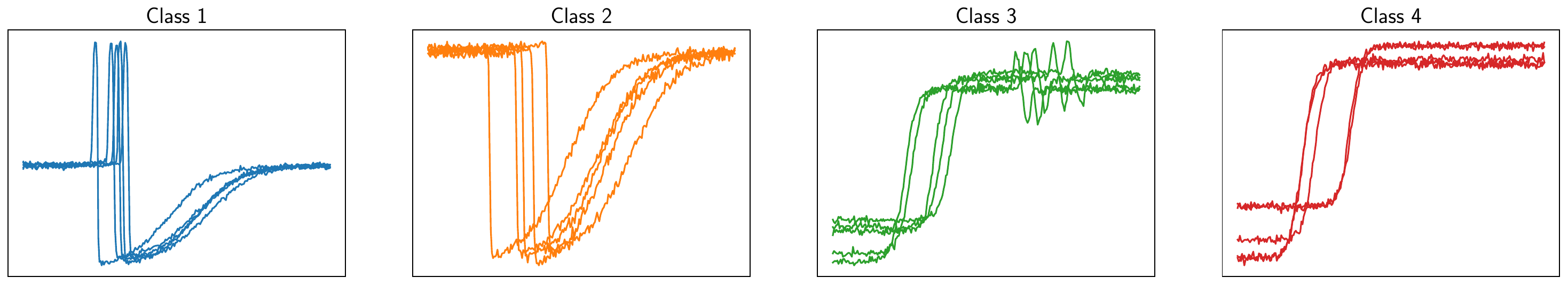}
    \caption{Samples of time series from each of the four classes of the Trace dataset from the UCR archive \cite{Dau2019}.}\label{Fig:Trace_Samples}
\end{figure}

\begin{figure}[p]
    \centering
    \includegraphics[width=0.98\linewidth]{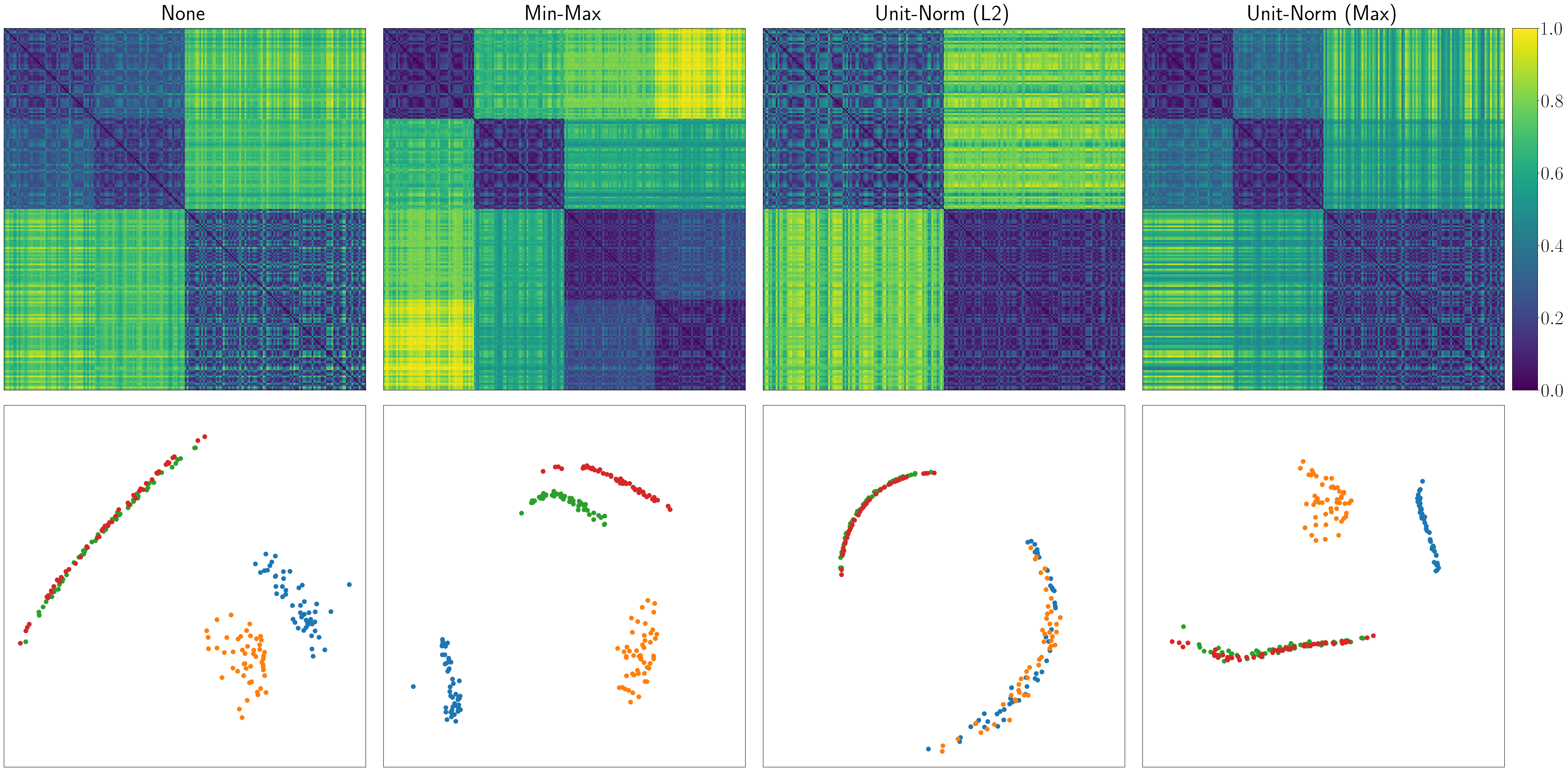}
    \caption{Pairwise distance matrices and corresponding MDS embeddings for the Trace dataset using \textbf{various normalisations}, raw representation and time warping edit distance.}\label{Fig:Norms}
\end{figure}

\begin{figure}[p]
    \centering
    \includegraphics[width=0.98\linewidth]{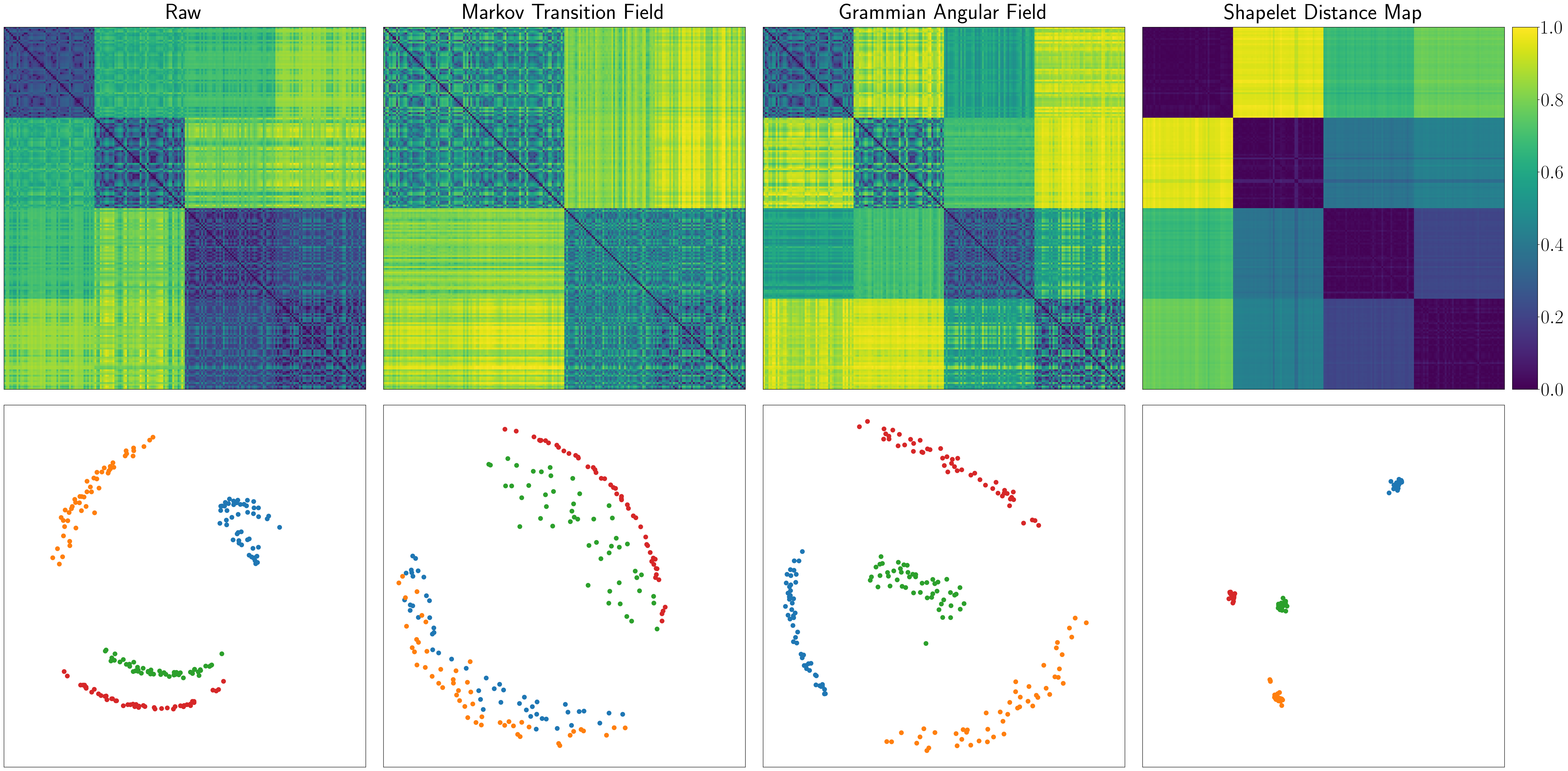}
    \caption{Pairwise distance matrices and corresponding MDS embeddings for the Trace dataset using min-max normalisation, \textbf{various representations} and Euclidean distance.}\label{Fig:Representations}
\end{figure} 

\begin{figure}[t]
    \centering
    \includegraphics[width=0.98\linewidth]{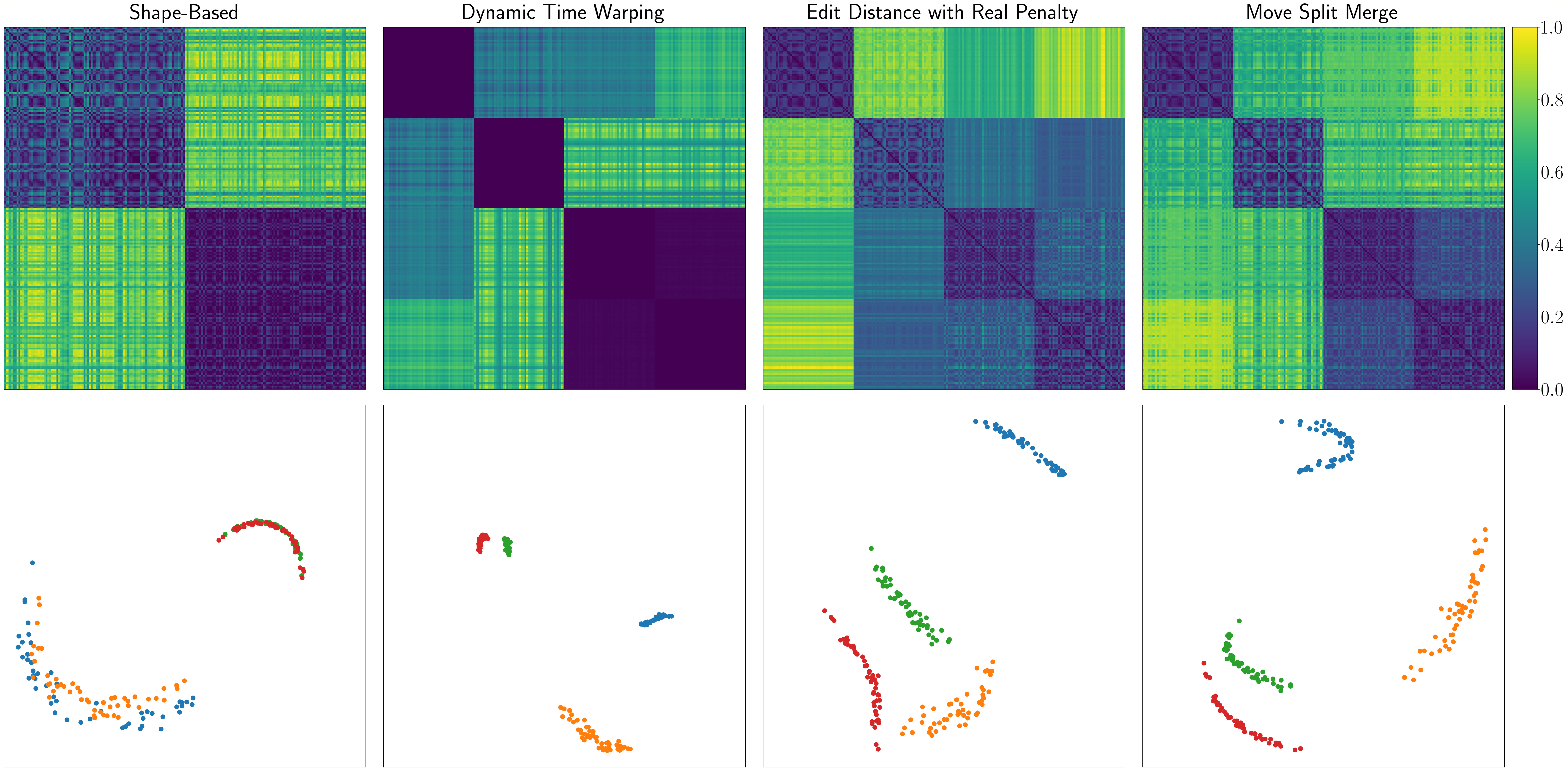}
    \caption{Pairwise distance matrices and corresponding MDS embeddings for the Trace dataset using min-max normalisation, raw representation and \textbf{various distances}.}\label{Fig:Distances}
\end{figure} 

\subsubsection*{Fixed Evaluation Scheme Issues}
It has been suggested that within fixed-SP evaluation schemes, RVIs are likely to demonstrate a bias towards partitions produced with the same or similar SPs \cite{Li2022}. If the fixed SP does not optimally capture the latent similarity structure in the dataset, superior partitions may not be recommended as such by an RVI equipped with this evaluation scheme.

In order to illustrate this, we introduce the Trace dataset from the UCR Time Series Classification Archive \cite{Dau2019}, a labelled synthetic dataset consisting of 200 time series and 4 classes. Sample series from each class are presented in \Cref{Fig:Trace_Samples}. Three sets of pairwise distance matrices have been computed for this dataset, with a single component of the SP varying within each set. These are presented in figures \Cref{Fig:Norms,Fig:Representations,Fig:Distances} where the normalisation procedure, representation method and distance measure have been varied respectively. It should be noted that all of the matrices have been scaled through division by the maximum distance, and the objects have been ordered identically in each matrix according to the ground-truth labels. This reordering reveals the effectiveness of each of the 12 SPs in capturing the similarity structure inherent within the dataset. The darker a pixel in the pairwise distance matrix, the closer the pair of objects. Note that along the main diagonals, all of the distances are zero, as all distances have been formulated as dissimilarity measures. Similar to the Visual Assessment of clustering Tendency (VAT) \cite{Bezdek2002}, dark blocks along the main diagonal are indicative of groups of objects which are close together, i.e. potential clusters. The distance matrices have been paired with one corresponding realisation of a 2D Multi-Dimensional Scaling (MDS) embedding \cite{Borg2005} to provide intuition when interpreting the distance matrices. MDS attempts to find a set of points in a low-dimensional space which optimally preserves the pairwise distances. The colours of the classes in \Cref{Fig:Trace_Samples} are consistent in the MDS plots. Additionally, the order of classes along the $x$ and $y$ axes of the distance matrices matches the order of the classes in \Cref{Fig:Trace_Samples}.

Observing each unique SP in \Cref{Fig:Norms,Fig:Distances,Fig:Representations} reveals that independently varying the normalisation procedure, representation method or distance measure components of an SP will produce complex, non-linear changes to the distribution of pairwise distances. Naturally, this is consistent with the definition of an SP. Some of the considered SPs, such as Unit-Norm (L2) and Unit-Norm (Max) from \Cref{Fig:Norms}, only manage to distinguish the existence of two or three clusters respectively. Even where four clusters have been identified, different SPs do so with varying degrees of cluster compactness and separation. For instance, compare the separation of classes 3 and 4 according to the Markov Transition Field SP in \Cref{Fig:Representations} to the Grammian Angular Field SP in \Cref{Fig:Representations}. Also compare the compactness of each class according to the Grammian Angular Field SP to the Shapelet Distance Map SP in \Cref{Fig:Representations}.

Let's now assume that the four classes represent the optimal partition of the Trace dataset for a practitioner's purposes. It would be ``unfair" to evaluate an accurate 4-cluster partition with a fixed-SP version of an RVI where the evaluation SP only identifies 2 or 3 clusters. The 4-cluster partition would not score well, even if it is objectively superior to a 2- or 3-cluster partition. The case where a hypothetical evaluation SP cannot identify \textit{any} inherent structure at all is worse again. Thus it is very likely that partitions which are more consistent with the similarity structure induced by the fixed evaluation SP will result in superior RVI values. Such partitions are most likely to be produced by clustering algorithms applying this fixed evaluation SP. According to this argument, it would be ``unfair" and unreliable to evaluate partitions produced by distinct SPs according to a single fixed SP. 

This strongly suggests that there is no guarantee that one can reliably select the ``optimal" SP by comparing RVIs computed with a fixed-SP evaluation scheme. That is unless one could know \textit{a priori} which SP would best capture the latent similarity structure of a dataset. Out of the candidates considered here for the Trace dataset, this ideal SP appears to be the Shapelet Distance Map representation. In such a case, that ideal SP is the one that should be used to compute the RVI --- though such computations would at that point, of course, be redundant.

\subsubsection*{Matching Evaluation Scheme Issues}

It may now seem obvious that the evaluation SP should always match the clustering SP, else biases can result from differing identifications of cluster structure with fixed schemes.
However, it has not been established empirically or theoretically, nor is it immediately obvious, that the matching-SP evaluation scheme is any more suitable when seeking out an optimal SP among multiple candidates. We have identified three factors which contribute to its unsuitability. 

\begin{figure}[!ht]
    \centering
    \includegraphics[clip,width=0.9\columnwidth]{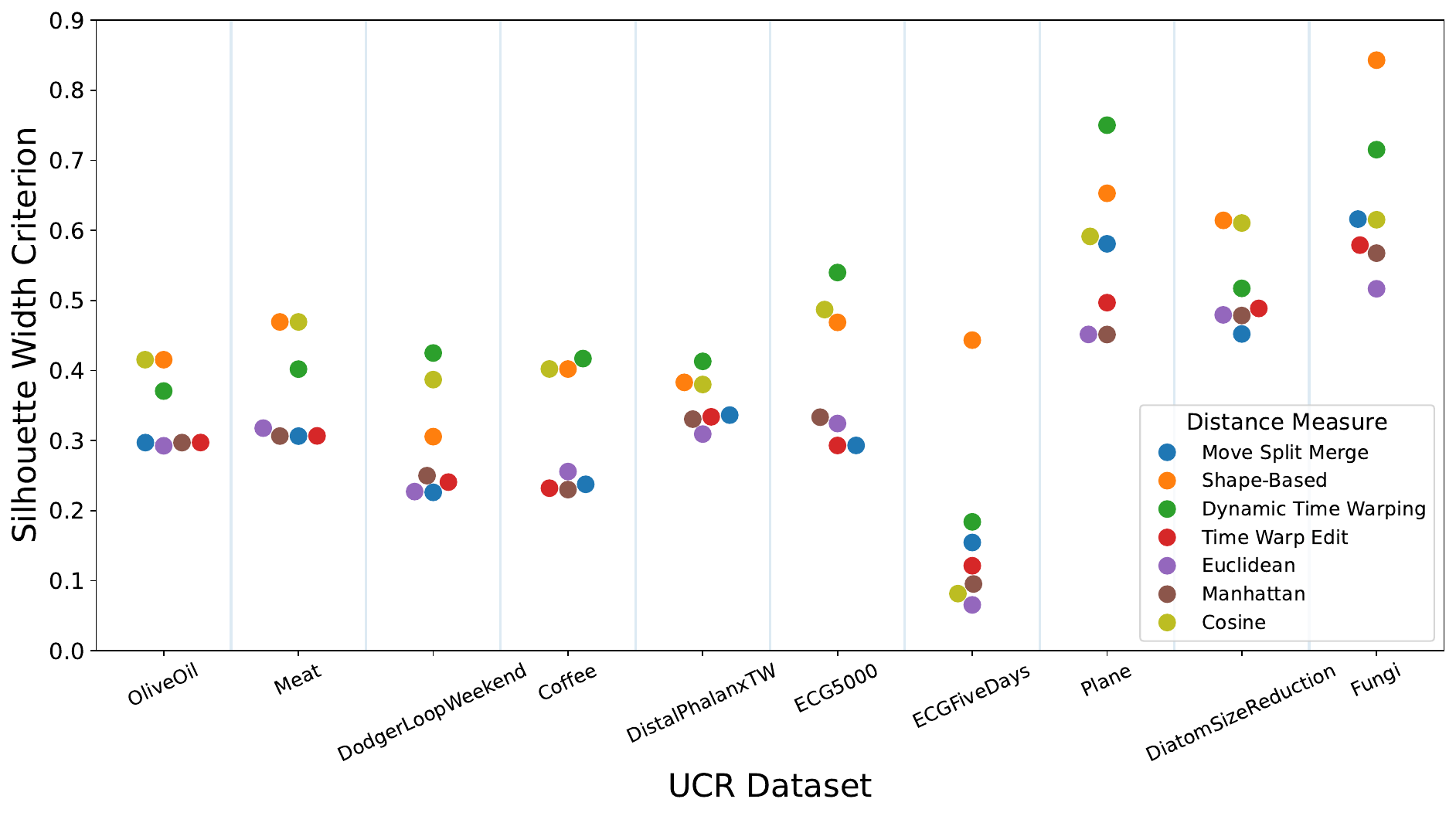}
    \caption{For each of ten selected datasets from the UCR archive \cite{Dau2019}, seven different versions of the SWC based on seven different SPs have been computed for the ground-truth partition from the archive. Note that for each SP, $z$-normalisation was applied and the raw data representation was used. The shape-based distance is the same employed in the $k$-shape clustering approach \cite{Paparrizos2015}. Dynamic Time Warping was computed with a $5\%$ Sakoe-Chiba band, Move-Split-Merge with cost parameter $1$, and Time Warp Edit with $\nu=0.05$ and $\lambda=1$.}
    \label{Fig:UCR_Different_PerfectPartition}
\end{figure}

Firstly, despite appearances, each coupling of RVI and SP actually produces a unique ``statistic" with its own distribution. As a result, RVIs equipped with different SPs will score the exact same partition differently. 
For instance, \Cref{Fig:UCR_Different_PerfectPartition} presents SWC values computed with a matching evaluation scheme for seven different SPs across ten datasets from the UCR Archive. If it is appropriate to perform SP-selection with
matching RVIs, the partition producing the maximum SWC would be indicative of the optimal SP for each dataset. 
However, for each of the datasets, it was only the ground truth partition that

\begin{figure}[!ht]
    \centering
    \includegraphics[width=0.8\textwidth]{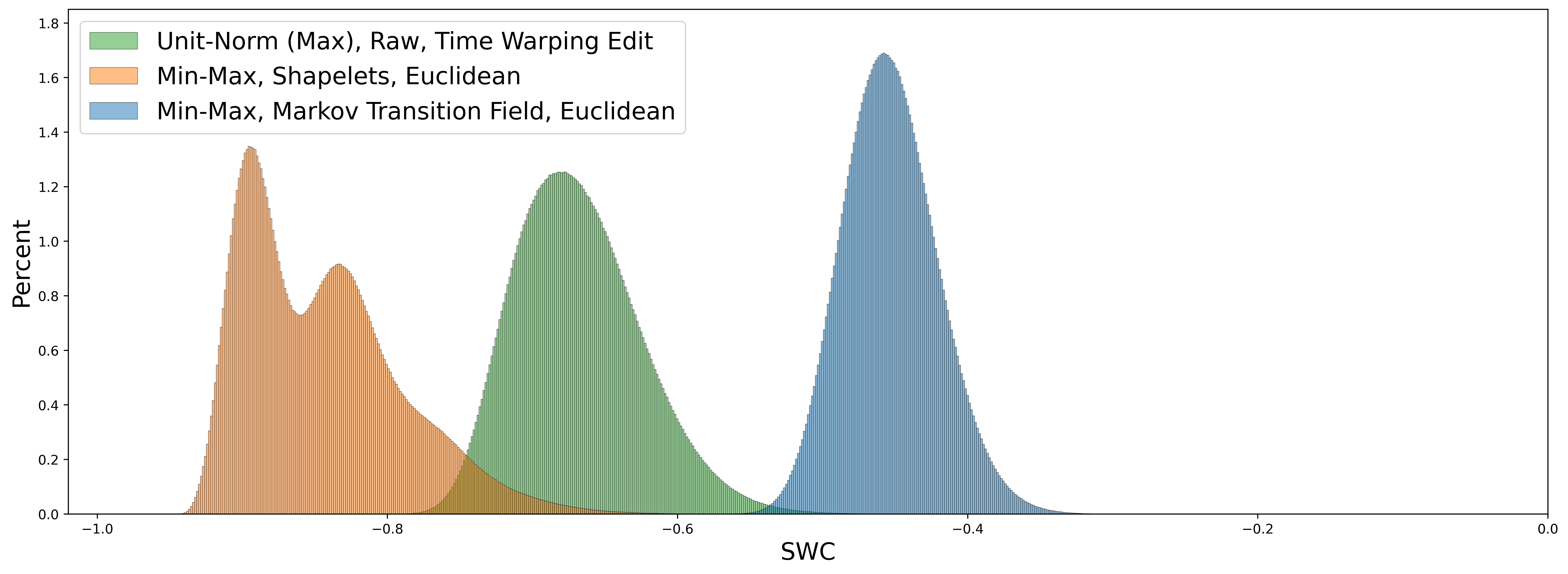}
    \caption{Empirical distributions of the SWC computed for the Trace dataset with three different SPs.}
    \label{Fig:SWC-Distributions}
\end{figure}

That each coupling of RVI and SP produces a unique statistic is further demonstrated in \Cref{Fig:SWC-Distributions}, where empirical distributions of the SWC have been computed for the Trace dataset with three different SPs from \Cref{Fig:Norms,Fig:Representations}. Each of these distributions vary across measures of centre, modality, spread and tail mass. For this dataset, the total number of possible non-trivial partitions is the Bell number $B_{200} - 2 \approx 6 \times 10^{275}$ \cite{e7661cb7-8363-3e1d-a4e2-6b1a1939c2c3,doi:10.1080/00029890.1964.11992270}, hence these distributions were obtained by uniformly sampling $3 \times 10^6$ such partitions according to the scheme presented in \cite{STAM1983231}. These clear distinctions between distributions are also maintained for uniformly random partitions with fixed numbers of clusters. 

Moreover, suppose we could feasibly evaluate all $B_{200}-2$ partitions using a fixed-SP version of an RVI, resulting in a ranking of the partitions (many of which are unlikely to be of interest to practitioners). This ranking can and will change when we embed the data in another SP and compute the alternative fixed-SP version of the RVI. 
This further supports the conclusion that RVIs based on different SPs are unfit for the sort of straightforward comparisons made when selecting between partitions with different numbers of clusters or produced by different clustering algorithms. This is because, in the latter case, the underlying ranking of partitions remains fixed under a single SP.

Secondly, when using RVIs for the task of SP-selection, practitioners are essentially trying to recover some theoretical, unobservable ranking of their candidate SPs. However, clustering algorithms are heuristics which search only a subspace of all $B_{N}$ partitions, and therefore they are liable to fall-short of the globally optimal partition, instead returning inferior local maxima or greedy solutions.
Hence there is no guarantee that the ranking of matching-SP RVI values computed for the purpose of SP-selection will coincide with the actual, unobserved ranking of the SPs anyway.

The final effect which might make the matching-SP scheme unreliable is the variable size of the bias effect discussed for fixed-SP schemes. If an RVI shows a larger bias towards some SPs than others, the ranking of SPs suggested by the matching-SP evaluation scheme becomes even more unreliable. 
This makes it difficult to know when comparing matching-SP RVIs if a particular partition is really better or worse, or if the bias conferred is more significant for one SP than another.

\paragraph{}\noindent
The following analogy can provide some intuition into the core issues facing both the fixed and matching evaluation schemes. Consider the problem of attempting to rank the performance of track and field athletes with different specialisations, or selecting the best performer amongst them. The athletes represent SPs, and their ranking within an event is analogous to the ranking of a corresponding partition according to an RVI. The fixed evaluation scheme is akin to ranking the athletes based on their performance in \textit{only} the 100m sprint. Clearly this scheme will be biased towards runners, particularly those athletes regularly competing in the 100m sprint. Meanwhile, the matching evaluation scheme is akin to ranking the athletes according to their performances within their individual events. But how can we compare 21.84 seconds in the 200m sprint to 22.85 meters in the shotput?

Whether a combinatorial scheme similar to that used in \cite{Iglesias2013} and \cite{Li2022} avoids the pitfalls described for the fixed and matching schemes also has not been established theoretically or experimentally. We have investigated this variety of evaluation scheme empirically by considering the arithmetic mean over all candidate fixed schemes, referred to as the mean-SP evaluation scheme.

	\section{Experimental Design} \label{Sec:Methodology-Main}

As previously discussed, this is the first study the authors are aware of that has attempted to empirically establish the suitability of RVIs for comparing the performance of different SPs for clustering. To this end we have conducted an extensive analysis over three different batteries of datasets. The experiments are used to address the following three questions introduced in \Cref{Sec:Introduction}: 
\begin{enumerate}
    \item[(i)] Do RVIs equipped with a fixed-SP evaluation scheme demonstrate an observable bias towards partitions generated from clustering approaches employing the same SP?
    \item[(ii)] Is there a difference in reliability between the fixed-SP, matching-SP and mean-SP evaluation schemes for either of the SP- or $k$-selection tasks?
    \item[(iii)] Are RVIs as reliable for SP-selection as they are for their conventional application of $k$-selection?
\end{enumerate}

\subsection{Methodology}
\label{Subsec:Methodology-Methodology}
The experimental methodology first introduced by Vendramin et al. \cite{Vendramin2010RelativeOverview} and employed in \cite{Jaskowiak2016OnCriteria,Vendramin2013OnCriteria,Rabbany2012RelativeAlgorithms,Moulavi2014Density-basedValidation} for the comparison of distinct relative validity criteria has been used alongside the complementary methodology, introduced by \cite{Gurrutxaga2011} and used in \cite{Arbelaitz2013AnIndices}, to establish insights into each of the three points above. The former involves analysing the \textit{correlations} between an RVI and an EVI amongst a subset of partitions. Previously this has been used to compare distinct RVIs, but in this paper it will be used to compare different fixed-SP and matching-SP versions of RVIs. It is proposed that superior versions of RVIs will more closely reflect the partition rankings established by an EVI, which is leveraging supervised information from labelled datasets. The latter methodology observes the rate at which the different versions of an RVI and an EVI are optimised for the same partition amongst a subset of partitions. As we are attempting to establish the suitability of RVIs for the task of SP-selection, we have chosen to compare their performance against the baseline task of $k$-selection. A similar level of performance on both tasks would be taken to suggest that, insofar as RVIs are dependable for $k$-selection, they can be relied upon for SP-selection.

The methodology introduced by Vendramin et al. was proposed to address several limitations identified in a prior approach, which gained prominence in the seminal work on RVI comparison by Milligan and Cooper \cite{Milligan1985}. Milligan and Cooper's methodology assumed that the accuracy of an RVI could be quantified by the frequency with which it suggested as the optimal partition, one which had the same number of clusters as the ground truth partition. An extensive examination of this success rate approach is presented in \cite{Vendramin2010RelativeOverview} where, amongst other critiques, it is recognised that unnatural partitions could be found by an RVI with the ``correct" number of clusters, whilst natural partitions may be found with the ``wrong" number of clusters. This criticism was independently echoed in \cite{Gurrutxaga2011} and referred to as the algorithm correctness assumption. The latter suggested that instead of counting the number of times an RVI recommends any partition with the same number of clusters as the ground-truth, one should instead count the number of times an RVI recommends the partition which also optimises an EVI. This improved success rate methodology was implemented in a comparative study \cite{Arbelaitz2013AnIndices}, and whilst an improvement on \cite{Milligan1985}, observing the \textit{coincidence of optima} similarly ignores all of the values of the RVIs for the remaining partitions. It was suggested in \cite{Vendramin2010RelativeOverview} that their novel methodology could be used to complement the success rate methodology. This is the view taken by the authors of the current study. Given that we are not comparing the relative performance of RVIs, but rather their suitability for different tasks, more angles for analysis of this suitability is beneficial. 

Conveniently, both of the above methodologies require the running of identical experiments. Namely, many partitions should be produced and evaluated by EVIs and by RVIs equipped with fixed-SP, matching-SP and mean-SP evaluation schemes. These partitions should be produced for an array of different datasets, with settings such as the clustering algorithm, number of clusters, and SP all varying. This same procedure also allows for an analysis of bias, allowing us to address each of the inquiries (i)-(iii) from above. The methodology for producing the partitions and their evaluation is described below:
\begin{enumerate}

    \item Take a battery of $N_{\delta}$ datasets with known ground-truth labels.

    \item Select an appropriate subset of $N_{\sigma}$ unique SPs for clustering the datasets. Various normalisations, representations, distances, or combinations thereof could be employed.

    \item Select a subset of $N_{\rho}$ RVIs, and one (or more) EVIs.
    
    \item For each combination of dataset and SP, vary the parameters of one or more clustering algorithms to produce $N_{\pi}$ partitions with assorted qualities and numbers of clusters. Note that we have used $N_{\alpha}$ clustering algorithms and $N_{\kappa}$ numbers of clusters, i.e. $N_{\pi} = N_{\alpha} \times N_{\kappa}$.

    \item For all of these $N_{\pi} \times N_{\sigma}$ partitions associated with one dataset, compute the EVI/s and use all $N_{\sigma}$ SPs to compute $N_{\sigma}$ versions of each RVI. \label{Item:Compute_RVIs}

    \item Subset the partitions such that all factors are static other than the number of clusters if assessing $k$-selection, or the SP if assessing SP-selection. Record the correlation and coincidence of optima with the EVI/s and: (i) the $N_{\sigma}$ versions of each RVI to assess them when using all $N_{\sigma}$ fixed-SP evaluation schemes; (ii) the average of the $N_{\sigma}$ versions of each RVI to assess them when using the mean-SP scheme, and; (iii) the version of each RVI where the SP matches that used in clustering to assess them when using the matching-SP evaluation scheme. Repeat for each relevant subset according to the task under assessment. \label{Item:Rpt_step_1}

    \item Repeat \cref{Item:Rpt_step_1} for each dataset. 
    
\end{enumerate}

\begin{table}[!ht]
    \small
    \centering
    \begin{tabular}{ccccccccccc} \toprule
        \multicolumn{3}{l}{} & \multicolumn{4}{c}{\footnotesize\textbf{Fixed-SP}} & \multicolumn{1}{l}{}\\ \cmidrule(l){4-7}
         \textbf{Partition} & \textbf{SP} & \textbf{ARI} & \textbf{SWC}\textsubscript{ED} & \textbf{SWC}\textsubscript{DTW} & \textbf{SWC}\textsubscript{SBD} & \textbf{SWC}\textsubscript{MSM} & \textbf{SWC}\textsubscript{Mean} & \textbf{SWC}\textsubscript{Match} & \textbf{OWM}\\ \midrule

        A & ED & 0.82 & \textbf{0.75} & 0.53 & 0.68 & 0.48 & 0.61 & \textbf{0.75} & 1 \\

        B & DTW & 0.74 & 0.64 & \textbf{0.61} & 0.69 & 0.47 & 0.60 & \textbf{0.61} & 0 \\

        C & SBD & 0.85 & 0.76 & 0.62 & \textbf{0.71} & 0.51 & 0.65 & \textbf{0.71} & 0 \\

        D & MSM & 0.79 & 0.69 & 0.65 & 0.70 & \textbf{0.53} & 0.64 & \textbf{0.53} & 0 \\ \midrule

        \textbf{CO} & \n & \n & 1 & 0 & 1 & 0 & 1 & 0 & \n \\
        
        \textbf{Corr.} & \n & \n & 1 & 0 & 0.4 & 0.4 & 0.9 & 0.6 & \n \\

        \bottomrule
    \end{tabular}
    \caption{An example to clarify the experimental methodology. ED, DTW, SBD and MSM represent different similarity paradigms, and have been used to produce each of the partitions A, B, C and D respectively for a static combination of dataset, clustering algorithm and number of clusters. Each partition has been scored according to an external index (ARI), 4 fixed-SP versions of the SWC (SWC\textsubscript{ED}, SWC\textsubscript{DTW}, SWC\textsubscript{SBD}, SWC\textsubscript{MSM}), the mean of the 4 fixed-SP versions of the SWC (SWC\textsubscript{Mean}) and the matching-SP version of the SWC (SWC\textsubscript{Match}). OWM is a boolean variable that is \texttt{True} when the SP used to generate the partition also produces the optimum fixed-SP version of the SWC. CO is another boolean variable which is \texttt{True} when the optimum value of SWC$_{x}$ coincides with the optimum value of the ARI. The final row of the table stores the correlation between SWC$_x$ and the ARI.}
    \label{Tab:MethodologyExample1}
\end{table}

For clarification of this procedure when assessing SP-selection, consider \Cref{Tab:MethodologyExample1} where four different SPs have been used to produce partitions for a a static combination of dataset, clustering algorithm and number of clusters. Note that each SP specifies a fixed-SP version of the RVI, which for this example is the SWC, and these are denoted by SWC\textsubscript{ED}, SWC\textsubscript{DTW}, SWC\textsubscript{SBD} and SWC\textsubscript{MSM}. The matching-SP version, SWC\textsubscript{Match} is constructed by taking the appropriate values (shown in bold) from the fixed-SP columns. The mean-SP version, SWC\textsubscript{Mean}, is constructed by taking the average of all of the fixed-SP columns. 

A boolean variable, Optimal When Matching (OWM), is \texttt{True} ($=1$) when the SP used to generate the partition also produces the optimum fixed-SP version of the SWC. This is used to assess the fixed-SP evaluation schemes for bias towards partitions produced with the same SP. If no such bias exists, then we could expect that OWM is a Bernoulli random variable with probability of success $1/N_{\sigma}$. A second boolean variable, Coincident Optima (CO), is \texttt{True} for a version of the RVI if its optimum over the partitions coincides with the optimum of the EVI. We use the term optimum as we considered a mixture of maximisation and minimisation criterion. Finally, the correlation can be computed between each version of the RVI and the EVI. A similar table could also be arranged to assess $k$-selection by allowing the number of clusters to vary for a static combination of dataset, clustering algorithm and SP. By computing both tables for all combinations of dataset, SP, clustering algorithm and numbers of clusters across three batteries of datasets, we can understand the performance of RVIs on the SP-selection task, relative to the $k$-selection task.

\subsection{Indices}
\label{Subsec:Methodology-Metrics}
Many EVIs have been proposed in the literature, typically classified into three categories according to the techniques they use to assess similarity between partitions. There are pair-counting, information theoretic and set-matching EVIs \cite{Warrens2022UnderstandingPairs,Arinik2021}. Most EVIs, including the ARI, rely on pair-counting techniques, which assess similarity by counting the number of object pairs that are either consistently grouped together or consistently placed in separate groups across the two partitions. Information theoretic EVIs, such as the Adjusted Mutual Information (AMI) \cite{Vinh2010}, treat partitions as categorical random variables and measure their similarity through mutual dependence between these variables. Set-matching EVIs, such as the F-measure or Purity \cite{Aggarwal2014}, attempt to find the best matches between clusters in the two partitions. A notable drawback of EVIs from this category is the exclusion of unpaired clusters and unmatched cluster portions from the similarity assessment \cite{Meila2007ComparingDistance}.

In comparing EVIs for clustering evaluation, \cite{Javed2020} found that ARI and AMI were particularly suitable due to their independence from the number of clusters and their adjustment for chance, with an expected value of zero for comparisons with random partitions. Milligan and Cooper \cite{Milligan1986AAnalysis} recommended ARI as the standard for external validation, a position supported by later research \cite{Milligan1996,Steinley2004PropertiesIndex,Javed2020}. Based on this evidence, we primarily report results using the ARI. We also computed the AMI, which produced consistent trends and observations, with corresponding results provided in the supplementary materials. Given the fundamentally different ways these EVIs compute similarity, this consistency suggests that our conclusions are robust to the choice of EVI.

The RVIs selected for our experiments were introduced in \Cref{Subsec:RVIs} and include the prototype-insensitive SWC, DI, CI, and AUCC. The prototype-sensitive CHI, DBI, and PBM are also used with the medoid prototype definition enforced for all similarity paradigms due to its general suitability. Where multiple medoids were obtained (such as for a cluster of size 2), one was selected at random. 
The correlation values were computed using Pearson's product-moment correlation coefficient. 
It should also be noted that for the DBI and CI, which are minimisation criteria, correlation analysis was performed using the negative of the calculated values. In the event that all of the partitions are identical, the ARI and RVI values will be constant and the correlation is undefined. Such instances have been excluded from further analyses. 

\subsection{Clustering Components}
\label{Subsec:Methodology-Clustering Algorithms}
The following experimental settings will be maintained across all three batteries, and settings unique to each battery will be detailed subsequently. All experiments have been performed using Python 3.9.4 on a linux-based HPC. 
As was demonstrated in \Cref{Sec:Whats_the_problem}, a unique SP can be produced by varying any one of three components of the clustering approach. However, for simplicity we will restrict our attention to varying only the distance measures. The distance measures used will be specified for each battery as they are introduced. 

The range of partitions will be produced by passing the corresponding pairwise distance matrices through the following nine clustering algorithms: BIRCH and spectral clustering with a Gaussian (RBF) kernel from \texttt{scikit-learn}, $k$-medoids from \texttt{scikit-learn-extra}, hierarchical agglomerative clustering with single, complete, average, weighted and wards linkages from \texttt{fastcluster}, and genieclust from \texttt{genieclust} \cite{Gagolewski2016}. The partition with optimal inertia using the PAM method was accepted from 30 initialisations of the $k$-medoids algorithm, with a maximum of 100 iterations. 
The genieclust threshold was set to 0.3 for all batteries.
The threshold parameter for BIRCH, the Gaussian kernel's width parameter $\delta$ for SC, and the range of $k$ (number of clusters) for the partitions will also be specified as each battery is introduced. We have opted to exclude the popular $k$-means algorithm and instead use the similar $k$-medoids algorithm due to concerns that the mean is not meaningful for some of the distance measures used in this study. 

\subsection{Datasets}
\label{Subsec:Methodology-Datasets}

\noindent \paragraph{} The experimental methodologies described in \Cref{Subsec:Methodology-Methodology} are not inert to criticisms, which have primarily been concerned with the adequacy of the ground-truth class labels. In \cite{Farber2010} it is noted that class labels do not always correspond to natural clusters, and that in such situations multiple data labellings may be required to capture all natural clusterings of a dataset.
This concern is shared in \cite{VanCraenendonck2015} where they observed that, whilst a high EVI score is easily interpreted, low scores are not necessarily indicative of low quality partitions. As will be discussed below, these concerns have been mitigated in the current study by selecting this combination of three batteries of datasets:
\begin{enumerate}
    \item[\Rom{1})] the \textit{Vendramin} et al. datasets \cite{Vendramin2010RelativeOverview};
    \item[\Rom{2})] \textit{Gagolewski's} \texttt{clustering-benchmarks} package datasets \cite{Gagolewski2022AAlgorithms};
    \item[\Rom{3})] the \textit{UCR} Time Series Classification archive datasets \cite{Dau2019}.
\end{enumerate}
These batteries and pertinent experimental settings have been summarised in \Cref{Tab:AllDatasetsBreakdown}.

\begin{table}[!ht]
    \small
    \begin{adjustbox}{center}
    \begin{tabular}{lccccc} \toprule
        \textbf{Battery} & \bm{$N_{\delta}$} & \bm{$N_{\sigma}$} & \bm{$N_{\alpha}$} & \bm{$N_{\kappa}$} & \bm{$\sum N_{\pi}$}\\ \midrule
        Vendramin & 972 & 6 & 9 & 49 & \num{2571912} \\
        Gagolewski & 57 & 6 & 9 & 21 & \num{64638} \\
        UCR & 112 & 7 & 9 & 11 & \num{77616} \\
        \bottomrule
    \end{tabular}
    \end{adjustbox}
    \caption{For each of the three batteries, this table details the number of: datasets in the battery ($N_{\delta}$), candidate SPs used for clustering ($N_{\sigma}$), clustering algorithms producing partitions ($N_{\alpha}$), unique numbers of clusters for those partitions ($N_{\kappa}$) and the total number of partitions produced for the battery ($N_{\delta} \times N_{\sigma} \times N_{\alpha} \times N_{\kappa}$).}
    \label{Tab:AllDatasetsBreakdown}
\end{table}

\subsubsection{The \textit{Vendramin} Battery}
\label{Subsubsec:Vendramin-Datasets}
Produced for \cite{Vendramin2010RelativeOverview} using the method introduced in Milligan and Cooper's seminal work \cite{Milligan1985}, the Vendramin battery consists of synthetic Euclidean datasets where the distribution of the objects within clusters follows a (mildly) truncated multi-variate normal distribution. This ensures that the datasets feature well-separated and compact Gaussian globular clusters which ``exhibit the properties of external isolation and internal cohesion". As various clustering algorithms will be used with varying parameters, including the number of clusters, we still obtained partitions spanning all levels of quality (see \Cref{Fig:ARI-Histograms-B}). Thus the relative simplicity of these datasets strikes a balance between complexity and ensuring that the issue described by \cite{Farber2010} is avoided almost entirely.

Each dataset features 500 objects embedded in $n$ dimensions, where $n \in \left \{2,3,4,22,23,24 \right \}$, and with $k^\star$ known clusters, where $k^\star \in \left \{2,4,6,12,14,16 \right \}$. The points were balanced between the clusters according to three different settings: a) approximately balanced, b) one cluster has $10\%$ of the objects, and the remainder are approximately evenly distributed amongst the remaining clusters, c) one cluster has $60\%$ of the objects for $k^\star \in \left \{2,4,6 \right \}$ or $20\%$ of the objects for $k^\star \in \left \{12,14,16 \right \}$, and the remainder are again approximately evenly distributed amongst the remaining clusters. The battery consists of a total of 972 datasets (6 dimensions $\times$ 6 numbers of clusters $\times$ 3 cluster balances $\times$ 9 replications).

For this battery we have used the raw data without applying any normalisation. The following set of distance measures were used to define the unique similarity paradigms: Euclidean Distance (ED), Manhattan Distance (MD), Cosine Distance (CoD), Canberra Distance (CaD), Braycurtis Distance (BD) and Chebyshev Distance (ChD). All of these distances are available from the \texttt{scipy} package. 
In accordance with performance findings in \cite{Gagolewski2021a} and preliminary experimentation, better clustering outputs were obtained when the BIRCH threshold parameter was set to $0.01$, and $\delta$ was chosen to be 10 for the SC Gaussian kernel. For the generated partitions we have emulated the larger range of $k$ from the original paper which featured in the presented results, i.e. $k \in \left \{ 2,3,\ldots,50 \right \}$.

\subsubsection{The \textit{Gagolewski} Battery}
\label{Subsubsec:Gagolewski-Datasets}

\begin{table}[!t]
    \footnotesize
    \begin{adjustbox}{center}
    \begin{tabular}{lrcccc|lrcccc} \toprule
        \textbf{Name} & \bm{$n$} & \bm{$d$} & \bm{$l$} & \bm{$k^\star$} & \bm{$k$} & \textbf{Name} & \bm{$n$} & \bm{$d$} & \bm{$l$} & \bm{$k^\star$} & \bm{$k$}\\ \midrule
        \textit{wut/circles} & 4000 & 2 & 1 & 4 & $[2,22]$            & \textit{sipu/a1} & 3000 & 2 & 1 & 20 & $[10,30]$\\
        \textit{wut/cross} & 2000 & 2 & 1 & 4 & $[2,22]$              & \textit{sipu/a2} & 5250 & 2 & 1 & 35 & $[25,45]$\\
        \textit{wut/graph} & 2500 & 2 & 1 & 10 & $[2,22]$             & \textit{sipu/a3} & 7500 & 2 & 1 & 50 & $[40,60]$\\
        \textit{wut/isolation} & 9000 & 2 & 1 & 3 & $[2,22]$          & \textit{sipu/aggregation} & 788 & 2 & 1 & 7 & $[2,22]$\\
        \textit{wut/labirynth} & 3546 & 2 & 1 & 6 & $[2,22]$          & \textit{sipu/compound} & 399 & 2 & 3 & 4,5,6 & $[2,22]$\\
        \textit{wut/mk1} & 300 & 2 & 1 & 3 & $[2,22]$                 & \textit{sipu/d31} & 3100 & 2 & 1 & 31 & $[21,41]$\\
        \textit{wut/mk2} & 1000 & 2 & 1 & 2 & $[2,22]$                & \textit{sipu/flame} & 240 & 2 & 1 & 2 & $[2,22]$\\
        \textit{wut/mk3} & 600 & 3 & 1 & 3 & $[2,22]$                 & \textit{sipu/jain} & 373 & 2 & 1 & 2 & $[2,22]$\\
        \textit{wut/mk4} & 1500 & 3 & 1 & 3 & $[2,22]$                & \textit{sipu/pathbased} & 300 & 2 & 2 & 3,4 & $[2,22]$\\
        \textit{wut/olympic} & 5000 & 2 & 1 & 5 & $[2,22]$            & \textit{sipu/r15} & 600 & 2 & 3 & 8,9,15 & $[2,22]$\\
        \textit{wut/smile} & 1000 & 2 & 2 & 4,6 & $[2,22]$            & \textit{sipu/s1} & 5000 & 2 & 1 & 15 & $[5,25]$\\
        \textit{wut/stripes} & 5000 & 2 & 1 & 2 & $[2,22]$            & \textit{sipu/s2} & 5000 & 2 & 1 & 15 & $[5,25]$\\
        \textit{wut/trajectories} & 10000 & 2 & 1 & 4 & $[2,22]$      & \textit{sipu/s3} & 5000 & 2 & 1 & 15 & $[5,25]$\\
        \textit{wut/trapped\_lovers} & 5000 & 3 & 1 & 3 & $[2,22]$    & \textit{sipu/s4} & 5000 & 2 & 1 & 15 & $[5,25]$\\
        \textit{wut/twosplashes} & 400 & 2 & 1 & 2 & $[2,22]$         & \textit{sipu/spiral} & 312 & 2 & 1 & 3 & $[2,22]$\\
        \textit{wut/windows} & 2977 & 2 & 1 & 5 & $[2,22]$            & \textit{sipu/unbalance} & 6500 & 2 & 1 & 8 & $[2,22]$\\
        \textit{wut/x1} & 120 & 2 & 1 & 3 & $[2,22]$                  & \textit{fcps/atom} & 800 & 3 & 1 & 2 & $[2,22]$\\
        \textit{wut/x2} & 120 & 2 & 1 & 4 & $[2,22]$                  & \textit{fcps/chainlink} & 1000 & 3 & 1 & 2 & $[2,22]$\\
        \textit{wut/x3} & 185 & 2 & 2 & 3,4 & $[2,22]$                & \textit{fcps/engytime} & 1000 & 3 & 2 & 2 & $[2,22]$\\
        \textit{wut/z1} & 192 & 2 & 1 & 3 & $[2,22]$                  & \textit{fcps/hepta} & 212 & 3 & 1 & 7 & $[2,22]$\\
        \textit{wut/z2} & 900 & 2 & 1 & 5 & $[2,22]$                  & \textit{fcps/lsun} & 400 & 2 & 1 & 3 & $[2,22]$\\
        \textit{wut/z3} & 1000 & 2 & 1 & 4 & $[2,22]$                 & \textit{fcps/target} & 770 & 2 & 1 & 6 & $[2,22]$\\
        \textit{graves/dense} & 200 & 2 & 1 & 2 & $[2,22]$            & \textit{fcps/tetra} & 400 & 3 & 1 & 4 & $[2,22]$\\
        \textit{graves/fuzzyx} & 1000 & 2 & 1 & 5 & $[2,22]$          & \textit{fcps/twodiamonds} & 800 & 2 & 1 & 2 & $[2,22]$\\
        \textit{graves/line} & 250 & 2 & 1 & 2 & $[2,22]$             & \textit{fcps/wingnut} & 1016 & 2 & 1 & 2 & $[2,22]$\\
        \textit{graves/parabolic} & 1000 & 2 & 2 & 2,4 & $[2,22]$     & \textit{other/iris} & 150 & 4 & 1 & 3 & $[2,22]$\\
        \textit{graves/ring} & 1000 & 2 & 1 & 2 & $[2,22]$            & \textit{other/iris5} & 105 & 4 & 1 & 3 & $[2,22]$\\
        \textit{graves/ring\_outliers} & 1030 & 2 & 1 & 5 & $[2,22]$  & \textit{other/square} & 1000 & 2 & 1 & 2 & $[2,22]$\\
        \textit{graves/zigzag} & 250 & 2 & 2 & 3,5 & $[2,22]$         &\\
        \bottomrule
    \end{tabular}
    \end{adjustbox}
    \caption{Details of the 57 \texttt{clustering-benchmarks} datasets. For each dataset, the table specifies the number of points ($n$), the number of features ($d$), the number of unique ground-truth labellings ($l$) and the associated numbers of ground-truth clusters ($k^\star$), and the range of $k$ values used to produce partitions.}
    \label{Tab:Gagolewski_Datasets}
\end{table}

The \texttt{clustering-benchmarks} package provides convenient access to a range of ``clustering benchmark collections referred to across the machine learning and data mining literature." Most of these collections of datasets are designed to challenge traditional clustering methods, being composed mainly of points embedded in Euclidean space with clusters of arbitrary shapes, varying densities or varying degrees of overlap. Another key characteristic differentiating this battery is the provision of multiple alternative labellings for a subset of the datasets. The EVI recorded for a clustering of a multiply labelled dataset is the maximum EVI computed with \textit{any} of the ground-truth partitions. This feature of the battery also addresses the concerns of \cite{Farber2010}.

We will use datasets from the five recommended collections (\textit{wut}, \textit{sipu}, \textit{fcps}, \textit{graves} and \textit{other}) in one aggregated battery. The individual datasets will be referred to using the format \textit{collection/name}.
Some of the reference labels for a selection of the datasets include a label allocating noise. These noisy labellings will not be used in our study, which means that 8 datasets with exclusively noisy labellings will be excluded. We have also omitted \textit{sipu/birch1}, \textit{sipu/birch2}, \textit{sipu/worms\_2} and \textit{sipu/worms\_64} as they are significantly larger than the other datasets in this battery (at least 10$\times$ more objects than the next largest dataset) and would create further computational hurdles. This leaves us with 57 datasets, of which 8 have two or more reference labellings. Pertinent details of these datasets are presented in \Cref{Tab:Gagolewski_Datasets}, including the inclusive range of integer $k$ that partitions were produced for, which was a window with 21 values centered around the ground truth numbers of clusters. The lower value was computed as $\max \left\{2, \left \lceil \left(\min{k^\star}+\max{k^\star}\right )/2\right \rceil-10 \right\}$, and the upper value as $\max \left\{22, \left \lceil \left (\min{k^\star}+\max{k^\star}\right )/2\right \rceil+10 \right\}$ for each dataset, where $\lceil \cdot \rceil$ denotes the ceiling function. 

As for the Vendramin battery, the raw data have been used without any normalisation, and the same set of distance measures have been used to define the unique similarity paradigms. 
The BIRCH threshold parameter was again set to $0.01$, and $\delta$ was chosen to be 10 for the SC Gaussian kernel. It should be noted that the combination of complete linkage hierarchical clustering algorithm and Canberra distance sometimes failed to produce partitions with the requested number of clusters on the lower end of the range. This was due to the last agglomeration in the dendrogram collecting $k=3,4,5$ or $6$ clusters into one. For this combination, 85/1120 of the partitions were trivial (all elements in one cluster) across 25/57 datasets. We have carried out analyses on the complete cases.

\subsubsection{The \textit{UCR} Archive Battery}
\label{Subsubsec:UCR-Datasets}

\begin{table}[!hbt]
    \footnotesize
    \begin{adjustbox}{center}
    \begin{tabular}{lrrrc|lrrrc} \toprule
        \textbf{Name} & \bm{$n$} & \bm{$d$} & \bm{$k^\star$} & \bm{$k$} & \textbf{Name} & \bm{$n$} & \bm{$d$} & \bm{$k^\star$} & \bm{$k$}\\ \midrule
        \textit{ACSF1} & 200 & 1460 & 10 & $[5,15]$                                      & \textit{Mallat} & 2400 & 1024 & 8 & $[3,13]$\\ 
        \textit{Adiac} & 781 & 176 & 37 & $[32,42]$                                      & \textit{Meat} & 120 & 448 & 3 & $[2,12]$\\ 
        \textit{ArrowHead} & 211 & 251 & 3 & $[2,12]$                                    & \textit{MedicalImages} & 1141 & 99 & 10 & $[5,15]$\\
        \textit{BME} & 180 & 128 & 3 & $[2,12]$                                          & \textit{MelbournePedestrian} & 3633 & 24 & 10 & $[5,15]$\\
        \textit{Beef} & 60 & 470 & 5 & $[2,12]$                                          & \textit{MiddlePhalanxOutlineAgeGroup} & 554 & 80 & 3 & $[2,12]$\\
        \textit{BeetleFly} & 40 & 512 & 2 & $[2,12]$                                     & \textit{MiddlePhalanxOutlineCorrect} & 891 & 80 & 2 & $[2,12]$\\ 
        \textit{BirdChicken} & 40 & 512 & 2 & $[2,12]$                                   & \textit{MiddlePhalanxTW} & 553 & 80 & 6 & $[2,12]$\\
        \textit{CBF} & 930 & 128 & 3 & $[2,12]$                                          & \textit{MixedShapesRegularTrain} & 2925 & 1024 & 5 & $[2,12]$\\ 
        \textit{Car} & 120 & 577 & 4 & $[2,12]$                                          & \textit{MixedShapesSmallTrain} & 2525 & 1024 & 5 & $[2,12]$\\ 
        \textit{Chinatown} & 363 & 24 & 2 & $[2,12]$                                     & \textit{MoteStrain} & 1272 & 84 & 2 & $[2,12]$\\
        \textit{ChlorineConcentration} & 4307 & 166 & 3 & $[2,12]$                       & \textit{NonInvasiveFetalECGThorax1} & 3765 & 750 & 42 & $[37,47]$\\
        \textit{CinCECGTorso} & 1420 & 1639 & 4 & $[2,12]$                               & \textit{NonInvasiveFetalECGThorax2} & 3765 & 750 & 42 & $[37,47]$\\
        \textit{Coffee} & 56 & 286 & 2 & $[2,12]$                                        & \textit{OSULeaf} & 442 & 427 & 6 & $[2,12]$\\
        \textit{Computers} & 500 & 720 & 2 & $[2,12]$                                    & \textit{OliveOil} & 60 & 570 & 4 & $[2,12]$\\
        \textit{CricketX} & 780 & 300 & 12 & $[7,17]$                                    & \textit{PhalangesOutlinesCorrect} & 2658 & 80 & 2 & $[2,12]$\\
        \textit{CricketY} & 780 & 300 & 12 & $[7,17]$                                    & \textit{Phoneme} & 2110 & 1024 & 39 & $[34,44]$\\
        \textit{CricketZ} & 780 & 300 & 12 & $[7,17]$                                    & \textit{PigAirwayPressure} & 312 & 2000 & 52 & $[47,57]$\\
        \textit{Crop} & 24000 & 46 & 24 & $[19,29]$                                      & \textit{PigArtPressure} & 312 & 2000 & 52 & $[47,57]$\\
        \textit{DiatomSizeReduction} & 322 & 345 & 4 & $[2,12]$                          & \textit{PigCVP} & 312 & 2000 & 52 & $[47,57]$\\
        \textit{DistalPhalanxOutlineAgeGroup} & 539 & 80 & 3 & $[2,12]$                  & \textit{Plane} & 210 & 144 & 7 & $[2,12]$\\
        \textit{DistalPhalanxOutlineCorrect} & 876 & 80 & 2 & $[2,12]$                   & \textit{PowerCons} & 360 & 144 & 2 & $[2,12]$\\
        \textit{DistalPhalanxTW} & 539 & 80 & 6 & $[2,12]$                               & \textit{ProximalPhalanxOutlineAgeGroup} & 605 & 80 & 3 & $[2,12]$\\
        \textit{DodgerLoopDay} & 158 & 288 & 7 & $[2,12]$                                & \textit{ProximalPhalanxOutlineCorrect} & 891 & 80 & 2 & $[2,12]$\\
        \textit{DodgerLoopGame} & 158 & 288 & 2 & $[2,12]$                               & \textit{ProximalPhalanxTW} & 605 & 80 & 6 & $[2,12]$\\
        \textit{DodgerLoopWeekend} & 158 & 288 & 2 & $[2,12]$                            & \textit{RefrigerationDevices} & 750 & 720 & 3 & $[2,12]$\\
        \textit{ECG5000} & 5000 & 140 & 5 & $[2,12]$                                     & \textit{Rock} & 70 & 2844 & 4 & $[2,12]$\\
        \textit{ECGFiveDays} & 884 & 136 & 2 & $[2,12]$                                  & \textit{ScreenType} & 750 & 720 & 3 & $[2,12]$\\
        \textit{EOGHorizontalSignal} & 724 & 1250 & 12 & $[7,17]$                        & \textit{SemgHandGenderCh2} & 900 & 1500 & 2 & $[2,12]$\\
        \textit{EOGVerticalSignal} & 724 & 1250 & 12 & $[7,17]$                          & \textit{SemgHandMovementCh2} & 900 & 1500 & 6 & $[2,12]$\\
        \textit{Earthquakes} & 461 & 512 & 2 & $[2,12]$                                  & \textit{SemgHandSubjectCh2} & 900 & 1500 & 5 & $[2,12]$\\
        \textit{ElectricDevices} & 16637 & 96 & 7 & $[2,12]$                             & \textit{ShapeletSim} & 200 & 500 & 2 & $[2,12]$\\
        \textit{EthanolLevel} & 1004 & 1751 & 4 & $[2,12]$                               & \textit{ShapesAll} & 1200 & 512 & 60 & $[55,65]$\\
        \textit{FaceAll} & 2250 & 131 & 14 & $[9,19]$                                    & \textit{SmallKitchenAppliances} & 750 & 720 & 3 & $[2,12]$\\
        \textit{FaceFour} & 112 & 350 & 4 & $[2,12]$                                     & \textit{SmoothSubspace} & 300 & 15 & 3 & $[2,12]$\\
        \textit{FacesUCR} & 2250 & 131 & 14 & $[9,19]$                                   & \textit{SonyAIBORobotSurface1} & 621 & 70 & 2 & $[2,12]$\\
        \textit{FiftyWords} & 905 & 270 & 50 & $[45,55]$                                 & \textit{SonyAIBORobotSurface2} & 980 & 65 & 2 & $[2,12]$\\
        \textit{Fish} & 350 & 463 & 7 & $[2,12]$                                         & \textit{StarLightCurves} & 9236 & 1024 & 3 & $[2,12]$\\
        \textit{FreezerRegularTrain} & 3000 & 301 & 2 & $[2,12]$                         & \textit{Strawberry} & 983 & 235 & 2 & $[2,12]$\\
        \textit{FreezerSmallTrain} & 2878 & 301 & 2 & $[2,12]$                           & \textit{SwedishLeaf} & 1125 & 128 & 15 & $[10,20]$\\
        \textit{Fungi} & 204 & 201 & 18 & $[13,23]$                                      & \textit{Symbols} & 1020 & 398 & 6 & $[2,12]$\\
        \textit{GunPoint} & 200 & 150 & 2 & $[2,12]$                                     & \textit{SyntheticControl} & 600 & 60 & 6 & $[2,12]$\\
        \textit{GunPointAgeSpan} & 451 & 150 & 2 & $[2,12]$                              & \textit{ToeSegmentation1} & 268 & 277 & 2 & $[2,12]$\\
        \textit{GunPointMaleVersusFemale} & 451 & 150 & 2 & $[2,12]$                     & \textit{ToeSegmentation2} & 166 & 343 & 2 & $[2,12]$\\
        \textit{GunPointOldVersusYoung} & 451 & 150 & 2 & $[2,12]$                       & \textit{Trace} & 200 & 275 & 4 & $[2,12]$\\
        \textit{Ham} & 214 & 431 & 2 & $[2,12]$                                          & \textit{TwoLeadECG} & 1162 & 82 & 2 & $[2,12]$\\
        \textit{HandOutlines} & 1370 & 2709 & 2 & $[2,12]$                               & \textit{TwoPatterns} & 5000 & 128 & 4 & $[2,12]$\\
        \textit{Haptics} & 463 & 1092 & 5 & $[2,12]$                                     & \textit{UMD} & 180 & 150 & 3 & $[2,12]$\\
        \textit{Herring} & 128 & 512 & 2 & $[2,12]$                                      & \textit{UWaveGestureLibraryAll} & 4478 & 945 & 8 & $[3,13]$\\
        \textit{HouseTwenty} & 159 & 2000 & 2 & $[2,12]$                                 & \textit{UWaveGestureLibraryX} & 4478 & 315 & 8 & $[3,13]$\\
        \textit{InlineSkate} & 650 & 1882 & 7 & $[2,12]$                                 & \textit{UWaveGestureLibraryY} & 4478 & 315 & 8 & $[3,13]$\\
        \textit{InsectEPGRegularTrain} & 311 & 601 & 3 & $[2,12]$                        & \textit{UWaveGestureLibraryZ} & 4478 & 315 & 8 & $[3,13]$\\
        \textit{InsectEPGSmallTrain} & 266 & 601 & 3 & $[2,12]$                          & \textit{Wine} & 111 & 234 & 2 & $[2,12]$\\
        \textit{InsectWingbeatSound} & 2200 & 256 & 11 & $[6,16]$                        & \textit{WordSynonyms} & 905 & 270 & 25 & $[20,30]$\\
        \textit{ItalyPowerDemand} & 1096 & 24 & 2 & $[2,12]$                             & \textit{Worms} & 258 & 900 & 5 & $[2,12]$\\
        \textit{LargeKitchenAppliances} & 750 & 720 & 3 & $[2,12]$                       & \textit{WormsTwoClass} & 258 & 900 & 2 & $[2,12]$\\
        \textit{Lightning7} & 143 & 319 & 7 & $[2,12]$                                   & \textit{Yoga} & 3300 & 426 & 2 & $[2,12]$\\
    \bottomrule
    \end{tabular}
    \end{adjustbox}
    \caption{Details of the 112 UCR classification archive datasets. For each dataset, the table specifies the number of points ($n$), the number of features ($d$), the number of ground-truth clusters ($k^\star$), and the range of $k$ values used to produce partitions.}
    \label{Tab:UCR_Datasets}
\end{table}

The UCR archive is the most extensive battery of publicly available labelled time series, providing a good variety of time series datasets where many different SPs can be optimal. Whilst the archive provides neither of the class label guarantees enjoyed by the other two batteries, it could be argued that non-standard SPs are more common for complex non-Euclidean datasets. Thus it is more relevant to perform these experiments on such datasets, and so this battery is used to complement the results from the other two. This battery is certainly a suitable candidate to be developed into a proper clustering benchmark battery with multiple alternative labellings, similar to the Gagolewski battery. We have used a subset of 112 datasets of the 128 supplied in the archive, and these are detailed in \Cref{Tab:UCR_Datasets}. This subset is the same that was employed in a time series clustering benchmark study \cite{Javed2020}, where datasets of variable lengths or with missing values were also excluded. Being a classification archive, a default train-test split is supplied for each dataset, and these have been combined for our purposes.

The first 85 datasets forming the archive were contributed with $z$-normalisation already applied, whilst some of the 43 newer datasets have undergone no normalisation. Only two of the datasets we used in this study were supplied to the archive without normalisation, thus for the sake of consistency, we have normalised all of the datasets using $z$-normalisation. These datasets are larger on average again than in the Gagolewski battery, hence we have considered a narrower window with 11 values centered around the ground truth number of clusters. The lower value was computed as $\max \left\{2, k^\star-5 \right\}$, and the upper value as $\max \left\{12, k^\star+5 \right\}$ for each dataset. 

The following set of distance measures were used to define the unique similarity paradigms: Euclidean Distance (ED), Manhattan distance (MD), Cosine Distance (CD), Shape-Based Distance (SBD) from the k-shape clustering approach, Dynamic Time Warping (DTW) with a $5\%$ Sakoe-Chiba band, Move-Split-Merge (MSM) with cost $c=1$, and Time Warping Edit Distance (TWED) with $\nu=0.05$ and $\lambda=1$. We used the \texttt{aeon} implementations for DTW, MSM and TWED, and the \texttt{kshape} implementation of SBD. The BIRCH threshold parameter was set to 0.1, and reduced by factors of 10 until the requested number of clusters was returned by the algorithm. Similarly, $\delta$ was set to 20 and incremented by 20 until the requested number of clusters was obtained. 

\section{Experimental Results}
\label{Sec:Experimental_Results}

\subsection{Analysis of Bias Towards Matching Similarity Paradigms}
\label{Subsec:Experimental_Results-Bias}

\begin{figure}[!ht]
    \centering
    \begin{subfigure}[t]{\textwidth}
        \centering
        \includegraphics[width=\textwidth]{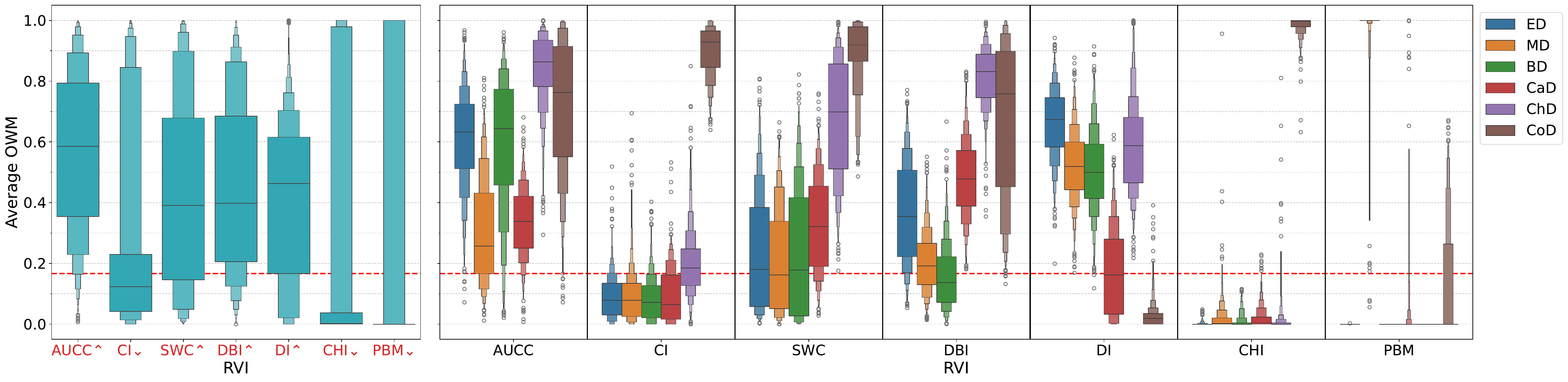}
        \caption{Vendramin}
        \label{Fig:BiasPlots-Vendramin}
    \end{subfigure}
    ~
    \begin{subfigure}[t]{\textwidth}
        \centering
        \includegraphics[width=\textwidth]{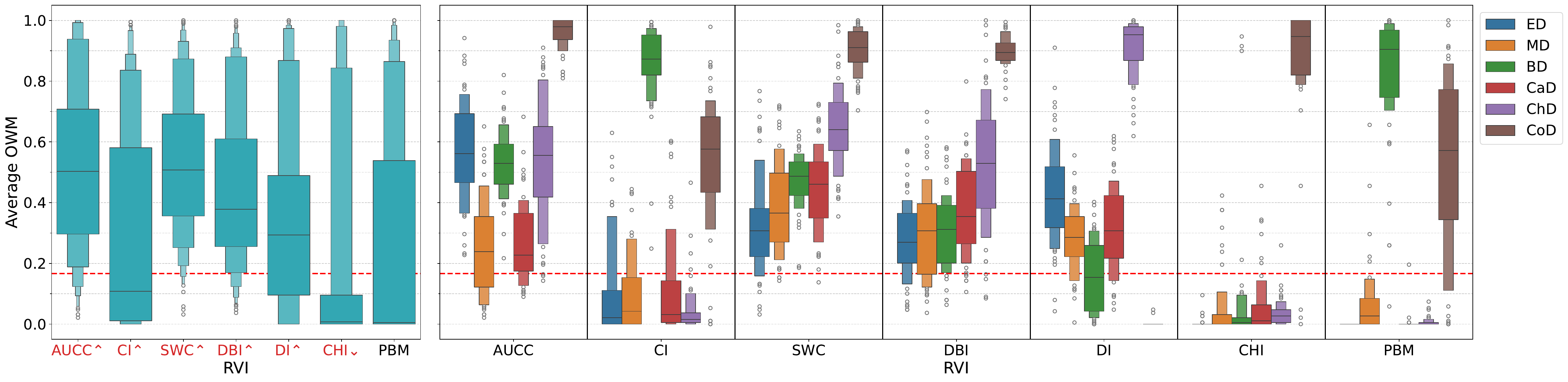}
        \caption{Gagolewski}
        \label{Fig:BiasPlots-Gagolewski}
    \end{subfigure}
    ~
    \begin{subfigure}[t]{\textwidth}
        \centering
        \includegraphics[width=\textwidth]{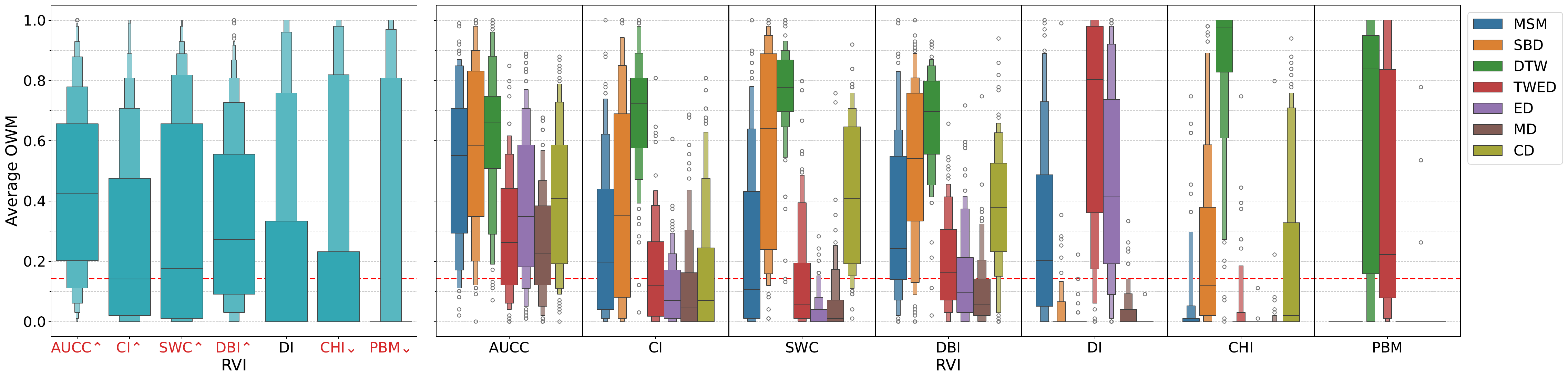}
        \caption{UCR}
        \label{Fig:BiasPlots-UCR}
    \end{subfigure}
    \caption{Enhanced boxplots of the \textit{distributions} of average OWM values for each battery grouped by RVI on the left, and further by SP on the right. Each point in both groupings represents the average OWM for partitions clustered using one SP on a single dataset. Averaging was performed over all of the corresponding partitions produced by the different clustering algorithms, with different numbers of clusters. The expected value ($1/N_{\sigma}$) in the absence of any bias is shown by the red horizontal dashed lines.}
    \label{Fig:BiasPlots}
\end{figure}

In \Cref{Sec:Whats_the_problem} it was suggested that RVIs may demonstrate a bias towards partitions produced using the same similarity paradigm they are computed with. As previously illustrated in \Cref{Tab:MethodologyExample1} with a toy example, a boolean variable, OWM, recorded whether the fixed-SP version of an RVI is optimal when it matches the clustering SP. The average of these booleans across all partitions of a single dataset from different clustering algorithms and numbers of clusters is used to indicate the bias for one combination of dataset, RVI and SP within each of the three batteries. If an RVI does not show a bias for the matching (or a particular) SP, then these OWM averages
should be symmetrically distributed about $1/N_{\sigma}$, i.e. $1/6$ for the Vendramin and Gagolewski batteries, and $1/7$ for the UCR battery. The distributions of average OWM values over the datasets from each battery are presented in \Cref{Fig:BiasPlots} as enhanced box-plots (boxen-plots), or ``letter value" plots \cite{letter-value-plot}. The distributions are grouped by RVI on the left, and further broken down by both RVI and SP on the right. The expected value in the absence of any bias is shown by the red horizontal dashed lines. 

Firstly we shall consider the distributions grouped by RVI. Interestingly, the distributions for each RVI are for the most part consistent across the three batteries in terms of both location and variation.
AUCC, SWC and DBI clearly demonstrated this theorised bias across all three batteries, with the DI showing similar bias for the first two batteries. Whilst not showing a positive bias, the CHI and PBM indices were still not symmetrically distributed around the expected value, suggesting that certain SPs were consistently producing the optimal version of these RVIs. 
The labels on the horizontal axis have been used to indicate the outcome of appropriately directed one-sided Wilcoxon signed rank tests at a $99.9\%$ confidence level. A red label indicates a significant result, followed by the direction, whilst a black label suggests failure to reject. These indicate that the medians for AUCC, SWC and DBI were statistically significantly greater than $1/N_{\sigma}$ for all three batteries, while the median for CHI was also found to be lower with statistical significance for all three batteries. The median for DI was greater with statistical significance for the Vendramin and Gagolewski batteries, and the median for PBM was lower for Vendramin and UCR. The CI was the only index with significant outcomes in both directions. 
The use of hierarchical clustering algorithms suggests some caution should be used when interpreting the output of any statistical test applied to the data from our experiments, as it is questionable whether the independence assumption is valid. This is due to the high degree of similarity between partitions with adjacent numbers of clusters in a hierarchical clustering solution. We have attempted to mitigate this by employing a range of linkages and non-hierarchical algorithms and using a high confidence level.

Now we shall discuss the distributions grouped by RVI and SP in the right-hand plots from \Cref{Fig:BiasPlots}. The extent of the bias appears to be quite extreme for particular combinations of RVI and SP. Consider the PBM index for the Vendramin battery (rightmost cell of \Cref{Fig:BiasPlots-Vendramin}), where PBM\textsubscript{ChD} was never optimal for ChD clusterings, whilst PBM\textsubscript{MD} was optimal for most of the MD clusterings. It is thus likely that PBM\textsubscript{MD} was overwhelmingly the optimal version of the PBM index for this battery, regardless of the clustering SP. A similar observation can be made for CI and CHI in \Cref{Fig:BiasPlots-Vendramin}, where the CoD version was frequently optimal for CoD clusterings. 

The fixed-CoD evaluation scheme demonstrates one of the largest and most consistent bias effects across the majority of the RVIs for the Vendramin and Gagolewski batteries. CoD is arguably the most unique SP amongst those considered for these batteries. In contrast, there is marked similarity between the ED, MD and BD SPs. These similarities and differences will be borne out in the partitions produced by these SPs. As a result, the CoD-versions of the RVIs are unlikely to recommend partitions which do not align with the CoD SP, i.e. partitions produced by any of the other five SPs. This is an issue if a practitioner is using such a fixed-SP evaluation scheme to select between SPs which are highly unique from one another (as would likely be the case if selecting an optimal SP), as this clearly amplifies the bias effect. Less bias is likely to be present if there is a high degree of similarity between all of the SPs, but one could argue that consequently there is significantly less value in comparing them to select the ``best" performer, which may at that point equate to a random decision.
Whilst some of the RVIs primarily experience this bias effect for a small subset of the more unique SPs (consider CI and CHI across all three batteries), others such as SWC, DBI and DI demonstrate this bias effect regardless of the SP. AUCC in particular appears to be the most susceptible to this. It is the only index without a median below the expected value for at least one SP on all three batteries. 

\subsection{Coincidence of EVI and RVI Optima}
\label{Subsec:Experimental_Results-CoincidenceRates}

\begin{figure}[!htb]
\centering
    \begin{subfigure}[t]{0.49\textwidth}
        \centering
        \includegraphics[width=\textwidth]{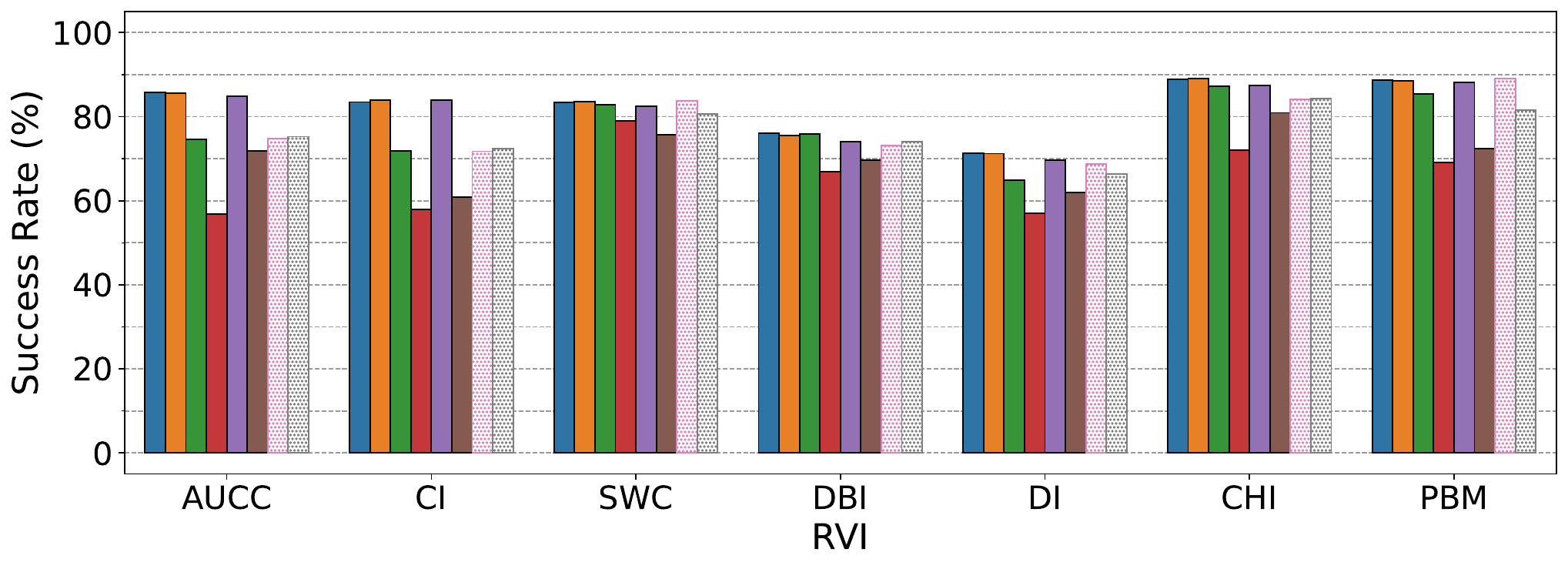} 
        \caption{Vendramin (over $k$)}
        \label{Fig:CoincidencePlots-Vend-K}
    \end{subfigure}%
    ~
    \begin{subfigure}[t]{0.49\textwidth}
        \centering
        \includegraphics[width=\textwidth]{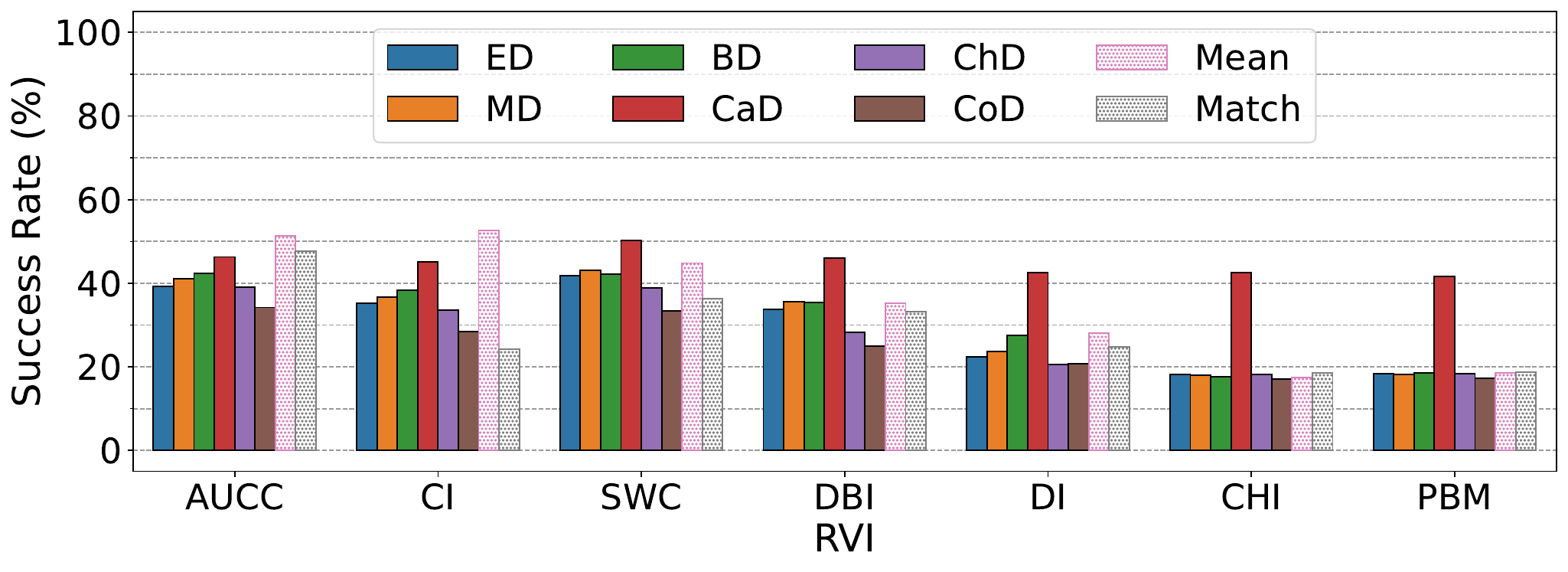}
        \caption{Vendramin (over SPs)}
        \label{Fig:CoincidencePlots-Vend-SP}
    \end{subfigure}
    ~
    \begin{subfigure}[t]{0.49\textwidth}
        \centering
        \includegraphics[width=\textwidth]{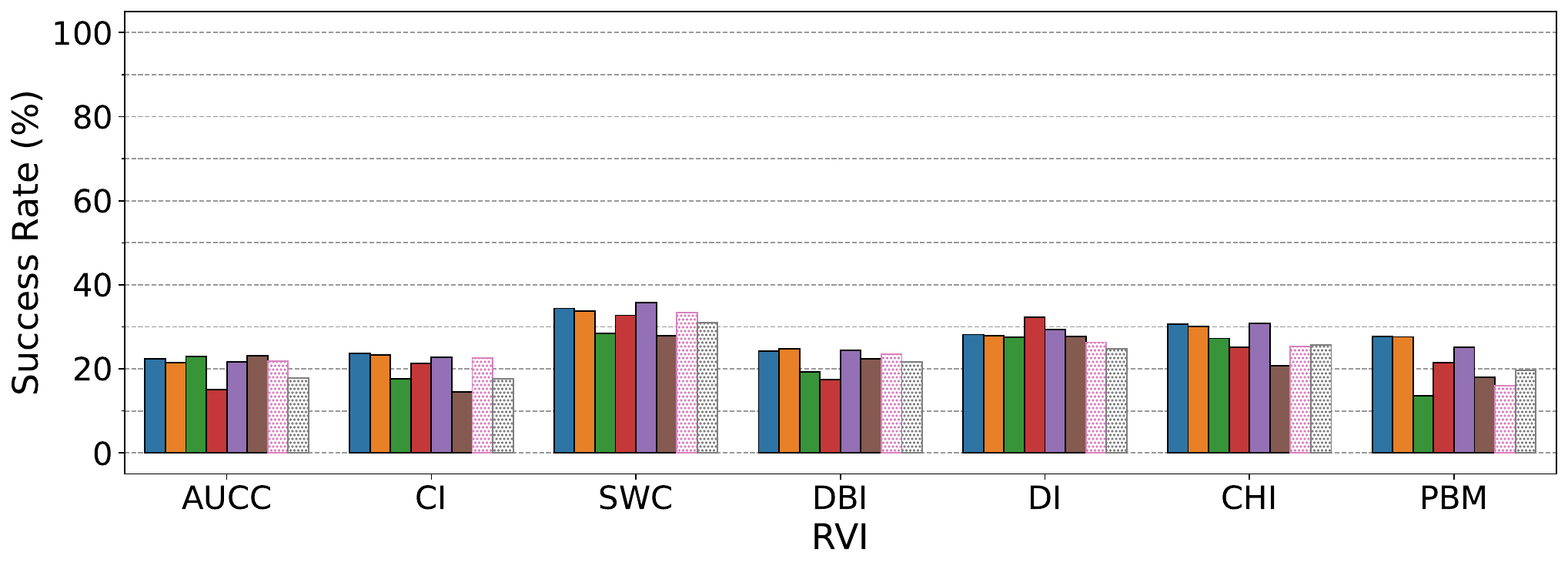} 
        \caption{Gagolewski (over $k$)}
        \label{Fig:CoincidencePlots-Gag-K}
    \end{subfigure}%
    ~
    \begin{subfigure}[t]{0.49\textwidth}
        \centering
        \includegraphics[width=\textwidth]{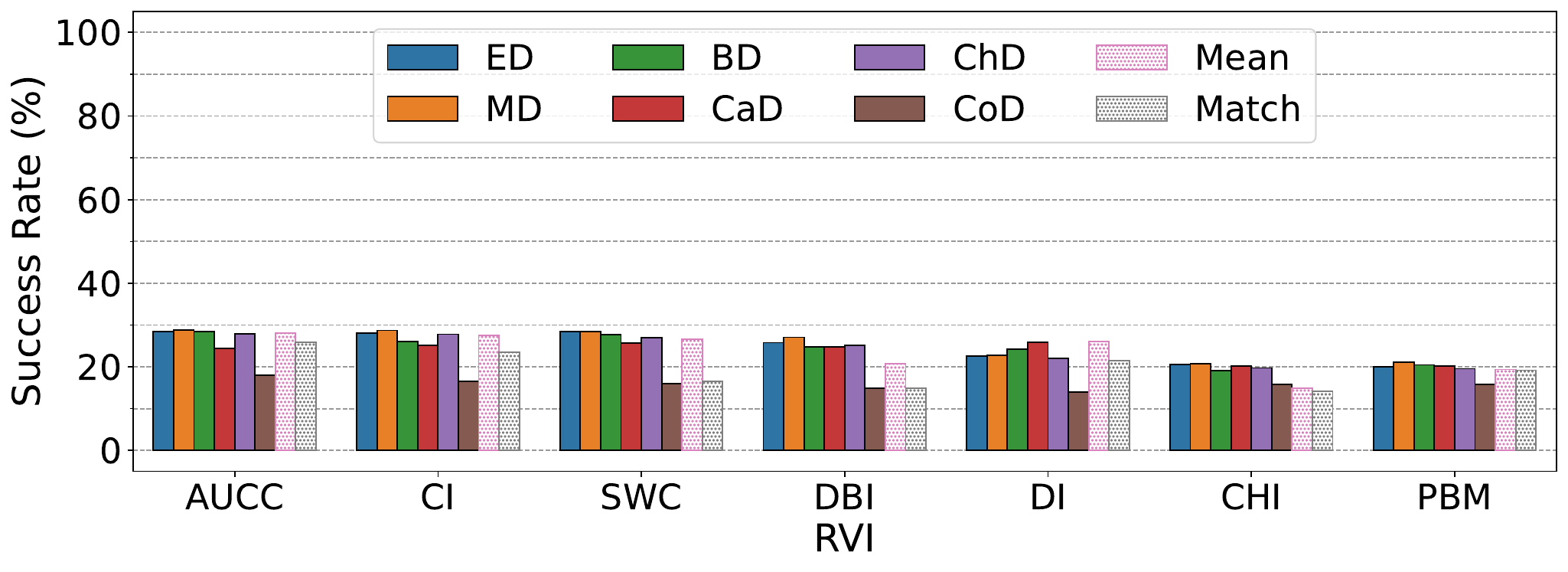}
        \caption{Gagolewski (over SPs)}
        \label{Fig:CoincidencePlots-Gag-SP}
    \end{subfigure}
       ~
    \begin{subfigure}[t]{0.49\textwidth}
        \centering
        \includegraphics[width=\textwidth]{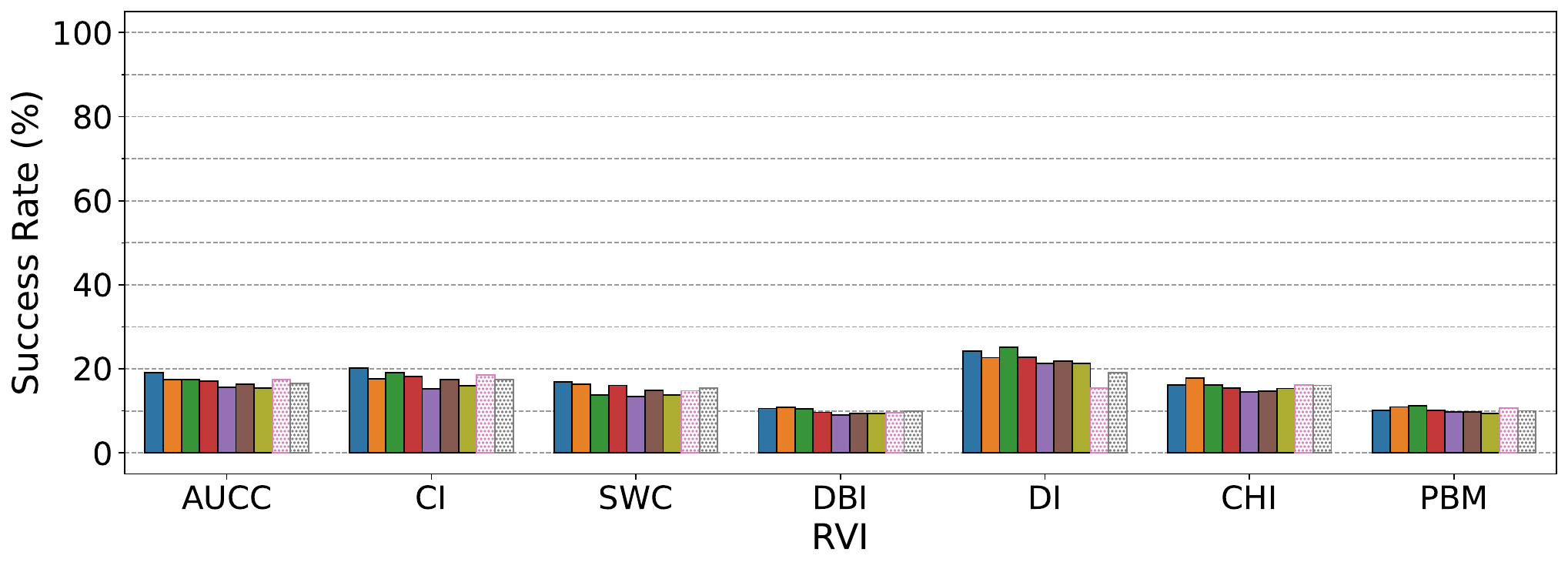} 
        \caption{UCR (over $k$)}
        \label{Fig:CoincidencePlots-UCR-K}
    \end{subfigure}%
    ~
    \begin{subfigure}[t]{0.49\textwidth}
        \centering
        \includegraphics[width=\textwidth]{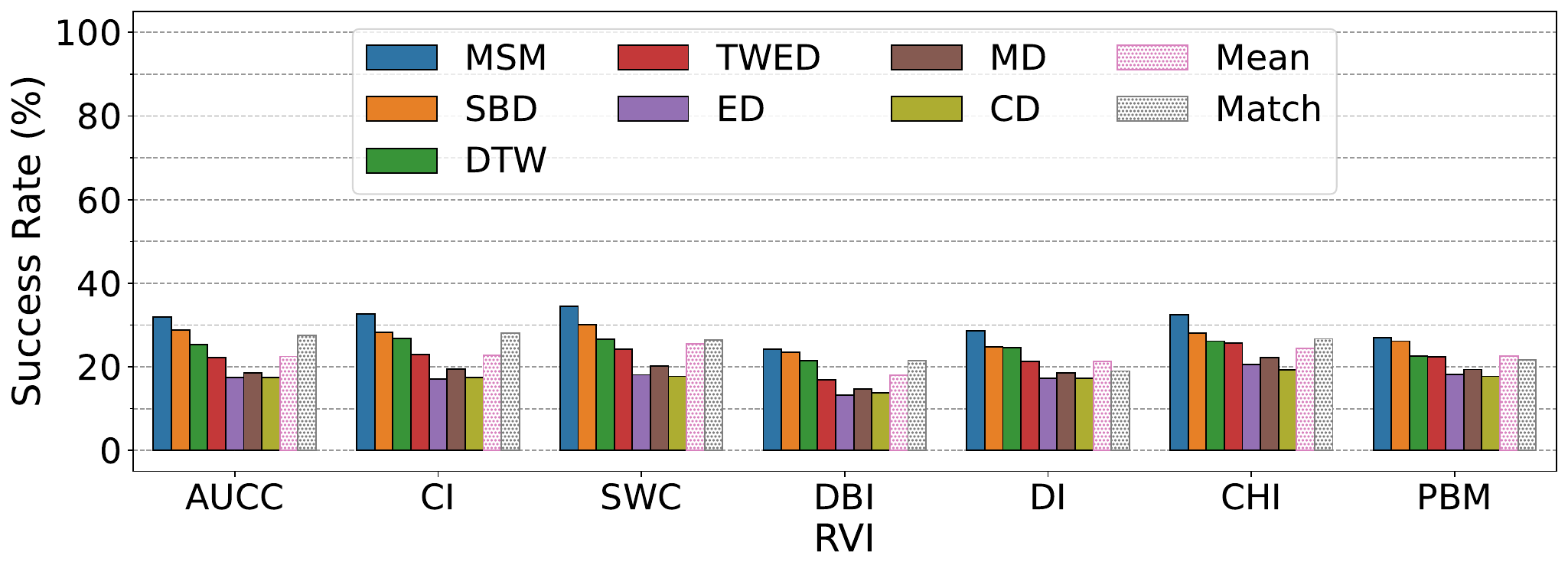}
        \caption{UCR (over SPs)}
        \label{Fig:CoincidencePlots-UCR-SP}
    \end{subfigure}
    \begin{subfigure}[t]{0.49\textwidth}
        \centering
        \includegraphics[width=\textwidth]{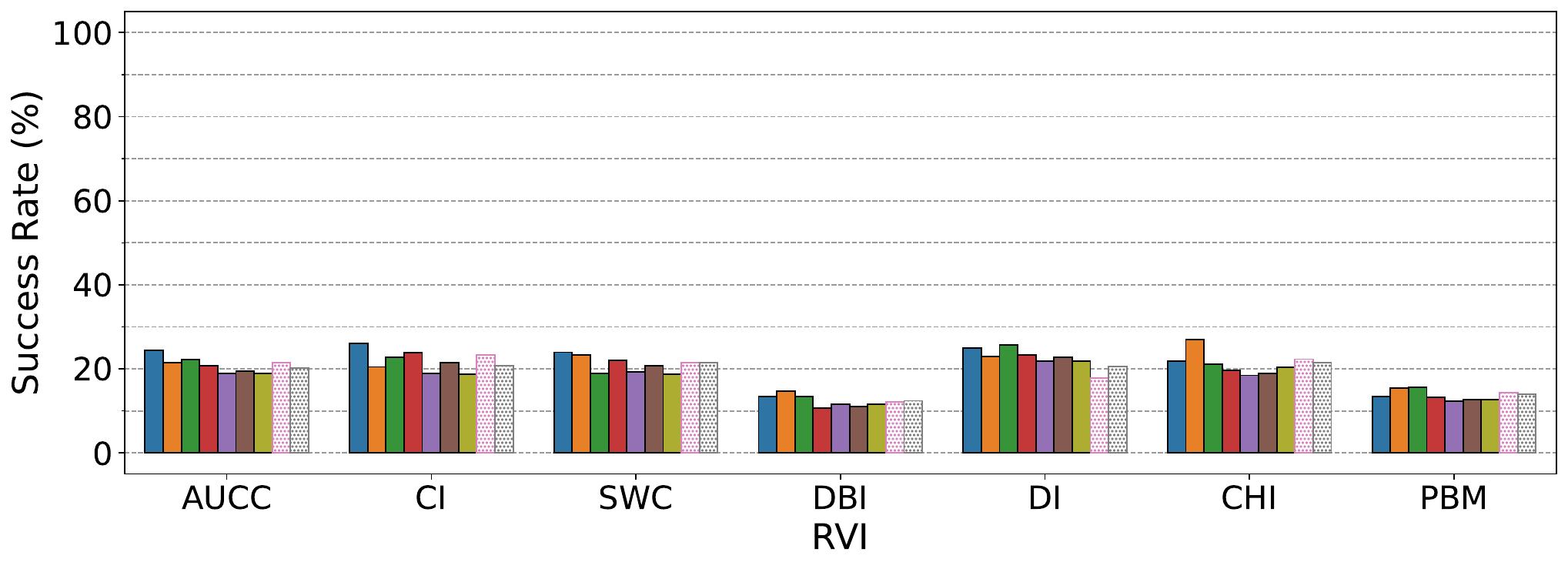} 
        \caption{UCR (over $k$ with max dataset ARI threshold of 0.6)}
        \label{Fig:CoincidencePlots-UCR-K-Threshold}
    \end{subfigure}%
    ~
    \begin{subfigure}[t]{0.49\textwidth}
        \centering
        \includegraphics[width=\textwidth]{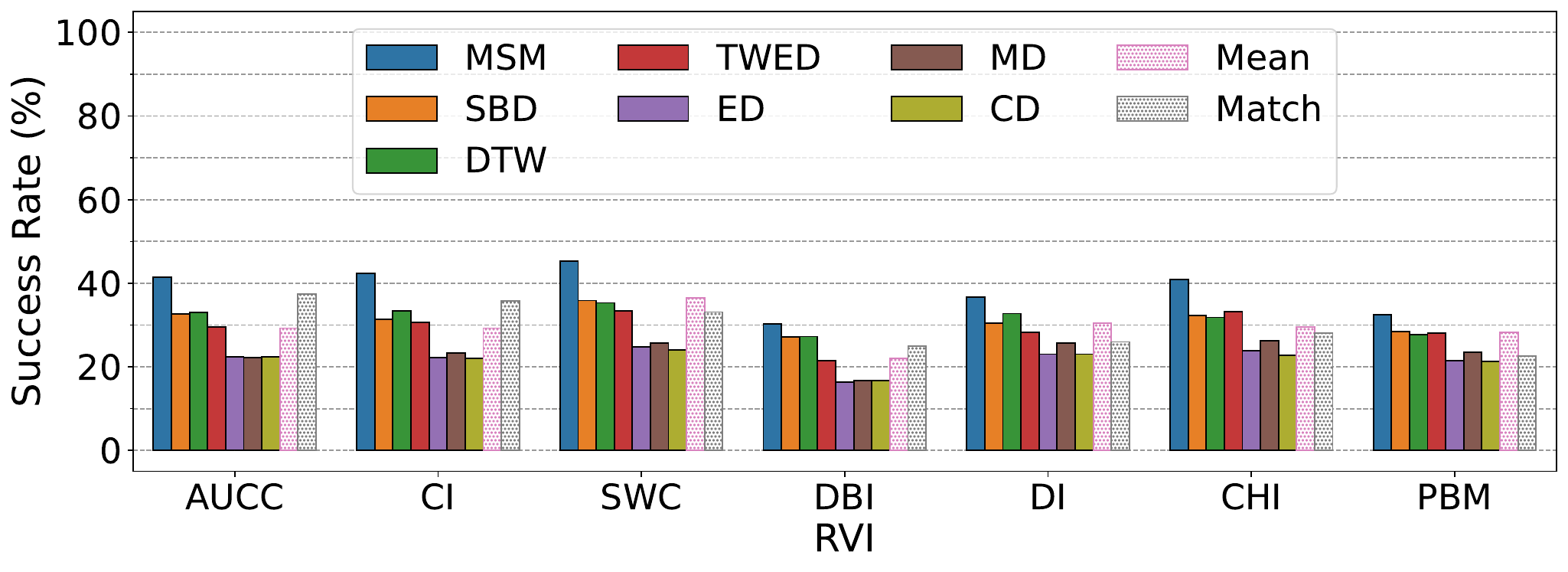}
        \caption{UCR (over SPs with max dataset ARI threshold of 0.6)}
        \label{Fig:CoincidencePlots-UCR-SP-Threshold}
    \end{subfigure}
    \caption{The success rates for coincidence of optimal values for ARI and the different versions of each RVI. The first three rows show the results for each battery, and the left and right columns show the success rates for the $k$- and SP-selection tasks respectively. The final row has been produced for a subset of 41 datasets from the UCR archive where at least one of the dataset partitions had an ARI in excess of 0.6 (see \Cref{Subsec:Experimental_Results-ThresholdingEVIs}). Note that the legend on the right plot serves both plots in each row.}
    \label{Fig:CoincidencePlots}
\end{figure}

\begin{figure}[!htb]
    \centering
    \includegraphics[width=0.95\textwidth]{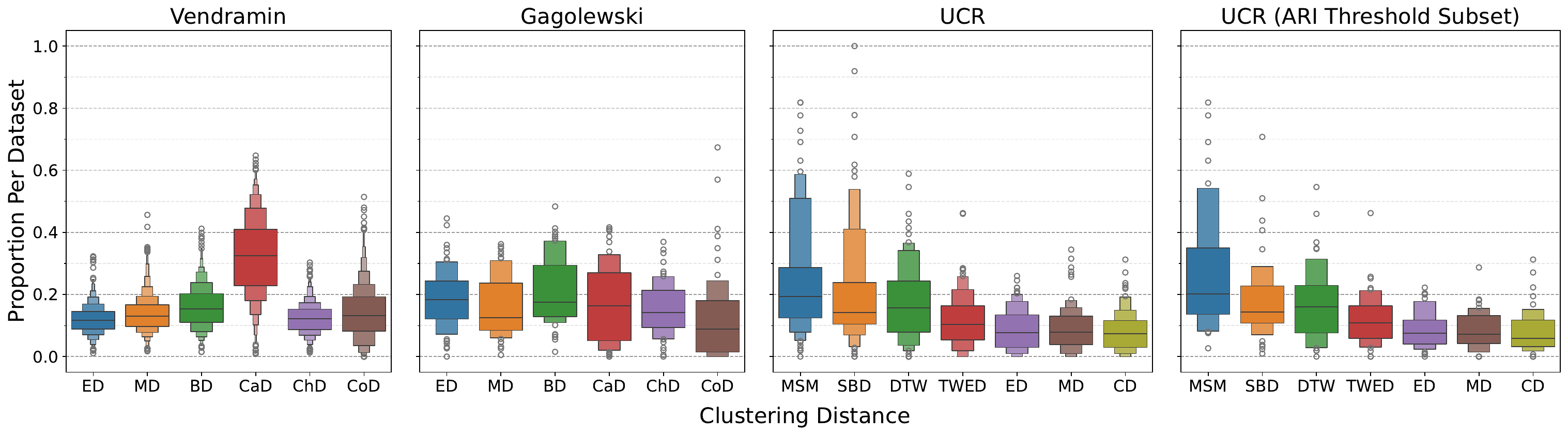}
    \caption{The proportion of times each SP generated the optimal partition according to the ARI for various combinations of clustering algorithm and $k$ for each of the datasets, i.e. which SPs were found to be optimal according to the ARI when producing \Cref{Fig:CoincidencePlots-Vend-SP,Fig:CoincidencePlots-Gag-SP,Fig:CoincidencePlots-UCR-SP,Fig:CoincidencePlots-UCR-SP-Threshold}.}
    \label{Fig:Max-ARI-Distances}
\end{figure}

As described in \Cref{Sec:Methodology-Main} with reference to \Cref{Tab:MethodologyExample1}, we have recorded a boolean variable, CO, which indicates whether the optimal value of an RVI coincides with the optimal value of the ARI over a subset of partitions. In order to assess the $k$-selection task, CO is computed over a subset of partitions which vary by the number of clusters, with the dataset, clustering algorithm and SP held static. The corresponding success rates shown in \Cref{Fig:CoincidencePlots-Vend-K,Fig:CoincidencePlots-Gag-K,Fig:CoincidencePlots-UCR-K} are obtained by averaging these CO values over each of the datasets, clustering algorithms and SPs. Similarly, to assess the SP-selection task, CO is computed over a subset of partitions which vary by the SP used for clustering, with the dataset, clustering algorithm and number of clusters held static. The corresponding success rates shown in \Cref{Fig:CoincidencePlots-Vend-SP,Fig:CoincidencePlots-Gag-SP,Fig:CoincidencePlots-UCR-SP} are obtained by averaging these CO values obtained over each of the datasets, clustering algorithms and numbers of clusters. 

Firstly we consider the success rates for $k$-selection on the Vendramin battery. The observed rates in this case are quite high for all fixed schemes,
which is to be expected for a battery with well-separated and compact globular clusters. However, more suitable distance measures such as ED, MD and ChD still appear to have performed better across all of the RVIs than CoD and CaD. Interestingly, there doesn't appear to be any notable improvement to the $k$-selection success rates when a fixed-SP scheme is replaced by a matching-SP scheme. This observation is consistent across all three batteries, and also holds for the mean-SP scheme. There appears to be no clear advantage in using the same SP to compute the RVI as was used to perform the clustering when conducting $k$-selection.

The success rates for the Vendramin battery deteriorate significantly when the task shifts to selecting the optimal SP for the same battery (\Cref{Fig:CoincidencePlots-Vend-SP}). For most of the fixed schemes, the rates are halved, or worse, as is the case for CHI and PBM. Interestingly, whilst CHI and PBM generally performed the best out of all the RVIs for $k$-selection, they performed the worst by far for SP-selection.
Overall, the mean-SP and matching-SP schemes did not perform significantly differently from the fixed-SP schemes for this battery on both tasks. The mean-SP scheme was however, superior to all of its constituent fixed-SP schemes in the case of CI and AUCC on the SP-selection task. Similarly, the matching-SP scheme was superior to all its constituents for AUCC on the SP-selection task. This was not the case for any RVIs in the other two batteries on either task.

For the Gagolewski and UCR batteries, the success rates on the $k$-selection task were drastically lower than those of the Vendramin battery. This observation aligns with what is known about the performance of these traditional indices which weren't designed for more complex, real-world datasets involving clusters with varying densities, overlap or arbitrary shapes \cite{Arbelaitz2013AnIndices}. The success rates by RVI across both tasks are largely consistent for the Gagolewski battery, with some RVIs (SWC, DI, CHI) showing slightly better performance on $k$-selection and others (AUCC, CI) showing slightly better performance on SP-selection. Interestingly however, the SP-selection task on the UCR battery saw modestly higher success rates for all RVIs other than DI when compared to $k$-selection. Furthermore, these success rate improvements were overall higher for the time series specific SPs, such as MSM, than for the generic SPs, such as ED --- the typical RVI default. However the success rates for SP-selection on both of these batteries were still quite low overall, generally not in excess of 30\%. 

For the Vendramin battery, all RVIs employing a fixed-CaD scheme rejected the standard trend, instead performing comparatively poorly on the $k$-selection task and showing only a minor reduction in performance for the SP-selection task. Furthermore, this scheme had the highest success rate of all the fixed-SP schemes for every RVI on the SP-selection task. This is likely due to the fact that, as shown in \Cref{Fig:Max-ARI-Distances}, the Canberra distance was most frequently producing partitions with the largest ARI for the datasets in this battery. And yet, this did not translate into competitive success rates on the $k$-selection task. In a similar way, the ranking of medians for the UCR battery SPs in \Cref{Fig:Max-ARI-Distances} mimicked the ranking of the fixed-SP scheme success rates in \Cref{Fig:CoincidencePlots-UCR-SP}, meaning this ranking could just be a chance effect due to the complexion of datasets within the battery.

\subsection{Correlation Analyses}
\label{Subsec:Experimental_Results-Correlation}

\begin{figure}[!htb]
    \centering
    \includegraphics[width=0.6\textwidth]{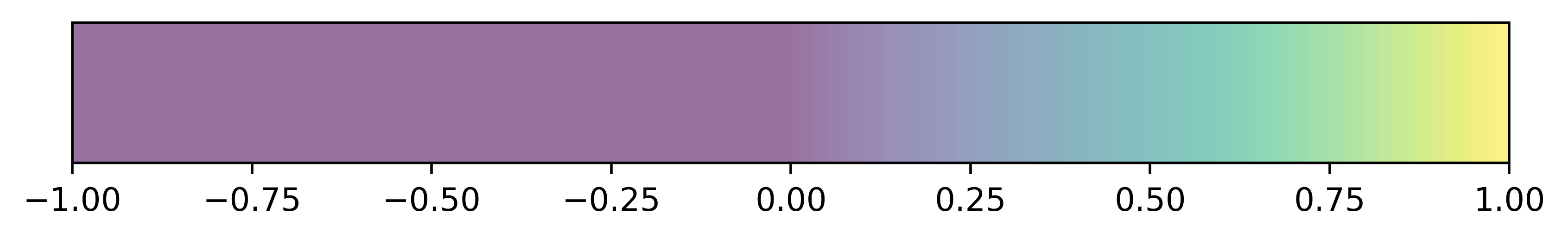}
    \caption{Reference scale for the Viridis colormap.}
    \label{fig:viridis_colorbar}
\end{figure}

The Pearson correlations between the ARI and all combinations of RVI and evaluation scheme have been summarised for each of the three batteries in \Cref{Tab:Vend-ARI-Pearson-All,Tab:Gag-ARI-Pearson-All,Tab:UCR-ARI-Pearson-All}, which use the Viridis colormap (\Cref{fig:viridis_colorbar}) to facilitate interpretation. Each table presents the median correlations for one battery on first the $k$-selection task at the top, followed by the SP-selection task below.
Each cell has been obtained by first computing the median of the distribution of correlations across all variations within an individual dataset, and then computing the median of this distribution across the battery. For SP-selection, this means that first, a median is computed from 441 correlations (49 numbers of clusters $\times$ 9 clustering algorithms) on all 972 datasets from the Vendramin battery (these distributions are provided in the supplementary materials). Then the median of these 972 medians is presented in the table.
Medians have been employed due to the skewness of the distributions of correlation values at both levels. The correlations have been rounded to 3 significant figures, but their relative magnitudes and overall trends are more important than the specific values.

\begin{table}[!htb]
    \footnotesize
    \begin{adjustbox}{center}
    \begin{tabular}{ccccccccc} \toprule
         \multicolumn{1}{l}{} & \multicolumn{8}{c}{\textbf{Evaluation Schemes}} \\ \cmidrule(l){2-9}
          \textbf{RVI} & \textbf{ED} & \textbf{MD} & \textbf{BD} & \textbf{CaD} & \textbf{ChD} & \textbf{CoD} & \textbf{Mean} & \textbf{Match}\\ \midrule
        \multicolumn{9}{c}{\textcolor{gray}{\textbf{Selecting Optimal Partition With Varying $k$}}} \\ \cmidrule(l){1-9}
        AUCC & \mc{0.756}  & \mc{0.759}  & \mc{0.747}  & \mc{0.736}  & \mc{0.752}  & \mc{0.748}  & \mc{0.760}  & \mc{0.747}  \\
        CI & \mc{0.726}  & \mc{0.726}  & \mc{0.711}  & \mc{0.753}  & \mc{0.703}  & \mc{0.635}  & \mc{0.715}  & \mc{0.715}  \\
        SWC & \mc{0.764}  & \mc{0.763}  & \mc{0.759}  & \mc{0.786}  & \mc{0.774}  & \mc{0.770}  & \mc{0.772}  & \mc{0.754}  \\
        DBI & \mc{0.492}  & \mc{0.471}  & \mc{0.482}  & \mc{0.537}  & \mc{0.530}  & \mc{0.395}  & \mc{0.471}  & \mc{0.485}  \\
        DI & \mc{0.172}  & \mc{0.183}  & \mc{0.180}  & \mc{0.069}  & \mc{0.153}  & \mc{0.204}  & \mc{0.188}  & \mc{0.160}  \\
        CHI & \mc{0.858}  & \mc{0.856}  & \mc{0.858}  & \mc{0.838}  & \mc{0.854}  & \mc{0.731}  & \mc{0.767}  & \mc{0.850}  \\
        PBM & \mc{0.850}  & \mc{0.852}  & \mc{0.849}  & \mc{0.759}  & \mc{0.851}  & \mc{0.712}  & \mc{0.853}  & \mc{0.838}  \\ \cmidrule(l){1-9}
        \multicolumn{9}{c}{\textcolor{gray}{\textbf{Selecting Optimal Partition With Varying SP}}} \\ \cmidrule(l){1-9}
        AUCC & \mc{0.613}  & \mc{0.658}  & \mc{0.741}  & \mc{0.717}  & \mc{0.616}  & \mc{0.605}  & \mc{0.831}  & \mc{0.702}  \\
        CI & \mc{0.524}  & \mc{0.591}  & \mc{0.699}  & \mc{0.677}  & \mc{0.462}  & \mc{0.433}  & \mc{0.804}  & \mc{0.346}  \\
        SWC & \mc{0.621}  & \mc{0.632}  & \mc{0.616}  & \mc{0.671}  & \mc{0.538}  & \mc{0.407}  & \mc{0.654}  & \mc{0.440}  \\
        DBI & \mc{0.439}  & \mc{0.533}  & \mc{0.535}  & \mc{0.528}  & \mc{0.196}  & \mc{0.216}  & \mc{0.355}  & \mc{0.382}  \\
        DI & \mc{0.095}  & \mc{0.182}  & \mc{0.377}  & \mc{0.463}  & \mc{-0.028}  & \mc{0.145}  & \mc{0.348}  & \mc{0.057}  \\
        CHI & \mc{-0.023}  & \mc{0.021}  & \mc{-0.013}  & \mc{0.456}  & \mc{-0.011}  & \mc{-0.139}  & \mc{-0.136}  & \mc{0.079}  \\
        PBM & \mc{0.000}  & \mc{0.048}  & \mc{0.027}  & \mc{0.438}  & \mc{-0.012}  & \mc{-0.093}  & \mc{0.027}  & \mc{-0.140}  \\
         \bottomrule
    \end{tabular}
    \end{adjustbox}
    \caption{Median Pearson correlations between the ARI and the different versions of each RVI for the $k$-selection task (top) and SP-selection task (bottom) on the Vendramin battery.}
    \label{Tab:Vend-ARI-Pearson-All}
\end{table}

\begin{table}[!htb]
    \footnotesize
    \begin{adjustbox}{center}
    \begin{tabular}{ccccccccc} \toprule
         \multicolumn{1}{l}{} & \multicolumn{8}{c}{\textbf{Evaluation Schemes}} \\ \cmidrule(l){2-9}
          \textbf{RVI} & \textbf{ED} & \textbf{MD} & \textbf{BD} & \textbf{CaD} & \textbf{ChD} & \textbf{CoD} & \textbf{Mean} & \textbf{Match} \\ \midrule
        \multicolumn{9}{c}{\textcolor{gray}{\textbf{Selecting Optimal Partition With Varying $k$}}} \\ \cmidrule(l){1-9}
        AUCC & \mc{0.252}  & \mc{0.259}  & \mc{0.265}  & \mc{0.177}  & \mc{0.204}  & \mc{0.285}  & \mc{0.235}  & \mc{0.135}  \\
        CI & \mc{0.246}  & \mc{0.217}  & \mc{-0.009}  & \mc{0.390}  & \mc{0.187}  & \mc{-0.138}  & \mc{0.203}  & \mc{0.144}  \\
        SWC & \mc{0.626}  & \mc{0.627}  & \mc{0.507}  & \mc{0.685}  & \mc{0.595}  & \mc{0.291}  & \mc{0.576}  & \mc{0.635}  \\
        DBI & \mc{0.311}  & \mc{0.356}  & \mc{0.085}  & \mc{0.201}  & \mc{0.261}  & \mc{0.097}  & \mc{0.153}  & \mc{0.151}  \\
        DI & \mc{0.217}  & \mc{0.205}  & \mc{0.205}  & \mc{0.138}  & \mc{0.266}  & \mc{0.026}  & \mc{0.253}  & \mc{0.170}  \\
        CHI & \mc{0.527}  & \mc{0.498}  & \mc{0.423}  & \mc{0.459}  & \mc{0.525}  & \mc{0.179}  & \mc{0.330}  & \mc{0.479}  \\
        PBM & \mc{0.405}  & \mc{0.491}  & \mc{0.032}  & \mc{0.191}  & \mc{0.454}  & \mc{0.053}  & \mc{0.029}  & \mc{0.215}  \\ \cmidrule(l){1-9} 
        \multicolumn{9}{c}{\textcolor{gray}{\textbf{Selecting Optimal Partition With Varying SP}}} \\ \cmidrule(l){1-9}
        AUCC & \mc{0.544}  & \mc{0.552}  & \mc{0.394}  & \mc{0.283}  & \mc{0.469}  & \mc{-0.124}  & \mc{0.461}  & \mc{0.296}  \\
        CI & \mc{0.551}  & \mc{0.547}  & \mc{0.260}  & \mc{0.333}  & \mc{0.512}  & \mc{-0.014}  & \mc{0.526}  & \mc{0.166}  \\
        SWC & \mc{0.535}  & \mc{0.549}  & \mc{0.485}  & \mc{0.476}  & \mc{0.518}  & \mc{-0.115}  & \mc{0.422}  & \mc{-0.006}  \\
        DBI & \mc{0.471}  & \mc{0.489}  & \mc{0.389}  & \mc{0.394}  & \mc{0.436}  & \mc{-0.116}  & \mc{-0.087}  & \mc{-0.146}  \\
        DI & \mc{0.449}  & \mc{0.488}  & \mc{0.151}  & \mc{0.254}  & \mc{0.463}  & \mc{-0.336}  & \mc{0.454}  & \mc{0.308}  \\
        CHI & \mc{0.258}  & \mc{0.206}  & \mc{0.059}  & \mc{0.206}  & \mc{0.238}  & \mc{-0.302}  & \mc{-0.295}  & \mc{-0.310}  \\
        PBM & \mc{0.252}  & \mc{0.201}  & \mc{0.066}  & \mc{0.206}  & \mc{0.237}  & \mc{-0.303}  & \mc{-0.101}  & \mc{-0.214}  \\
         \bottomrule
    \end{tabular}
    \end{adjustbox}
    \caption{Median Pearson correlations between the ARI and the different versions of each RVI for the $k$-selection task (top) and SP-selection task (bottom) on the Gagolewski battery.}
    \label{Tab:Gag-ARI-Pearson-All}
\end{table}

For both the $k$- and SP-selection tasks on the Vendramin battery, the matching-SP scheme always performed worse than one or multiple of the fixed schemes, including the common RVI default ED. This observation was consistent for both the Gagolewski and UCR batteries as well. The correlations for the mean-SP scheme were similar to the fixed-SP and matching-SP schemes for $k$-selection on the Vendramin battery, but the mean-SP was actually superior to the matching-SP scheme for most of the RVIs on SP-selection. This is also true for the Gagolewski battery, whilst there was a high degree of similarity between the two schemes for the UCR battery. 

If we focus on SP-selection for the UCR battery (\Cref{Tab:UCR-ARI-Pearson-All}), all of the fixed schemes that used generic feature-vector distances such as MD, CD, and ED (which is the default distance measure for most RVIs) performed much worse than their time series counterparts, such as MSM, SBD and DTW. As mentioned in \Cref{Sec:Whats_the_problem}, this is likely due to their inability to accurately capture the relevant similarity structure within these time series datasets. For a similar reason, the fixed-CoD scheme has performed poorly on this task for the Gagolewski battery. For the Vendramin and Gagolewski batteries the ED, MD and ChD fixed schemes all perform quite similarly on both $k$- and SP-selection, which is to be expected due to similarities between the distance measures themselves.

Overall, these observations from the correlation results are largely consistent with observations in the preceding section concerned with the coincidence of RVI and EVI optima. Indeed if we focus on the Vendramin battery for a moment, some highly similar observations can be made. For instance, the median correlations overall tend to be higher for the $k$-selection task than for the SP-selection task. Furthermore, the mean-SP scheme has a higher median correlation for the SP-selection task than any of its constituents for AUCC and CI, and the fixed-CaD scheme has amongst the highest median correlations for all RVIs (notably so for CHI and PBM) on the same task. We also see that CHI and PBM are the best performing RVIs for the $k$-selection task, but mostly have median correlations around 0 for the SP-selection task. However, the correlation results are more damning for the DI on $k$-selection in all three batteries than the success rate results, suggesting that the DI does not align well with the ARI for $k \neq k^\star$. The DI was also found to perform quite poorly in other comparative studies \cite{Arbelaitz2013AnIndices,Vendramin2010RelativeOverview}. 

\begin{table}[!htb]
    \footnotesize
    \begin{adjustbox}{center}
    \begin{tabular}{cccccccccc} \toprule
         \multicolumn{1}{l}{} & \multicolumn{9}{c}{\textbf{Evaluation Schemes}} \\ \cmidrule(l){2-10}
          \textbf{RVI} & \textbf{MSM} & \textbf{SBD} & \textbf{DTW} & \textbf{TWED} & \textbf{ED} & \textbf{MD} & \textbf{CD} & \textbf{Mean} & \textbf{Match} \\ \midrule
        \multicolumn{10}{c}{\textcolor{gray}{\textbf{Selecting Optimal Partition With Varying $k$}}} \\ \cmidrule(l){1-10}
        AUCC & \mc{0.372}  & \mc{0.293}  & \mc{0.368}  & \mc{0.356}  & \mc{0.320}  & \mc{0.345}  & \mc{0.320}  & \mc{0.384}  & \mc{0.312}  \\
        CI & \mc{0.415}  & \mc{0.413}  & \mc{0.548}  & \mc{0.329}  & \mc{0.169}  & \mc{0.358}  & \mc{0.366}  & \mc{0.436}  & \mc{0.412}  \\
        SWC & \mc{-0.179}  & \mc{-0.228}  & \mc{-0.211}  & \mc{-0.118}  & \mc{-0.265}  & \mc{-0.185}  & \mc{-0.224}  & \mc{-0.246}  & \mc{-0.139}  \\
        DBI & \mc{-0.179}  & \mc{-0.154}  & \mc{-0.188}  & \mc{-0.169}  & \mc{-0.192}  & \mc{-0.178}  & \mc{-0.175}  & \mc{-0.214}  & \mc{-0.152}  \\
        DI & \mc{-0.092}  & \mc{-0.207}  & \mc{-0.134}  & \mc{-0.150}  & \mc{-0.234}  & \mc{-0.147}  & \mc{-0.237}  & \mc{-0.195}  & \mc{-0.126}  \\
        CHI & \mc{-0.107}  & \mc{0.059}  & \mc{0.072}  & \mc{-0.159}  & \mc{-0.148}  & \mc{-0.230}  & \mc{-0.041}  & \mc{-0.013}  & \mc{-0.121}  \\
        PBM & \mc{-0.479}  & \mc{-0.289}  & \mc{-0.061}  & \mc{-0.509}  & \mc{-0.508}  & \mc{-0.506}  & \mc{-0.269}  & \mc{-0.403}  & \mc{-0.349}  \\ \cmidrule(l){1-10}
        \multicolumn{10}{c}{\textcolor{gray}{\textbf{Selecting Optimal Partition With Varying SP}}} \\ \cmidrule(l){1-10}
        AUCC & \mc{0.450}  & \mc{0.362}  & \mc{0.231}  & \mc{0.127}  & \mc{-0.173}  & \mc{-0.038}  & \mc{-0.173}  & \mc{0.089}  & \mc{0.180}  \\
        CI & \mc{0.444}  & \mc{0.343}  & \mc{0.232}  & \mc{0.143}  & \mc{-0.147}  & \mc{-0.041}  & \mc{-0.108}  & \mc{0.158}  & \mc{0.216}  \\
        SWC & \mc{0.544}  & \mc{0.359}  & \mc{0.280}  & \mc{0.232}  & \mc{-0.139}  & \mc{0.008}  & \mc{-0.131}  & \mc{0.205}  & \mc{0.207}  \\
        DBI & \mc{0.099}  & \mc{0.050}  & \mc{0.070}  & \mc{-0.067}  & \mc{-0.297}  & \mc{-0.208}  & \mc{-0.278}  & \mc{-0.044}  & \mc{-0.001}  \\
        DI & \mc{0.114}  & \mc{0.066}  & \mc{-0.008}  & \mc{-0.034}  & \mc{-0.216}  & \mc{-0.176}  & \mc{-0.225}  & \mc{-0.087}  & \mc{-0.037}  \\
        CHI & \mc{0.329}  & \mc{0.262}  & \mc{0.236}  & \mc{0.241}  & \mc{-0.004}  & \mc{0.063}  & \mc{-0.054}  & \mc{0.163}  & \mc{0.153}  \\
        PBM & \mc{0.209}  & \mc{0.188}  & \mc{0.093}  & \mc{0.115}  & \mc{-0.074}  & \mc{0.000}  & \mc{-0.085}  & \mc{0.076}  & \mc{0.061}  \\
         \bottomrule
    \end{tabular}
    \end{adjustbox}
    \caption{Median Pearson correlations between the ARI and the different versions of each RVI for the $k$-selection task (top) and SP-selection task (bottom) on the UCR battery.}
    \label{Tab:UCR-ARI-Pearson-All}
\end{table}

\begin{table}[!htb]
    \footnotesize
    \begin{adjustbox}{center}
    \begin{tabular}{cccccccccc} \toprule
         \multicolumn{1}{l}{} & \multicolumn{9}{c}{\textbf{Evaluation Schemes}} \\ \cmidrule(l){2-10}
          \textbf{RVI} & \textbf{MSM} & \textbf{SBD} & \textbf{DTW} & \textbf{TWED} & \textbf{ED} & \textbf{MD} & \textbf{CD} & \textbf{Mean} & \textbf{Match} \\ \midrule
        \multicolumn{10}{c}{\textcolor{gray}{\textbf{Selecting Optimal Partition With Varying $k$}}} \\ \cmidrule(l){1-10}
        AUCC & \mc{0.620}  & \mc{0.458}  & \mc{0.565}  & \mc{0.512}  & \mc{0.400}  & \mc{0.419}  & \mc{0.400}  & \mc{0.579}  & \mc{0.458}  \\
        CI & \mc{0.593}  & \mc{0.470}  & \mc{0.693}  & \mc{0.466}  & \mc{0.371}  & \mc{0.442}  & \mc{0.362}  & \mc{0.589}  & \mc{0.497}  \\
        SWC & \mc{0.359}  & \mc{0.187}  & \mc{0.070}  & \mc{0.349}  & \mc{0.088}  & \mc{0.130}  & \mc{0.039}  & \mc{0.161}  & \mc{0.181}  \\
        DBI & \mc{-0.074}  & \mc{-0.109}  & \mc{-0.128}  & \mc{-0.079}  & \mc{-0.207}  & \mc{-0.221}  & \mc{-0.187}  & \mc{-0.191}  & \mc{-0.092}  \\
        DI & \mc{0.000}  & \mc{-0.188}  & \mc{0.000}  & \mc{-0.018}  & \mc{-0.241}  & \mc{-0.080}  & \mc{-0.252}  & \mc{-0.028}  & \mc{-0.000}  \\
        CHI & \mc{0.004}  & \mc{0.193}  & \mc{0.263}  & \mc{0.013}  & \mc{-0.041}  & \mc{-0.005}  & \mc{0.095}  & \mc{0.096}  & \mc{0.085}  \\
        PBM & \mc{-0.436}  & \mc{-0.048}  & \mc{0.096}  & \mc{-0.512}  & \mc{-0.473}  & \mc{-0.474}  & \mc{-0.080}  & \mc{-0.256}  & \mc{-0.210}  \\ \cmidrule(l){1-10}
        \multicolumn{10}{c}{\textcolor{gray}{\textbf{Selecting Optimal Partition With Varying SP}}} \\ \cmidrule(l){1-10}
        AUCC & \mc{0.667}  & \mc{0.560}  & \mc{0.442}  & \mc{0.428}  & \mc{0.099}  & \mc{0.148}  & \mc{0.099}  & \mc{0.466}  & \mc{0.465}  \\
        CI & \mc{0.750}  & \mc{0.552}  & \mc{0.472}  & \mc{0.515}  & \mc{0.094}  & \mc{0.157}  & \mc{-0.055}  & \mc{0.497}  & \mc{0.362}  \\
        SWC & \mc{0.736}  & \mc{0.618}  & \mc{0.565}  & \mc{0.541}  & \mc{0.230}  & \mc{0.280}  & \mc{0.185}  & \mc{0.609}  & \mc{0.319}  \\
        DBI & \mc{0.259}  & \mc{0.269}  & \mc{0.277}  & \mc{-0.011}  & \mc{-0.285}  & \mc{-0.143}  & \mc{-0.195}  & \mc{0.047}  & \mc{0.093}  \\
        DI & \mc{0.415}  & \mc{0.200}  & \mc{0.273}  & \mc{0.197}  & \mc{-0.021}  & \mc{-0.002}  & \mc{-0.038}  & \mc{0.104}  & \mc{0.044}  \\
        CHI & \mc{0.498}  & \mc{0.293}  & \mc{0.297}  & \mc{0.412}  & \mc{-0.006}  & \mc{0.199}  & \mc{-0.084}  & \mc{0.291}  & \mc{0.169}  \\
        PBM & \mc{0.309}  & \mc{0.233}  & \mc{0.216}  & \mc{0.223}  & \mc{-0.040}  & \mc{0.062}  & \mc{-0.039}  & \mc{0.182}  & \mc{0.198}  \\
         \bottomrule
    \end{tabular}
    \end{adjustbox}
    \caption{Median Pearson correlations between the ARI and the different versions of each RVI for the $k$-selection task (top) and SP-selection task (bottom) on the 41 datasets from the UCR battery which have at least one partition achieving an ARI $>0.6$.}
    \label{Tab:UCR-Threshold-ARI-Pearson-All}
\end{table}

Different RVIs performed well on the time series data for $k$-selection compared to the feature-vector data. CHI, PBM and SWC appeared to perform the best for the Vendramin and Gagolewski batteries. On the UCR battery, AUCC and CI were the only RVIs that didn't record a negative median correlation, though the median correlations were still relatively small. This is also of note due to the fact that the failure of MSM, SBD, DTW and TWED to satisfy the identity of indiscernibles has not negatively impacted the performance of the corresponding fixed-SP schemes for the AUCC on either task. It should also be noted that there’s no indication that prototype-sensitive and prototype-insensitive RVIs perform very differently on these two tasks for any of the batteries, though we only considered one prototype definition. The generally low correlation values for the UCR battery make it difficult to recommend using any of these RVIs for either task when facing complex, real-world datasets, and for the sake of dependability, better alternatives should be preferred (see \Cref{Sec:Discussion}).

\begin{figure}[!ht]
    \centering
    \includegraphics[width=1.0\textwidth]{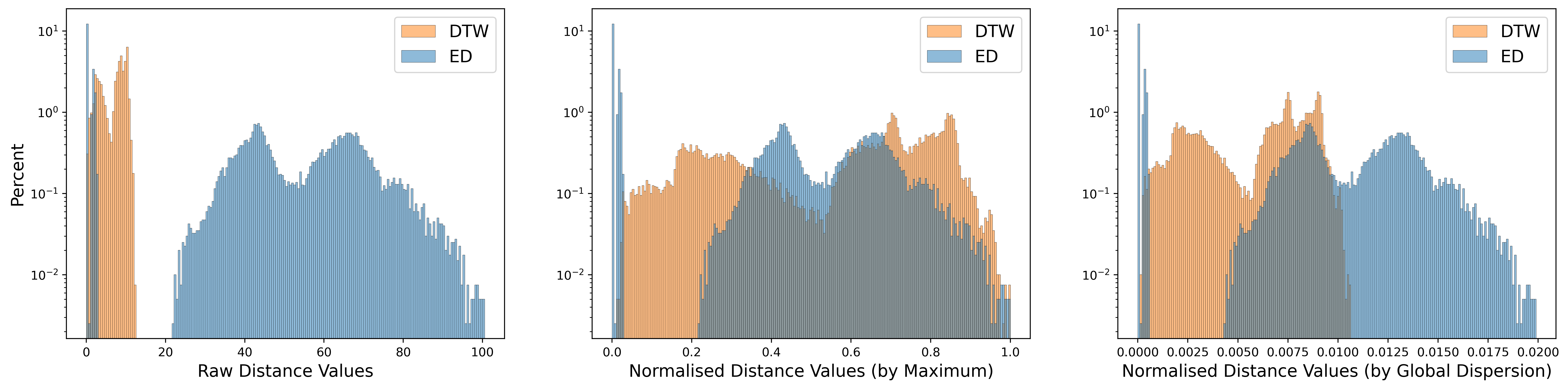}
    \caption{Log-scaled histograms of the strict upper triangle for two distance measures computed on the Trace dataset. The raw distance values are shown in the left-hand plot, and have been scaled using the maximum value for the middle plot, and using the global dispersion for the right-hand plot.}
    \label{Fig:Distance-Distributions}
\end{figure}

\paragraph{}
Finally, noting that PBM is the only RVI not invariant to multiplicative scaling, one might consider whether scaling the pairwise distances for each distance measure to a similar range might improve the correlation for PBM's matching-SP evaluation scheme. Not all SPs will have distance computations with comparable ranges. For instance, CD and SBD produce values within the range $\left [ 0, 2 \right ]$, whilst ED and the flexible time series extension, DTW, both produce values in the range $\left [ 0, \infty \right ]$. Even if the range of distance computations is similar for two SPs, this doesn't imply that the distribution of values within that range will be comparable for a particular dataset. In fact, it should be expected that DTW values will be concentrated much lower than ED values due to the inherent flexibility of the measure (see left subfigure in \Cref{Fig:Distance-Distributions}). One simple scaling approach could involve dividing all pairwise distances by the maximum pairwise distance, ensuring all values range between 0 and 1. Another approach adopted in \cite{DeCarvalho2012} scaled the distance matrices by the ``global dispersion". Each pairwise distance was divided by the sum of all deviations from the global (or grand) prototype, which for PBM would have the effect of removing $E_1$ from \Cref{Eqn:PBM}. These two simple scaling approaches, hereon referred to as maximum and global scaling respectively, are illustrated for ED and DTW with the Trace dataset in \Cref{Fig:Distance-Distributions}. To assess their impact on correlations, we applied them to the evaluation distances used in the matching-SP scheme for the PBM index.

For the Vendramin battery, both modes of scaling caused a slight improvement in the performance of PBM when selecting the optimal SP. The median correlations for the maximum and global scaling methods increased from $-0.140$, to $0.063$ and $0.062$ respectively. This outcome was not repeated in the more difficult batteries. For the Gagolewski battery, scaling reduced the already meagre median correlation from $-0.214$, to $-0.321$ and $-0.230$ respectively. For the UCR battery, an improvement was only noted for maximum scaling (from $0.061$ to $0.087$), whilst global scaling resulted in a reduction ($0.048$). In any case, scaling does not appear to have a significant impact on the correlation for the PBM index when using a matching evaluation scheme.

\subsection{Minimum EVI Requirements}
\label{Subsec:Experimental_Results-ThresholdingEVIs}

It is noteworthy that studies utilising the UCR archive for benchmarking the performance of existing clustering algorithms \cite{Javed2020} or justifying the introduction of novel clustering approaches \cite{a17020061} often display extremely low ground-truth recovery rates for many of the datasets. Observing the correlation or coincidence rates between RVI and EVI values for a dataset where none of the clustering approaches managed to recover any meaningful clustering structure related to the ground-truth labels is of questionable value. If a broad range of different clustering approaches have been applied to such datasets, it is likely that the ground-truth labels are not \textit{discoverable} for any ``off-the-shelf" clustering method that may be reasonably applied in practical circumstances. It is conceivable that a specialist distance measure or algorithm could be constructed to perfectly separate the clusters into the groupings observed in the ground-truth data, but this would defeat the need to apply any clustering approach. 

\begin{figure}[!ht]
    \centering
    \includegraphics[width=0.7\textwidth]{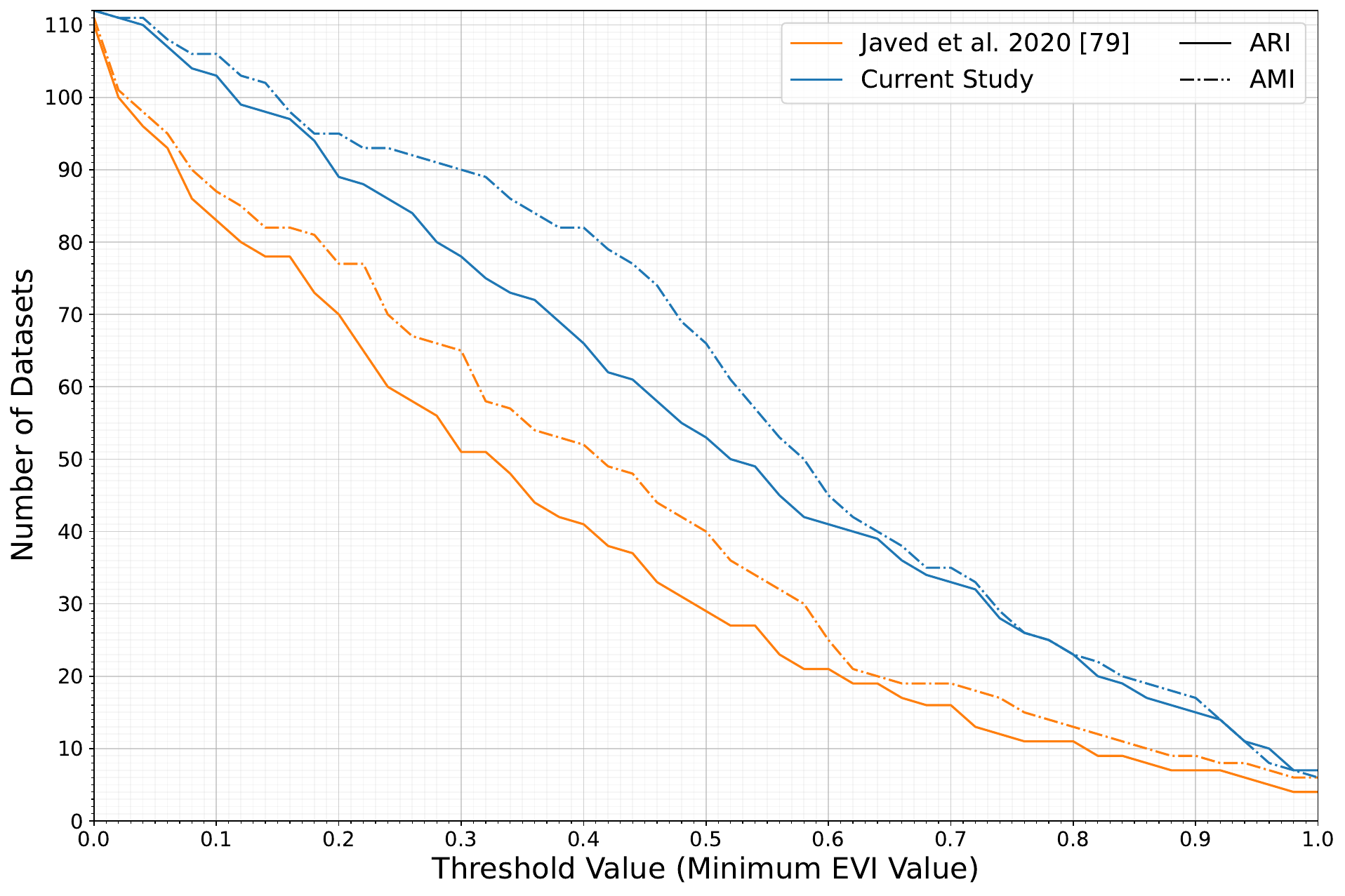}
    \caption{The number of datasets from the UCR archive clustered in \cite{Javed2020} where at least one of their 8 included clustering approaches produced a partition whose ARI or AMI exceeded the threshold values along the $x$-axis. A similar analysis is provided for the partitions from the present study.}
    \label{Fig:Javed2020-Threshold_Plot}
\end{figure}

In \cite{Javed2020}, ARI and AMI values are computed for 8 partitions, each of which is produced by a different clustering approach, for the same 112 UCR datasets used in this paper. \Cref{Fig:Javed2020-Threshold_Plot} displays the number of datasets where \textit{at least one} of the 8 corresponding ARIs and AMIs were in excess of the threshold value on the $x$-axis. The same has been plotted for the 693 partitions generated in this study from each combination of 7 distance measures, 9 clustering algorithms and 11 numbers of clusters. Very few datasets in \cite{Javed2020} had any partitions discovered with even a modicum of similarity to the ground-truth labels. In fact, \Cref{Fig:Javed2020-Threshold_Plot} shows that 42 (i.e. $112-70$) datasets had no partitions with ARI values in excess of 0.2, while only 7 datasets had an ARI in excess of above 0.9. However the breadth of partitions in our study affords only marginal improvements, suggesting that the issue is indeed with the meaningfulness of the ground-truth labels for clustering purposes.

\begin{figure}[!ht]
    \centering
    \begin{tabular}{cc}
        \includegraphics[width=0.49\textwidth]{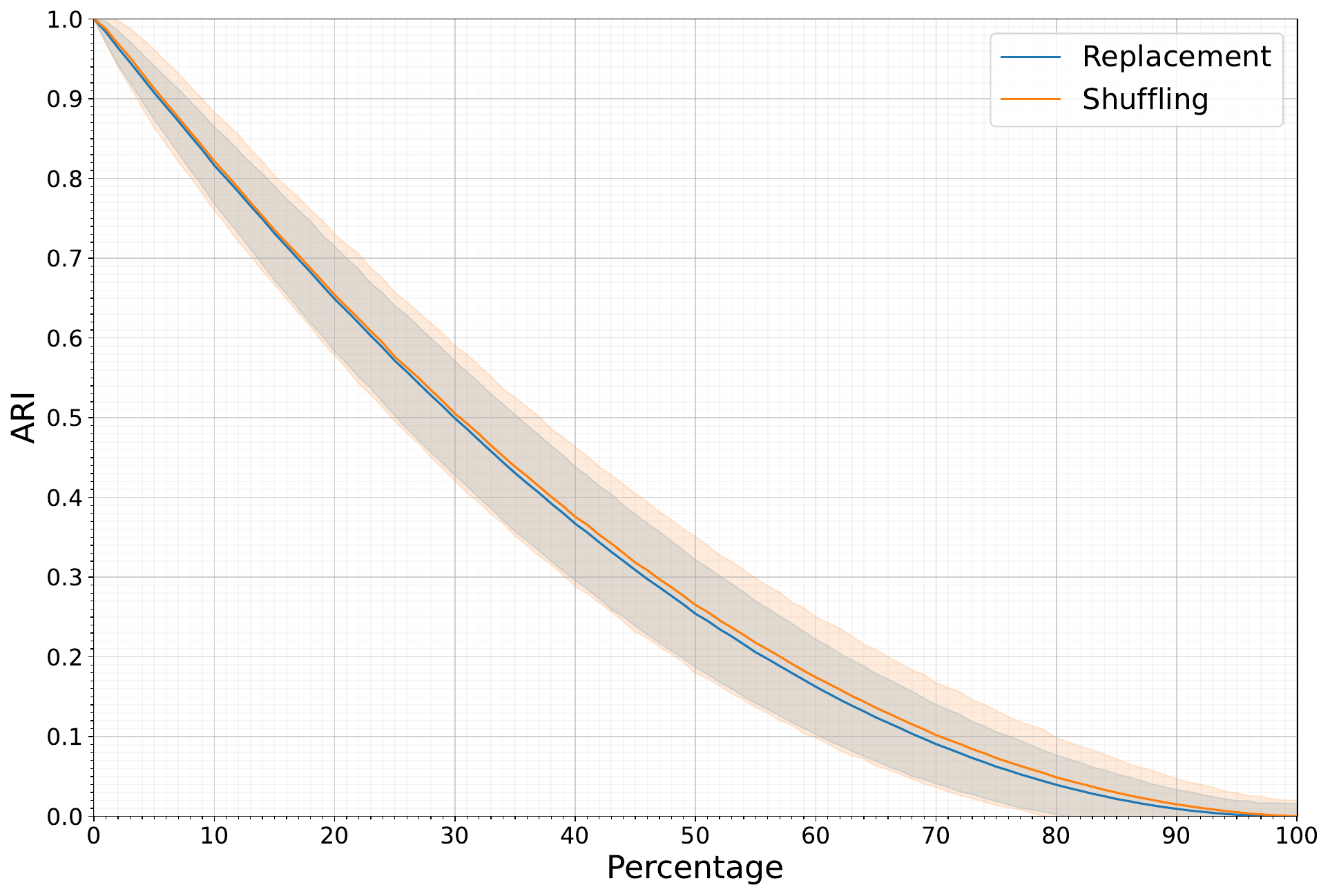} &   \includegraphics[width=0.49\textwidth]{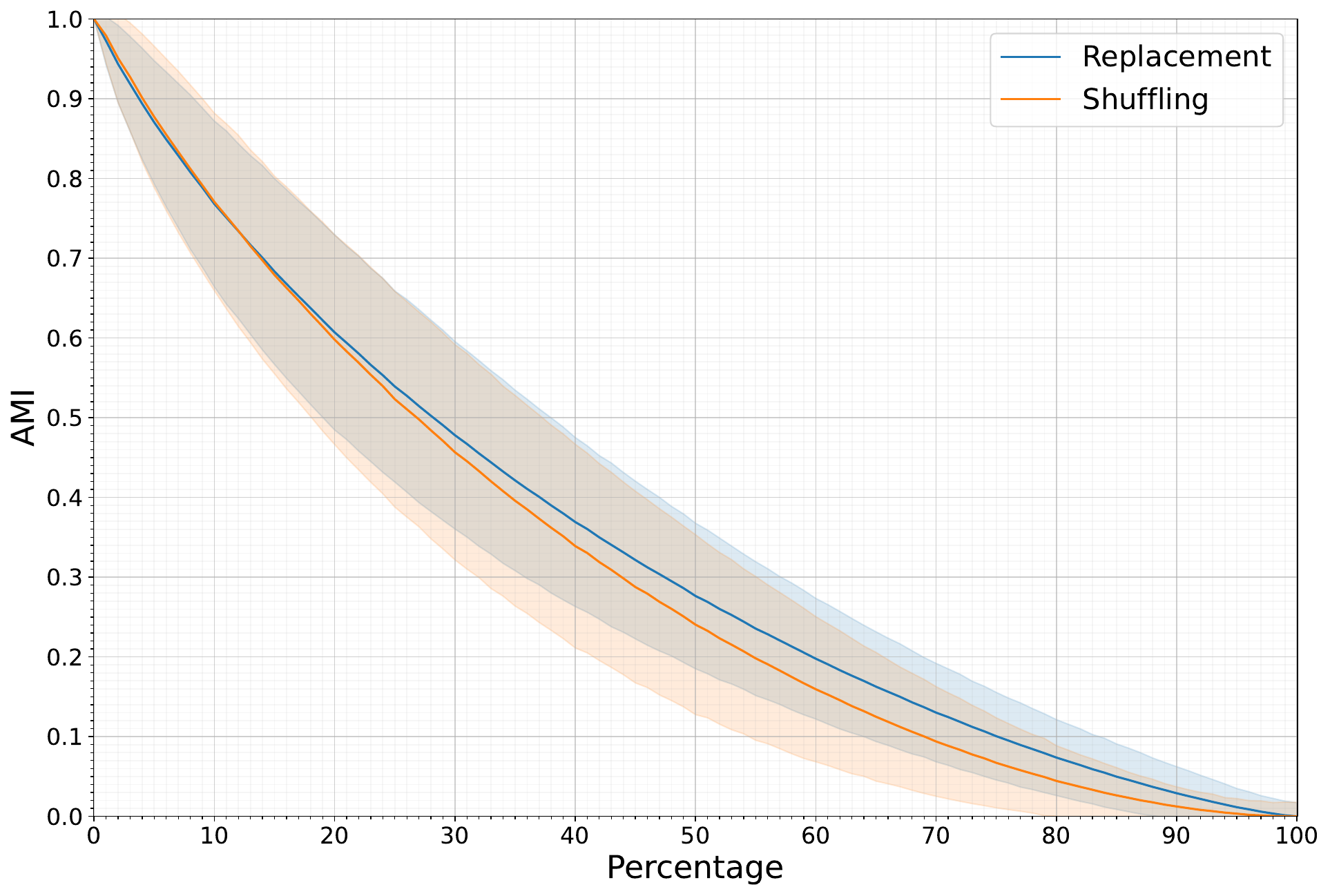} \\
        (a) ARI & (b) AMI \\[6pt]
    \end{tabular}
    \caption{For each dataset in the UCR Archive, a percentage of the ground-truth class labels were selected and either shuffled or replaced uniformly at random to create a new labelling. The figures plot the a) ARI and b) AMI values computed between these two partitions across all datasets and with 100 repetitions.}
    \label{Fig:Label_Shuffling}
\end{figure}

In \cite{Gagolewski2021}, the authors consider how well RVIs concur with expert knowledge as represented by ground truth labels. They generate a large number of partitions for each of their datasets and observed that some datasets were inherently hard to cluster, with no algorithm producing partitions that matched the ground truth labels well. They specifically refer to three datasets which had maximum ARIs of 0.036, 0.181 and 0.401. Moreover, as part of their phased evaluation framework, \cite{Javed2020} excluded datasets with an $\text{ARI}<0.05$ from winning tally counts, acknowledging that this was approaching randomness.

Whilst there is a multitude of ways that clustering partitions could score low ARIs, for some basic intuition consider \Cref{Fig:Label_Shuffling}, where ARI and AMI values have been computed for versions of the UCR ground truth labellings that have been tampered with to varying degrees. Multiple random subsets, whose size was determined according to the percentages shown on the $x$-axis, were selected from the ground truth labels and either shuffled or replaced with uniformly random labels without changing the number of clusters. 
Based on the observations in \Cref{Fig:Label_Shuffling}, an $\text{ARI}<0.05$ equates to over $80\%$ of the ground-truth labels being shuffled or replaced. We consider this threshold insufficient for our purposes, and instead suggest an ARI of 0.6 would be more appropriate. This would equate to no more than $25\%$ of the ground-truth labels having been shuffled or replaced, which is still a significant degradation of the ground-truth clustering.

\begin{figure}[!th]
    \centering
    \begin{subfigure}{0.25\textwidth}
        \centering
        \includegraphics[height=4.2cm]{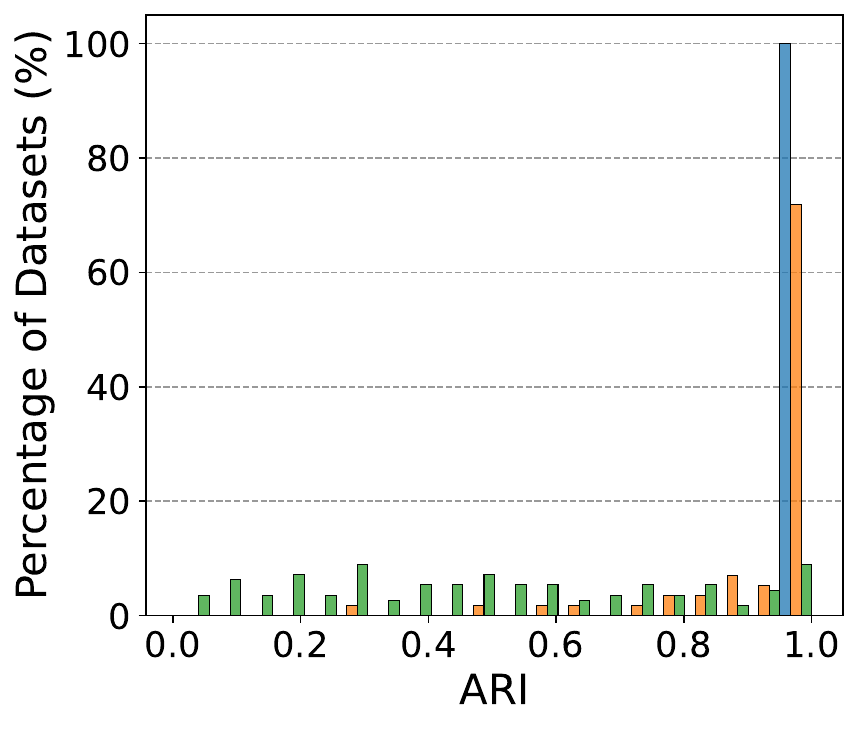}
        \caption{Maximum ARI}
        \label{Fig:ARI-Histograms-A}
    \end{subfigure}%
    \begin{subfigure}{0.75\textwidth}
        \centering
        \includegraphics[height=4.2cm]{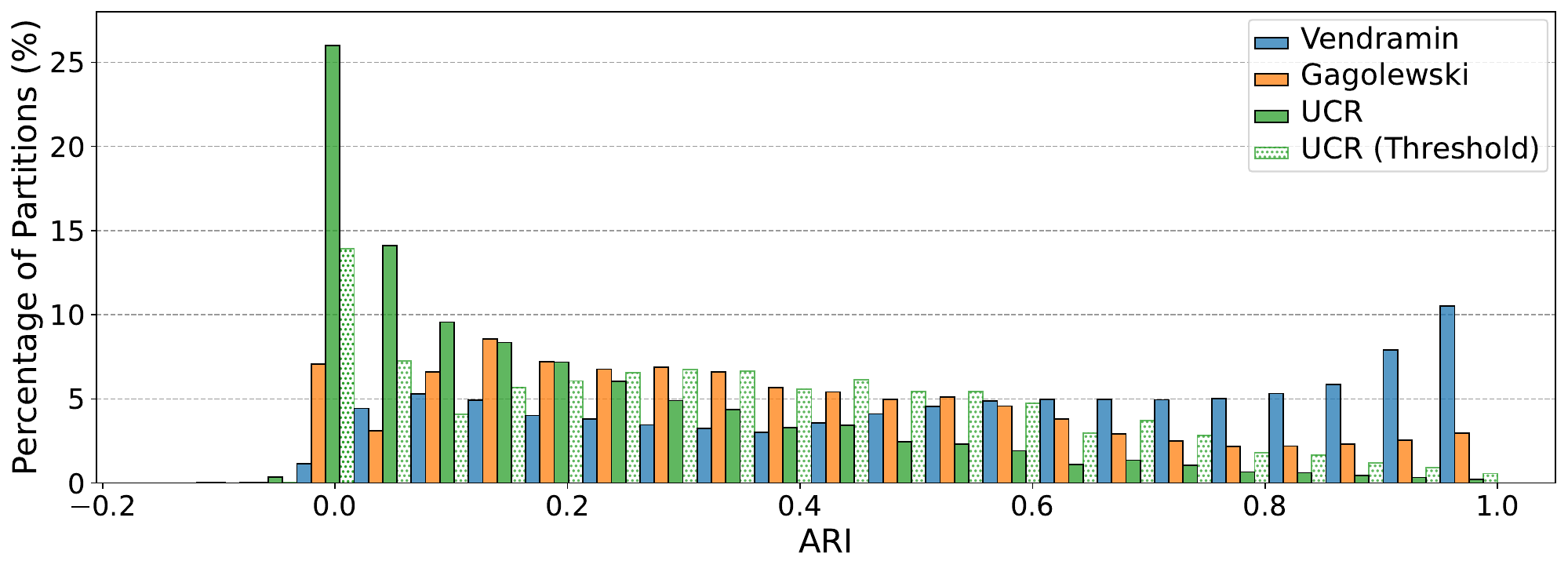}
        \caption{All ARI}
        \label{Fig:ARI-Histograms-B}
    \end{subfigure}
    \caption{Histograms of (a) the maximum ARI value per dataset, and (b) ARI values for every partition.}
    \label{Fig:ARI-Histograms}
\end{figure}

It should be noted that there is likely a difference in terms of the discovered structure between a partition achieving an ARI of 0.2 for a dataset where the best partition scored over 0.9, compared to a dataset where 0.2 is achieved by the best partition. The former can occur for instance where the ground-truth clusters have been split into a sufficient number of subclusters hierarchically. In the latter case it is likely that the ground-truth labels are not discoverable, or that there is no discoverable clustering structure to the data.
\Cref{Fig:ARI-Histograms-A} presents a histogram of the maximum ARI values achieved for each dataset from the three batteries. All apart from two of the 972 Vendramin datasets have at least one partition returning an ARI of 1.0. The two exceptions both have a maximum ARI of $0.996$. The Gagolewski battery is also mostly composed of datasets with recoverable labels, with at least one partition for $54/57$ datasets achieving a maximum ARI above 0.6, and the remaining three having maxima of 0.560, 0.261 and 0.460. As revealed in \Cref{Fig:Javed2020-Threshold_Plot}, the UCR battery is much less dependable. 

By applying a threshold requiring \textit{at least one} partition have an ARI in excess of 0.6 to the UCR battery, we obtain a subset of 41 datasets. The resulting distribution of ARI values for this subset is shown in \Cref{Fig:ARI-Histograms-B} alongside the other batteries. The ARI values of this subset still cover the same range as the full UCR battery, but is notably less right skewed. The coincidence and correlation analyses in \Cref{Subsec:Experimental_Results-CoincidenceRates,Subsec:Experimental_Results-Correlation} have been reproduced for this subset and are presented in \Cref{Fig:CoincidencePlots-UCR-K-Threshold,Fig:CoincidencePlots-UCR-SP-Threshold} and \Cref{Tab:UCR-Threshold-ARI-Pearson-All} respectively. 

As could be expected, the exclusion of datasets without clustering structure or non-discoverable labels results in improvements in the overall success rates. For the SP-selection success rates, it is noteworthy that the fixed-MSM scheme appears to pull further ahead of the other fixed schemes for most of the RVIs. This is likely due to the fact that MSM was responsible for more of the maximal ARI partitions ($32\%$) than the other distance measures ($17\%$ for the next closest). This means that the MSM SP is optimally capturing the similarity structure in more of the datasets. With most of the RVIs the fixed-MSM scheme also demonstrated a clear bias towards partitions produced with MSM for the UCR archive (\Cref{Fig:BiasPlots-UCR}).

Significant improvements to the median correlations were also observed for many combinations of RVIs and evaluation schemes. Notably, the $k$-selection median correlations for SWC\textsubscript{MSM} improved by 0.538. Interestingly, DBI, DI and PBM still showed extremely poor correlations for this task. These also happen to be three indices which were not encouraged for use with non-Euclidean data and distance measures (see Appendix \hyperref[Appendix:RVIs]{A}). 
The SP-selection median correlations for the mean-SP scheme improved more significantly than for the matching-SP scheme, bringing the relationship between the two schemes more in line with observations for the other two batteries.
	
	
	\section{Discussion}
\label{Sec:Discussion}

Our experiments were designed to probe for answers to three main questions:
\begin{enumerate}
    \item[(i)] Do RVIs equipped with a fixed-SP evaluation scheme demonstrate an observable bias towards partitions generated from clustering approaches employing the same SP?
    \item[(ii)] Is there a difference in reliability between the fixed-SP, matching-SP and mean-SP evaluation schemes for either of the SP- or $k$-selection tasks?
    \item[(iii)] Are RVIs as reliable for SP-selection as they are for their conventional application of $k$-selection?
\end{enumerate}

We now address each of these questions in turn, drawing on the experimental results presented in \Cref{Sec:Experimental_Results}.

\subsubsection*{(i) Bias in Fixed-SP Evaluation Schemes}
Our experimental results provide strong evidence for the existence of a bias effect in fixed-SP evaluation schemes. None of the seven RVIs conformed with the behaviour expected in the absence of bias, as demonstrated by their failure to produce distributions of OWM values symmetrically distributed around the expected value of $1/N_{\sigma}$. For all of the RVIs, the extent of this bias effect manifested non-uniformly across the different SPs. Notably, certain fixed-SP schemes consistently yielded the optimal value across the considered fixed-SP versions of some RVIs --- regardless of the SP used for clustering. This different sort of bias effect was most pronounced for the more unique SPs within the candidate set, such as the cosine distance in the Vendramin and Gagolewski batteries. This observation suggests that practitioners face greater bias effects when selecting between diverse SPs compared to choosing among more similar ones. Such bias amplification can lead to certain SPs being artificially favoured, regardless of their actual suitability to the data.
These findings strongly suggest that practitioners should avoid the use of fixed-SP evaluation schemes for SP-selection, particularly when comparing fundamentally different SPs.

\subsubsection*{(ii) Comparison of Evaluation Schemes}

For SP-selection, our analysis revealed no clear advantage in using any particular evaluation scheme. The matching-SP scheme consistently demonstrated similar or inferior success rates and correlations compared to fixed-SP schemes across all three batteries. The mean-SP scheme performed similarly, showing no systematic improvement over fixed-SP approaches.

The relative performance of different fixed-SP schemes for SP-selection varied based on the nature of the data. Recall that in \Cref{Sec:Whats_the_problem} it was suggested that a fixed-SP evaluation scheme would be effective for SP-selection only if the ideal SP was used to compute the RVI. This was borne out in our experimental results. For time series datasets in the UCR battery, fixed schemes using time series-specific SPs, such as MSM, outperformed those using generic feature-vector SPs like ED (the typical RVI default). Similarly, for the Vendramin and Gagolewski batteries, fixed schemes using SPs better-suited to the data tended to perform better --- for instance, the Canberra distance showed higher success rates for the Vendramin battery, aligning with its tendency to produce partitions with the largest ARI for these datasets. However, these marginal improvements are difficult to reconcile with the biases observed for fixed-SP schemes and the requirement to know which SP is best suited to the data a priori.

Regarding $k$-selection, we found that there was no obvious advantage when employing any of the three evaluation schemes. Thus, based on our discussion in \Cref{Sec:Whats_the_problem}, we suggest practitioners use the same SP for evaluation that was used for clustering when comparing partitions from different clustering algorithms with different numbers of clusters. This ensures that conclusions are consistent with the spatial embedding of the dataset used to obtain the partitions.

\subsubsection*{(iii) Reliability of RVIs for SP-selection}

RVIs were found to be much less reliable for SP-selection than $k$-selection on well-behaved datasets with well-separated and compact globular clusters, such as those found in the Vendramin battery. For these datasets, success rates for SP-selection were typically half or less of those for $k$-selection. Furthermore, an RVI's proficiency at $k$-selection on this battery did not predict proficiency in SP-selection, with some top-performing indices for $k$-selection showing markedly poor results in SP-selection. This was particularly evident for CHI and PBM, which performed exceptionally well for $k$-selection but showed markedly poor performance in SP-selection, with success rates dropping dramatically and correlations hovering around zero.

The findings weren't as clear for the batteries featuring datasets with more real-world complexity. For the Gagolewski and UCR batteries, the success rates and correlations on the $k$-selection task were drastically lower than those for the Vendramin battery. This observation aligns with known limitations of these seven traditional indices when encountering real-world data or complex datasets involving clusters with varying densities, overlap, or arbitrary shapes \cite{Arbelaitz2013AnIndices}. The success rates and correlations for SP-selection were largely consistent with those for $k$-selection on the Gagolewski battery, with some RVIs (SWC, DI, CHI) showing slightly better performance on $k$-selection and others (AUCC, CI) showing slightly better performance on SP-selection. Interestingly, the SP-selection task on the UCR battery revealed a contrasting pattern, with modestly higher success rates for most RVIs compared to $k$-selection. As previously mentioned, these improvements were particularly pronounced when using time series-specific SPs rather than generic feature-vector SPs. This advantage was also reflected in higher correlations for these cases. However, it is important to note that the absolute success rates for SP-selection on both the Gagolewski and UCR batteries remained quite low overall --- generally not exceeding 30\%.

\subsection{Recommendations}
\label{Subsec:Discussion-Recommendations}
The authors interpret these observations as strong evidence for the unsuitability of RVIs to the SP-selection task when using any of the evaluation schemes considered. A breadth of more appropriate, reliable tools and strategies exist for selecting an SP which will either be ``good enough" for an exploratory analysis, or optimal in some particular concrete sense. We have gathered some of these alternatives that circulate in the literature and will discuss them in the remainder of this section.

\subsubsection*{Matching Data Characteristics}
This first alternative involves selecting a subset of relevant SPs whose features and characteristics align with those of the dataset and the domain under analysis. This is an obvious first step for any practitioner wanting to effectively cluster their data. Whilst this approach will likely not prove sufficient to make a final selection of one SP, it is useful for narrowing the field of candidates. For example,
a Shapelet representation \cite{Zakaria2012} may be appropriate where time series contain extraneous data, are of variable lengths or have missing data. However, Shapelets are less likely to be effective in situations where features or events should be aligned within the time domain. Coordinating dataset and SP characteristics can also be useful for reducing the range of candidate parameters for a given representation or distance. For instance, if it is important that the clustering process focus on events closely aligned in time, then a heavily constrained version of DTW may be appropriate with a small warping window. 

Each dataset or clustering problem will come with a unique set of characteristics, and if these are considered in conjunction with the requirements of the clustering, the range of candidate SPs and their parameters can be reasonably reduced. Some research has attempted to formalise this relationship between datasets and specifically clustering \textit{algorithms} \cite{Ferrari2015,Pimentel2019,Jilling2020} by drawing upon the meta-learning literature \cite{Smith-Miles2008}. Meta-learning uses dataset-level features to learn a mapping between datasets and algorithms, allowing for the automatic recommendation of appropriate algorithms for unseen datasets on a range of different data mining tasks. The authors are not aware of any research that has attempted to define meta-features for the recommendation of different similarity paradigms for clustering. The breadth of datasets, and methods, that would be required to train a useful, widely applicable meta-model make this approach to SP-selection impractical without a concerted effort in this direction. 

\subsubsection*{Visualisation}
An obvious second alternative that can be useful for comparing SPs is visualisation. This alternative is especially practical for time series data --- albeit limited in scope by the length and number of time series. Dimensionality reduction techniques such as PCA, tSNE and MDS can also make visualisation possible for high-dimensional feature-vector data, though they are prone to information loss, with preservation of clustering structures not guaranteed \cite{Moon2019}. Furthermore, such visualisations can quickly become overwhelming for SP-selection if the number of competing methods is large. Especially if $k$ is unknown and multiple clustering algorithms are being considered. The visual assessment of clustering tendency \cite{Kumar2020a,Wang2010a,Bezdek2002} is another form of visual evaluation which can be useful for determining whether a candidate SP has been able to capture any clustering structure in a dataset.

\subsubsection*{External Validation with High Quality, Application Specific Labeled Datasets}
The third alternative involves comparing SPs using external validation against labelled datasets. Strong performance in a benchmark study (such as \cite{Javed2020}) across datasets from a multitude of domains is certainly grounds to consider an SP as a candidate, but it should not be assumed that this performance will necessarily extrapolate to novel datasets \cite{Wolpert1997}. This approach specifically requires high quality, domain and problem specific labelled datasets, which may be real or synthetic. This calibre of labelled dataset is typically not readily available, and can be difficult or expensive to procure, often requiring the involvement of domain experts. Real world datasets could be labelled fully by hand, or partial labellings could be propagated through the dataset. External validation against such datasets can result in a useful selection bias towards SPs that align with those cluster concepts included by the expert. However this could also lead to clustering approaches that will overlook some fruitful, unanticipated groupings in new datasets. In lieu of real labelled datasets, synthetically generated datasets can also be effective \cite{Hennig2018}. In this case, there is a balance that needs to be struck between realistic complexity and reliability of cluster labels \cite{Kim2020a}. Conditional generative adversarial networks are a promising recent advancement that can be adapted for the production of credible synthetic labelled feature vector and time series datasets \cite{Zhang2018GenerativeGrids, Smith2020, Pinceti2021}.

\subsubsection*{Application Targeted Evaluation}
The final alternative we consider is born of a clustering philosophy which differs subtly from the typical exploratory philosophy. This philosophy declares that clustering methodologies cannot be evaluated effectively in isolation from some clustering objective \cite{Hennig2015a,Montero2014TSclust:Clustering}. It was noted in \cite{Li2022} that when selecting distance measures for clustering, optimal performance according to internal validation with RVIs doesn't necessarily translate to optimal performance in downstream application tasks. This idea is over 100 years old \cite{VonLuxburg2012}, as \cite{Mercier1912ALogic} claims: ``The nature of the classification that we make \ldots\ must have direct regard to the purpose for which the classification is required. In as far as it serves the purpose, the classification is a good classification, however `artificial' it may be. In as far as it does not serve this purpose, it is a bad classification, however `natural' it may be." 

The authors of \cite{VonLuxburg2012} identify that much of the disagreement and uncertainty around clustering evaluation and comparison is a side-effect of the diversity of problems for which we rely on clustering. They advocate for the construction of a taxonomy of clustering \textit{problems}. For example, one clustering problem might be to define categories for subsequent use in a data processing pipeline, while another might be to validate a hypothesis about groupings in the data, and yet another could be to model the data generating processes for groups of objects. They argue that these different problems will require their own domain-dependent evaluation procedures and that interactions between statisticians and domain experts cannot be overlooked in this process. 

This philosophy of application specificity can modestly guide evaluation \cite{Hennig2019ClusterUser}, as is the case for \cite{Iglesias2013}: ``Our intended applications mainly use clustering for pattern or representative discovery, so we find suitable validity methods that focus on representativeness or give an important role to the representatives." In the extreme, this philosophy has guided the creation of novel optimisation criteria. One such example comes from the interval meter time series clustering literature, where indirect observations of the error on downstream forecasting applications are used as an indication of relative clustering quality \cite{Chaouch2014,Teeraratkul2018,Tajeuna2018}. For other examples of novel evaluation criteria, see \cite{Dent2015,Toussaint2020}. 

Despite the obvious benefit of specificity providing concrete grounds for comparison between SPs, this approach still suffers from varying degrees of subjectivity, sensitivity and expert bias. Furthermore, not every application of clustering has goals that translate so concretely into objective functions. However, we encourage practitioners to examine whether their specific application could allow for such tangible evaluation criteria.
	
	
	\section{Conclusion}
\label{Sec:Conclusion}

Relative validity indices are the most widely used tool for clustering evaluation, but it is critical to understand their limitations. Originally suggested as an aid for selecting the optimal partition from amongst a set which vary by the number of clusters or the responsible clustering algorithm, RVIs have increasingly been used to compare the clustering performance of different similarity paradigms --- that is, to select optimal distance measures, representation methods, normalisation procedures, and their parameterisations. Our comprehensive investigation demonstrates that RVIs are not well-suited for this task, a finding that should not be surprising given that it was never their intended purpose. This finding is particularly significant given the widespread and often unchallenged use of RVIs for this purpose in current research.

Through extensive experimentation involving over 2.7 million clustering partitions, we have established several crucial findings regarding the use of RVIs for SP-selection. Firstly, fixed-SP evaluation schemes exhibit clear bias towards partitions generated using the same SP, with this effect amplified when comparing SPs with fundamentally different formulations. Secondly, RVIs demonstrate lower reliability for SP-selection compared to $k$-selection, with some top-performing indices for $k$-selection showing the poorest results for SP-selection. Thirdly, matching-SP and mean-SP evaluation schemes not only fail to offer meaningful advantages over fixed-SP schemes, but bear their own limitations. These observations should not be viewed as isolated to the seven RVIs examined in this study; rather, they likely extend to other RVIs given the fundamental nature of the limitations identified. The success of RVIs in choosing the best SP should be viewed as an exception, rather than the rule.

Based on these findings, we recommend that RVIs only be used to rank or select optimal partitions from collections produced using the same SP --- where the partitions in the collection may differ only by their numbers of clusters or the responsible clustering algorithm. For this purpose our results suggest that the choice between fixed Euclidean and matching evaluation schemes is unlikely to significantly impact performance, as both performed similarly in our experiments for $k$-selection across all batteries. Nevertheless, for more meaningful evaluations we recommend using the matching-SP scheme to ensure that the RVI is exposed to the same similarity structure used to obtain the partitions.

Our main recommended alternatives for SP-selection are necessarily application-specific, aligning with the long-standing but often overlooked principle that clustering evaluation cannot be meaningfully separated from its intended purpose. The caveat with these alternatives is that they require more effort from practitioners, though we believe that this additional complexity is warranted given the demonstrated unreliability of RVI-based approaches. These alternatives include: carefully selecting SPs based on dataset characteristics and domain requirements; using visualisation techniques where feasible; validating against high quality, domain-specific labelled data; and evaluating SPs based on concrete application objectives. While none of these alternatives offers the simplicity of RVI-based evaluation, they provide more reliable and meaningful paths to SP-selection. We particularly emphasise the importance of considering the specific purpose motivating the use of clustering, as the optimal SP for one application may not be suitable for another, even within the same domain.

We also advocate for the adoption of some standard terminology to describe clustering approaches which may combine different normalisation procedures, representation methods, distance measures, clustering algorithms and prototype definitions. Without distinguishing between a clustering algorithm and a clustering approach, it is all too easy for practitioners to fall for the pernicious misconception exemplified in \cite{Anzanello2011} where it is suggested that the SWC is ``independent of the clustering technique". This is true for clustering algorithms, but not of clustering approaches, where the SP can vary. Similarly, where ambiguous, we encourage practitioners to explicitly state which SP was used when computing RVIs for $k$-selection --- either the default Euclidean or the matching SP. For those who persist with using RVIs for SP selection, we urge them to clearly indicate whether a fixed-SP or matching-SP evaluation scheme was employed. Such transparency would also be advisable for software packages offering RVI implementations.

This work should not be viewed as a criticism of previous research that has used RVIs for SP comparison. Instead, it serves as a careful examination of a widely-adopted practice. While it was outside the scope of the current study to develop novel methodologies, we hope our findings will inspire the research community to pursue more robust approaches to SP-selection, moving beyond the current reliance on tools that are inadequate for this critical task.


        \appendix
\crefalias{section}{appendix}
\renewcommand{\thesection}{Appendix \Alph{section}.}
\section{Relative Validity Indices}\label{Appendix:RVIs}

In this section we provide details regarding the computation of each RVI used within our experiments and explore their original scope and intended purposes. As mentioned in \Cref{Subsec:RVIs}, the RVIs have been split into two categories: prototype-sensitive and prototype-insensitive.

\subsection*{Notation}
First we shall establish some notation. Suppose we have obtained a \textit{hard} partition of a set $\bm{\mathcal{S}} = \left \{ \bm{x}_1, \bm{x}_2, \ldots, \bm{x}_N \right \}$, given by $\bm{\mathcal{C}} = \left \{ C_1, C_2, ..., C_k \right \}$. Whilst the $\bm{x}_i$ could represent any generic object to be clustered (for which we can define a dissimilarity function), we consider $\bm{x}_i \in \mathbb{R}^{n}$ as the data encountered in this paper was restricted to feature vectors and time series. Note that the hard partition $\bm{\mathcal{C}}$ satisfies $\bigcup_{j=1}^k C_j = \bm{\mathcal{S}}$, $C_j \neq \emptyset$ and $C_j \cap C_\ell = \emptyset$ for $j \neq \ell$. Furthermore, let $\bm{c}_j$ denote the prototype of cluster $C_j$ and $\bm{c}$ denote the prototype of $\bm{\mathcal{S}}$, i.e. the grand prototype. Note also that $|C_j|$ represents the number of objects in $C_j$, also known as the cardinality of the set $C_j$. 

These RVIs all require some computation of distances either between objects, between prototypes or between objects and prototypes. In the experiments, many different dissimilarity measures were used for this purpose. For the sake of consistency, the notation $d(\cdot, \cdot)$ will be used to represent any generic distance measure between the $\bm{x}_i$. These distances will not necessarily fulfil all of the conditions of a strict metric, in particular for time series distance measures where the triangle inequality or identity of indiscernibles are often violated.

\subsection*{Prototype-Insensitive RVIs}

\paragraph{Silhouette Width Criterion (SWC)}
Probably the most popular RVI, the SWC is given by a summation of individual scores computed for each object being clustered. Compactness is quantified by considering the average distance of each object to all of the other objects in the same cluster. Similarly, separation is quantified as the average distance of each object to all of the objects in the nearest cluster. Formally, consider $\bm{x}_i \in C_j$, and define $a_{j,i}$ as the average distance of $\bm{x}_i$ to every other object in $C_j$. Also define $b_{j,i} = \min_{\ell, \ell \neq j} d_{\ell,i}$, where $d_{\ell,i}$ is the average distance of $\bm{x}_i$ to every other object in a \textit{different} cluster $C_\ell$. The Silhouette score for $\bm{x}_i$ is then computed as the normalised difference between these two average distances,
\begin{equation*}
   s_{\bm{x}_i} = \frac{b_{j,i} - a_{j,i}}{\max \{a_{j,i}, b_{j,i}\}} \, .
   \label{Eqn:SWC_1}
\end{equation*}
These scores can be considered per object or per cluster, as is the case for Silhouette plots, or can be averaged over all $\bm{x}_i \in \bm{\mathcal{S}}$ to produce,
\begin{equation*}
   SWC = \frac{1}{N} \sum \limits_{i=1}^{N} s_{\bm{x}_i} \, .
   \label{Eqn:SWC_2}
\end{equation*}
The SWC is a maximisation criterion, as $b_{j,i}$ should be maximised and $a_{j,i}$ should be minimised. Further, if $C_j$ is a singleton cluster, i.e. $|C_j| = 1$, $s_{\bm{x}_i} := 0$ in order to prevent the trivial solution of $k = N$ being recommended. Whilst the original SWC is prototype-insensitive, some variants are not. A popular variant known as the simplified SWC \cite{Hruschka2009} replaces the computation of average distances between objects with distances between objects and relevant cluster prototypes.

In \cite{Kaufman1990FindingAnalysis} it is offered that SWC values above 0.7 suggest the discovery of strong clusters, values from $0.51 - 0.7$ reasonable clusters, $0.26-0.5$ weak or artifical clusters and that no clusters have been found for values less than 0.25. In practice however, these ranges should be interpreted with caution. One reason is that a homogeneous dataset with a single far enough outlier could be regarded as a strong clustering structure. Furthermore, recall that it was suggested in \Cref{Sec:Whats_the_problem} that RVIs based on unique SPs should be recognised as independent statistics. Such proposed ranges are thus virtually meaningless when comparing partitions produced with different SPs, as the exact same partition could have a SWC above 0.7 and below 0.5 according to different SPs (see the Plane dataset in \Cref{Fig:UCR_Different_PerfectPartition}).

In the original paper \cite{Rousseeuw1987Silhouettes:Analysis}, it was suggested that the SWC could suitably be applied to any ratio-scale dissimilarities, with Euclidean distance used as an example. This preserves the invariance of the SWC to multiplicative scaling of the pairwise distance matrix. It is also touted as an advantage that the SWC only depends on the dissimilarities and the partition. As a result of this, the authors suggest that the SWC is suitable for comparing partitions produced by different algorithms, with different numbers of clusters.

\paragraph{Dunn Index (DI)}\label{Sec:DI}
A popular index which takes the general form:
\begin{equation*}
    DI = \frac{\min \limits_{j \neq \ell \in \{1,\ldots,k\}} \delta_{j,\ell}}{\max \limits_{j \in \{1,\ldots,k\}} \Delta_j} \, ,
    \label{Eqn:DI_general_form}
\end{equation*}
where $\delta_{j,\ell}$ is a set distance between clusters $j$ and $\ell$, and $\Delta_j$ is the diameter of the $j^\textsuperscript{th}$ cluster. Originally $\delta_{j,\ell}$ was defined as the distance between the closest two objects in $C_j$ and $C_{\ell}$, and $\Delta_j$ was defined as the distance between the furthest objects in $C_j$, i.e.
\begin{equation*}
    \delta_{j,\ell} = \min \limits_{\bm{x}_h \in C_j, \,\bm{x}_i \in C_{\ell}} \left\{ d \left (\bm{x}_h, \bm{x}_i \right ) \right\}, \quad \Delta_j = \max \limits_{\bm{x}_h \in C_j, \,\bm{x}_i \in C_{j}} \left\{ d \left (\bm{x}_h, \bm{x}_i \right ) \right\}
    \label{Eqn;DI_components}
\end{equation*}

Many variants of the DI exist which involve different definitions of the set distance and diameter in order to correct for sensitivity to outliers \cite{Bezdek1995, Vendramin2010RelativeOverview}. This particular set distance is actually the single linkage definition of set distance commonly encountered in agglomerative hierarchical clustering. The set distance is a measure of separation between any two clusters, while diameter is a measure of cluster compactness. The ratio of these quantities means that the DI is also a maximisation criterion.

In the original paper \cite{Dunn1974Well-separatedPartitions}, the data were assumed to derive from a general real inner product space with an induced metric. The index is suggested for determining the optimal partition from a family of ``$k$-partitions" of the same form as $\bm{\mathcal{C}}$. This index has since been commonly applied for comparing partitions produced by
different algorithms, with different numbers of clusters \cite{Bezdek1995}.

\paragraph{C-Index (CI)}\label{Appendix:CIX}
The CI compares the sum of all within-group distances ($\theta$) with a worst and best case scenario. If there is a total of $\omega = \frac{1}{2}\sum_{j=1}^k |C_j| \left( |C_j| -1 \right)$ within-group distances, these two extremes are computed by sorting, in ascending order, all of the non-trivial pairwise distances and summing the first and last $\omega$ distances for the best ($\min \theta$) and worst ($\max \theta$) cases respectively. The index is then computed as
\begin{equation*}
    CI = \frac{\theta - \min \theta}{\max \theta - \min \theta} \, .
    \label{Eqn:CI_1}
\end{equation*}
Note that $\theta$ is the sum of all off-diagonal terms in the upper triangle of the pairwise distance matrix which correspond to objects that share a cluster, i.e.
\begin{equation*}
    \theta = \sum \limits_{h=1}^{N-1} \sum \limits_{i=h+1}^N I(h,i) \cdot d\left(\bm{x}_h, \bm{x}_i \right) ,
    \label{Eqn:CI_2}
\end{equation*}
where
\begin{equation*}
    I(h,i) = \begin{cases}1~&{\text{ if }}~\bm{x}_h\text{ and }\bm{x}_i\text{ are in the same cluster},\\0~&{\text{ if }}~\bm{x}_h\text{ and }\bm{x}_i\text{ are in different clusters}.\end{cases}
    \label{Eqn:CI_3}
\end{equation*}
According to the CI, better partitions will produce values of $\theta$ closer to $\min \theta$, hence the CI is a minimisation criterion.

From \cite{Bezdek2016TheValidity}: ``The C-index...was introduced in 1970 as a way to define and identify a \textit{best} crisp partition on $n$ objects..." The original paper \cite{Dalrymple-Alford1970MeasurementRecall} suggests the CI in the context of free-recall clustering. The theory and application space of this index was not addressed in this paper. Rather it was subsequently discussed in \cite{Hubert1976ARecall}, \cite{Hubert1985} and \cite{Bezdek2016TheValidity}. The CI readily accepts a generic distance matrix, though formalisations are often prepared with the assumption of an induced norm in a vector space \cite{Bezdek2016TheValidity}. It is implied that the CI is capable of comparing partitions with different numbers of clusters, but it is not explicitly suggested that partitions from different algorithms should or should not be compared.

\paragraph{Area Under Curve for Clustering (AUCC)}\label{Appendix:AUCC}
The AUCC \cite{Jaskowiak2022} is a linear transformation of Baker and Hubert's Gamma criterion \cite{Baker1975} with a superior computational complexity. It is also the only RVI we are aware of which has incorporated a theoretical expected value for random clusterings - something typically only accounted for in the context of external clustering evaluation. The AUCC makes use of the Area Under the Receiver Operating Characteristics Curve (AUC-ROC), a popular supervised learning performance measure. By representing a clustering partition as a binary relation between objects, and normalising the pairwise distance matrix to range between $0$ and $1$, the clustering problem can be recast in a format compatible with a typical ROC analysis. The normalised dissimilarity values correspond to classification thresholds, and the corresponding binary labels are treated as the \textit{true} classes. AUCC is thus a maximisation criterion ranging between 0 and 1. The expected value under a relevant null model of random clusterings is 0.5, regardless of $k$ or cluster imbalances. It should also be noted that solutions under-estimating $k$ are penalised more heavily than solutions over-estimating $k$. 

In the original paper \cite{Jaskowiak2022}, it is suggested that the AUCC is defined for objects embedded in a space where a dissimilarity measure between objects can be defined which adheres to non-negativity, identity of indiscernibles and symmetry. Interestingly, the authors also suggest that the dissimilarity used to compute the AUCC ``\textit{must} be the very same (or equivalent) to the one employed during the clustering phase". The experiments performed in \cite{Jaskowiak2022} suggest the authors support the use of the AUCC for comparing partitions produced by different clustering algorithms, with different numbers of clusters.

\subsubsection*{Prototype-Sensitive RVIs}

\paragraph{Calinski-Harabasz Index (CHI)}\label{Appendix:CHI}
This index is also referred to as the Variance Ratio Criterion (VRC), and is computed as,
\begin{equation*}
    CHI = \frac{\Tr{(\mathbf{B})}}{\Tr{(\mathbf{W})}} \times \frac{N-k}{k-1} \, , 
\end{equation*}
where $\Tr(\cdot)$ is the trace of a matrix, and $\mathbf{B}$ and $\mathbf{W}$ are the $n \times n$ between-cluster and within-cluster dispersion matrices respectively. Separation is quantified as the sum of between-cluster variances, given by $\Tr{(B)}$, while the sum of within-cluster variances, given by $\Tr{(W)}$, measures compactness. Hence this is a maximisation criterion. The normalisation term prevents the index from monotonically increasing with $k$. Computing these traces is done in the original formulation without obtaining the full matrices by using the following:
\begin{equation*}
    \Tr{(\mathbf{B})} = \Tr{(\mathbf{T})} - \Tr{(\mathbf{W})} \, , \quad \Tr{(\mathbf{T})} = \sum \limits_{i=1}^{N} d(\bm{x}_i, \bm{c})^p \, , \quad \Tr{(\mathbf{W})} = \sum \limits_{j=1}^{k} \sum \limits_{\bm{x}_i \in C_j} d(\bm{x}_i, \bm{c}_j)^p \, ,
\end{equation*}
where $p=2$ and $d(\cdot,\cdot)$ is the ED. In order to make the CHI generically applicable to non-Euclidean measures, we have used a variant with $p=1$. This is consistent with the presentation of this option to users in the \texttt{WeightedCluster} package \cite{Studer2013} implementation for \Rlogo, and the formulation presented for community mining evaluation in \cite{Rabbany2012RelativeAlgorithms}.

In the original paper \cite{Calinski1974AAnalysis}, it is acknowledged that their proposed method could be extended to instances where the points are not from an ordinary Euclidean space. If only a measure of pairwise distances between objects can be established, then the objects can be meaningfully subjected to cluster analysis with the VRC. It is not explicitly suggested that the CHI can be used to compare partitions from different algorithms, but it is recommended as an informal indicator of the ``best number" of clusters. They acknowledge that this is in spite of any satisfactory probabilistic theory justifying such a use, and rather due to the CHI displaying desirable mathematical properties. The authors pragmatically suggest choosing the value of $k$ where the CHI demonstrates either an absolute maxima, early local maxima or at least a comparatively rapid increase.

\paragraph{Davies-Bouldin Index (DBI)} \label{Appendix:DBI}
Where the CHI is a ratio of sums, the DBI is a sum of ratios. It is defined as follows,
\begin{equation*}
    DBI = \frac{1}{k} \sum \limits_{j=1}^{k} D_j,
\end{equation*}
where $D_j = \max_{j \neq \ell} \left \{D_{j,\ell} \right \}$, $D_{j,\ell} = \left (\bar{d}_j+ \bar{d}_\ell \right )/d_{j,\ell}$ is the within-to-between cluster spread for clusters $C_j$ and $C_\ell$, $\bar{d}_j$ and $\bar{d}_\ell$ are the average within-cluster distances of $C_j$ and $C_\ell$ respectively, and $d_{j,\ell}$ is the distance between those two clusters. Explicitly,
\begin{align}
    \bar{d}_j = \frac{1}{|C_j|} \sum \limits_{\bm{x}_i \in C_j} d(\bm{x}_i, \bm{c}_j),\, \text{ and } \,\,d_{j,\ell} = d \left ( \bm{c}_j, \bm{c}_\ell \right ) \, .
    \label{Eqn:DBI_Dist_to_and_between_prototypes}
\end{align}
The smaller the values of $D_j$, the smaller the worst within-to-between cluster spread involving cluster $C_\ell$, hence DBI is a minimisation criterion.

In the original paper \cite{Davies1979AMeasure}, the DBI is proposed in terms of the Minkowski family of distance metrics. The authors explicitly state that the DBI ``can be used to compare the validity of data partitions regardless of how those partitions were generated." Subsequently they only ever refer to partitions varying in terms of clustering algorithms and numbers of clusters. It is unlikely that they were advocating for the DBI to be used to compare partitions from different SPs.

\paragraph{Pakhira-Bandyopadhyay-Mauli Index (PBM)} \label{Appendix:PBM} 
The PBM for a partition is given by,
\begin{equation}
    PBM = \left ( \frac{1}{k} \frac{D_K}{E_K} E_1 \right) ^p \, .
    \label{Eqn:PBM}
\end{equation}
PBM uses the maximum distance between cluster prototypes to measure the separation of the clustering, given by $D_K = \max_{j,\ell \in \left \{ 1,2,\ldots,k \right \} } d(\bm{c}_j,\bm{c}_\ell)$. Cluster compactness is summarised by the $E_k$ term, which denotes the sum of distances to each object's cluster prototype, i.e. $E_K = \sum_{j=1}^{k} \sum_{\bm{x}_i \in C_j} d(\bm{x}_i, \bm{c}_j)$. 
$E_1$ denotes the sum of distances to the grand prototype, i.e. $E_1 = \sum_{i=1}^{N} d(\bm{x}_i,\bm{c})$. In the original paper, $p=2$, but as for CHI, we have chosen to set $p=1$ for general applicability to non-Euclidean distance measures.
As $D_K$ should be maximised and $E_K$ minimised, PBM is another maximisation criteria.

The original paper \cite{Pakhira2004ValidityClusters} suggests the index is capable of selecting the optimal number of clusters for an ``underlying clustering technique".
The index is explained and decisions justified with reference to spherical cluster concepts in Euclidean space. A vector norm notation is also utilised without any indication that this could be extended to other distance measures. 
It should be noted that PBM is the only index we considered that is not scale-invariant, and this is due to the normalisation by $E_1$. This term was dropped in \cite{Rabbany2012RelativeAlgorithms}, but has been retained here to be faithful to the original formulation, and to observe the performance of a scale-variant RVI. Whilst invariance was expressly advertised for the SWC, all of DI, CI, AUCC, CHI and DBI are also invariant to multiplicative scaling.
 

	\bibliographystyle{acm}
	\bibliography{references}








\newpage
\section*{Supplementary Materials}

\subsection*{Coincidence of EVI and RVI Optima}
\begin{figure}[!htb]
\centering
    \begin{subfigure}[t]{0.49\textwidth}
        \centering
        \includegraphics[width=\textwidth]{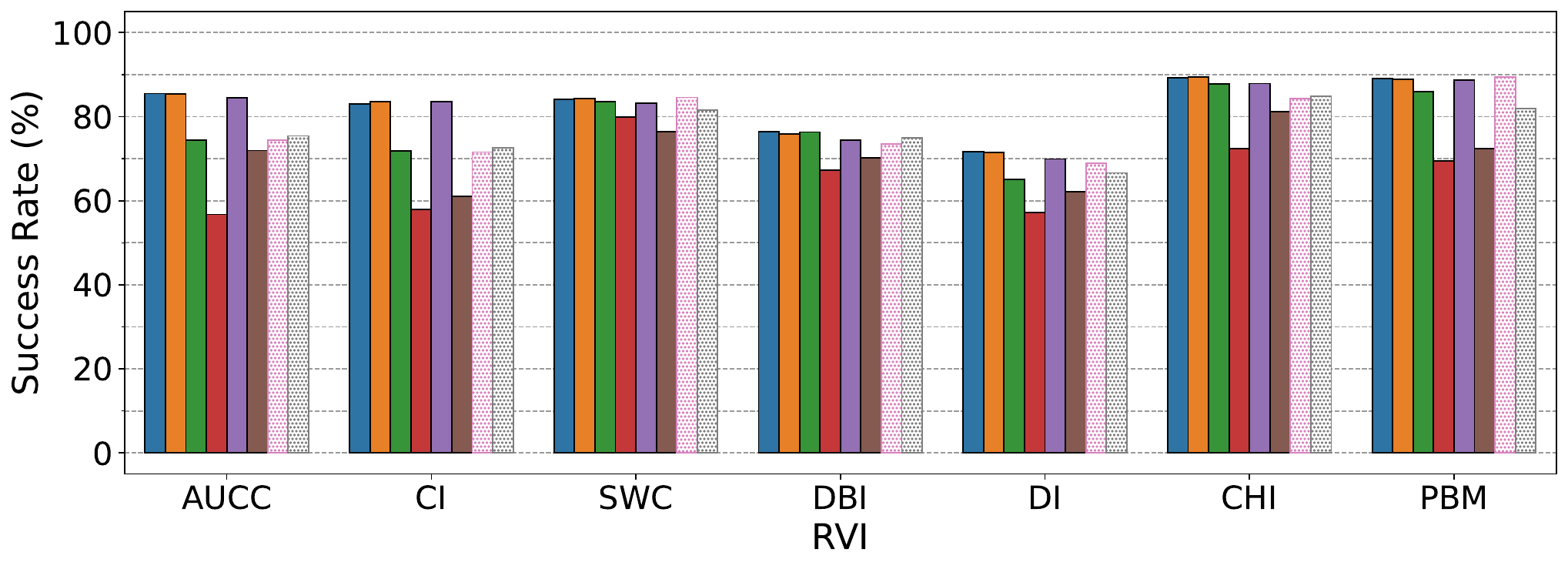} 
        \caption{Vendramin (over $k$)}
        \label{Fig:CoincidencePlots-Vend-K-ami}
    \end{subfigure}%
    ~
    \begin{subfigure}[t]{0.49\textwidth}
        \centering
        \includegraphics[width=\textwidth]{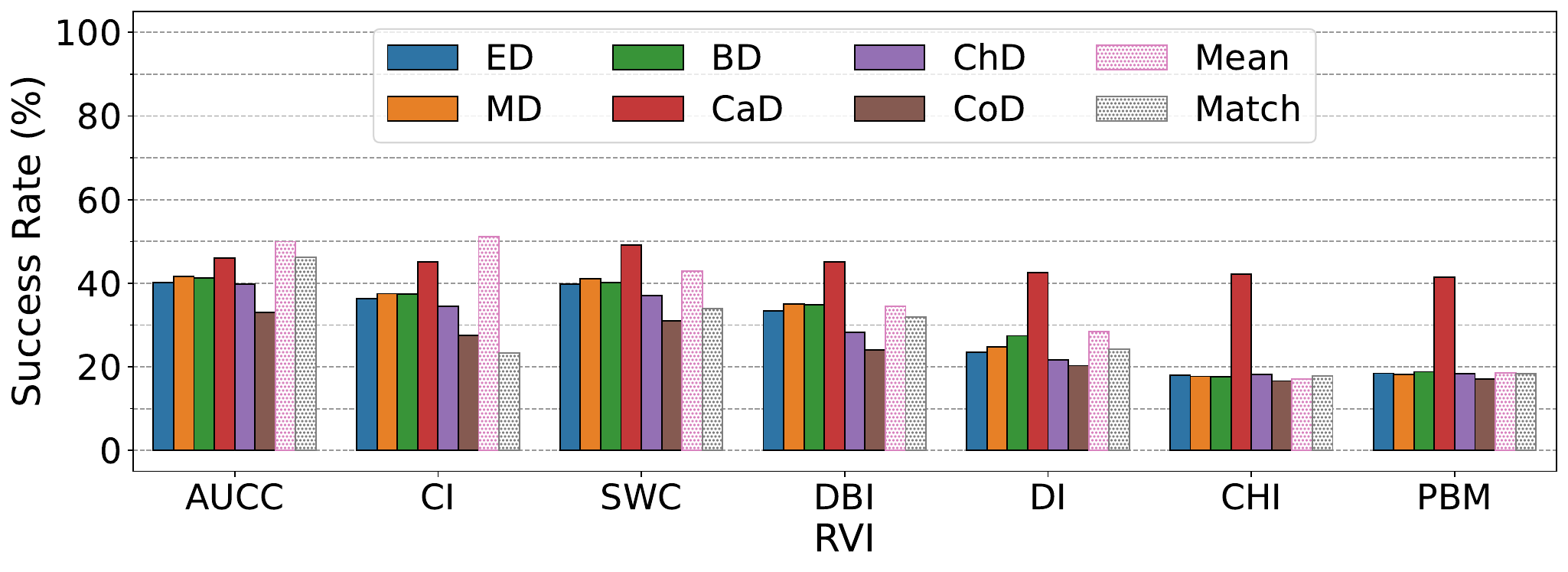}
        \caption{Vendramin (over SPs)}
        \label{Fig:CoincidencePlots-Vend-SP-ami}
    \end{subfigure}
    ~
    \begin{subfigure}[t]{0.49\textwidth}
        \centering
        \includegraphics[width=\textwidth]{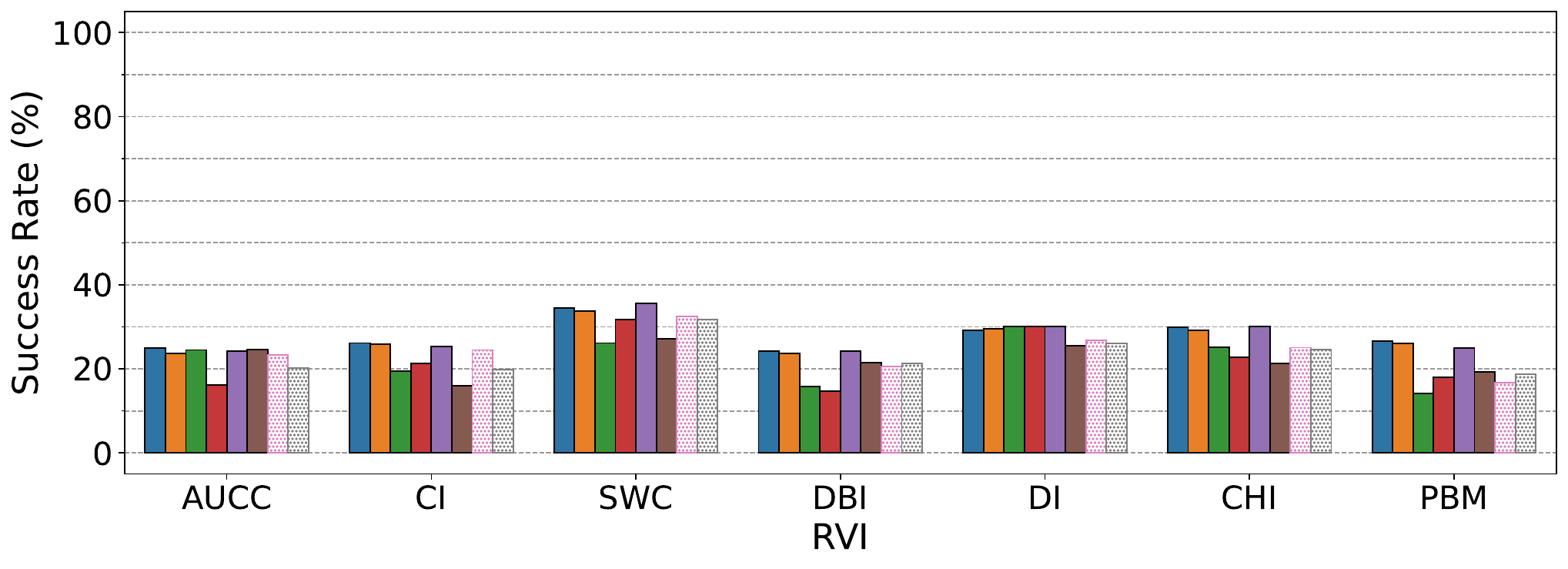} 
        \caption{Gagolewski (over $k$)}
        \label{Fig:CoincidencePlots-Gag-K-ami}
    \end{subfigure}%
    ~
    \begin{subfigure}[t]{0.49\textwidth}
        \centering
        \includegraphics[width=\textwidth]{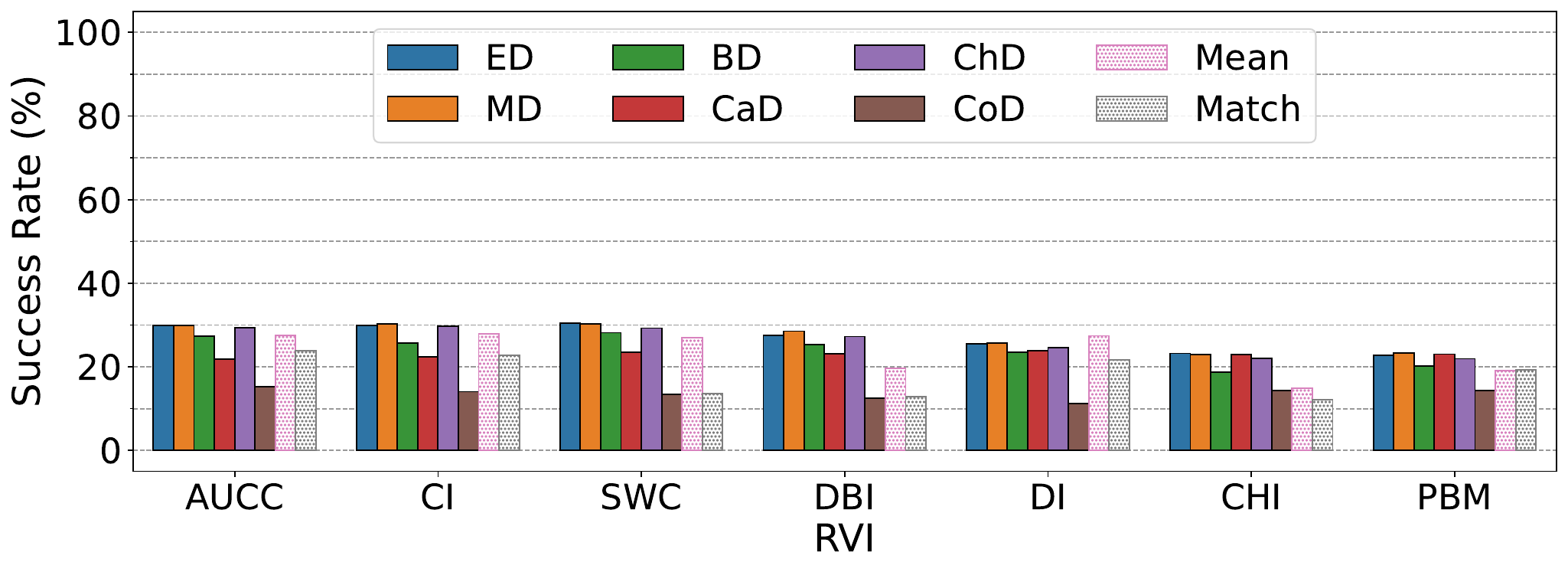}
        \caption{Gagolewski (over SPs)}
        \label{Fig:CoincidencePlots-Gag-SP-ami}
    \end{subfigure}
       ~
    \begin{subfigure}[t]{0.49\textwidth}
        \centering
        \includegraphics[width=\textwidth]{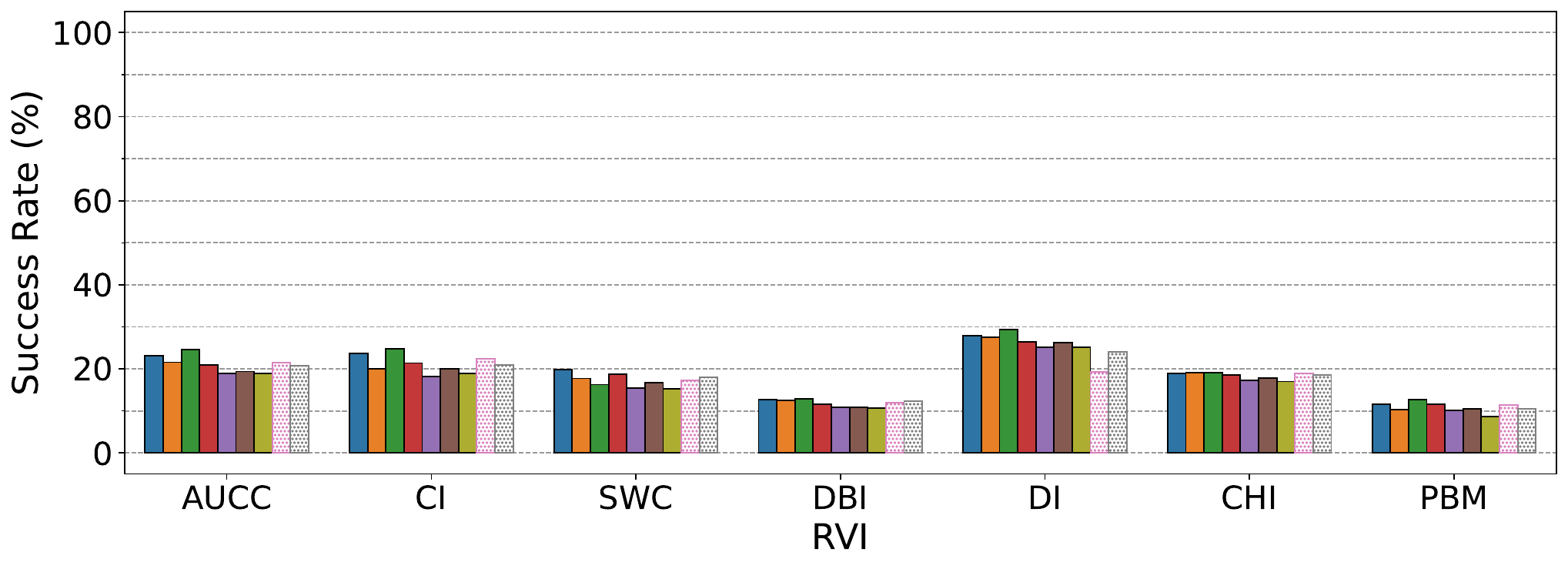} 
        \caption{UCR (over $k$)}
        \label{Fig:CoincidencePlots-UCR-K-ami}
    \end{subfigure}%
    ~
    \begin{subfigure}[t]{0.49\textwidth}
        \centering
        \includegraphics[width=\textwidth]{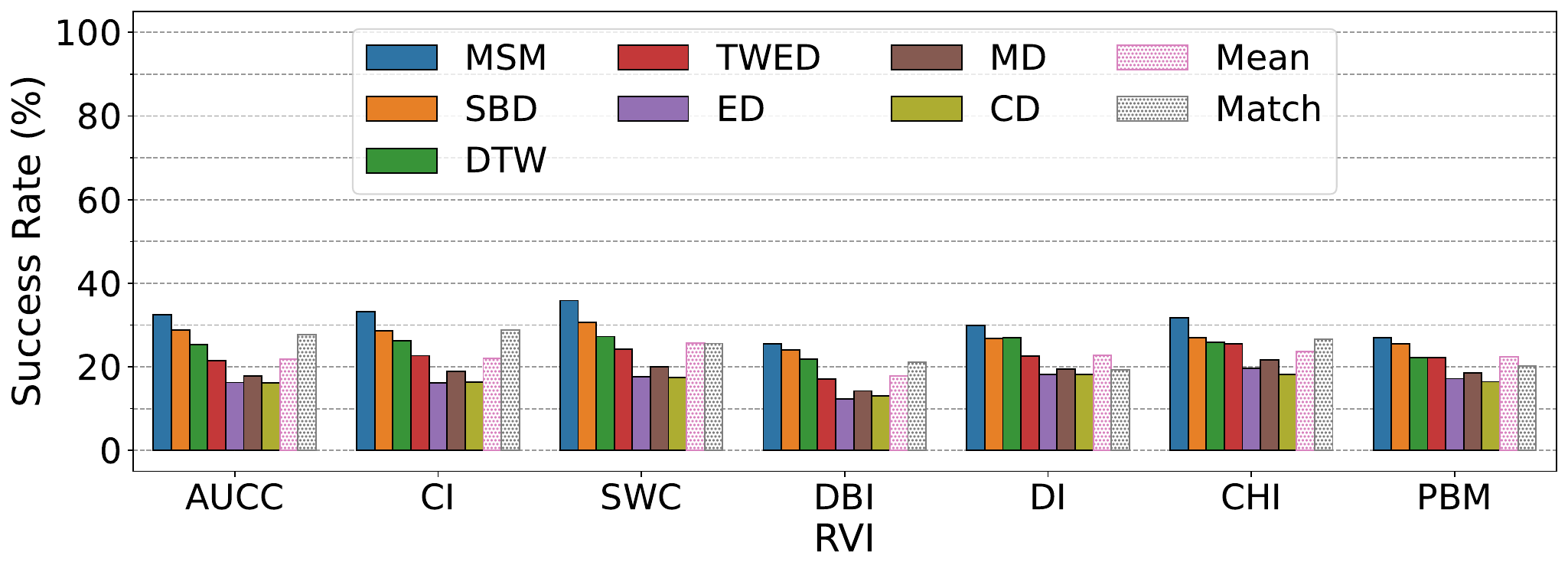}
        \caption{UCR (over SPs)}
        \label{Fig:CoincidencePlots-UCR-SP-ami}
    \end{subfigure}
    \begin{subfigure}[t]{0.49\textwidth}
        \centering
        \includegraphics[width=\textwidth]{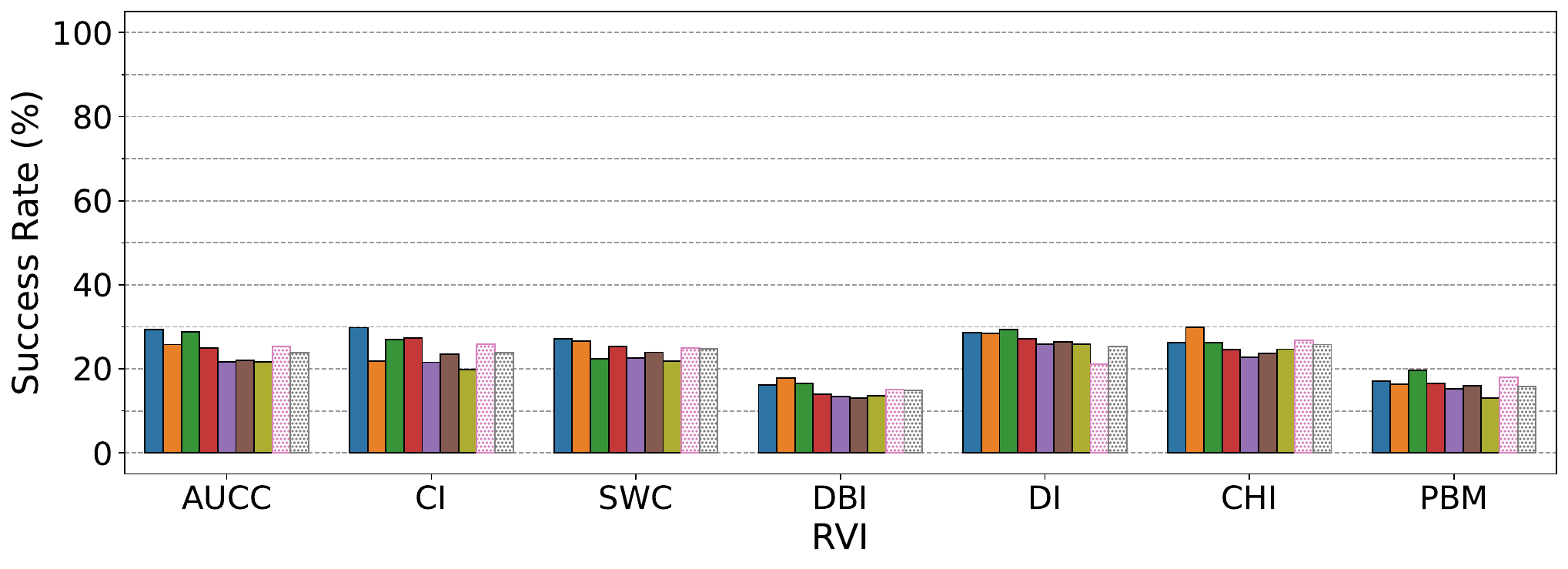} 
        \caption{UCR (over $k$ with max dataset ARI threshold of 0.6)}
        \label{Fig:CoincidencePlots-UCR-K-Threshold-ami}
    \end{subfigure}%
    ~
    \begin{subfigure}[t]{0.49\textwidth}
        \centering
        \includegraphics[width=\textwidth]{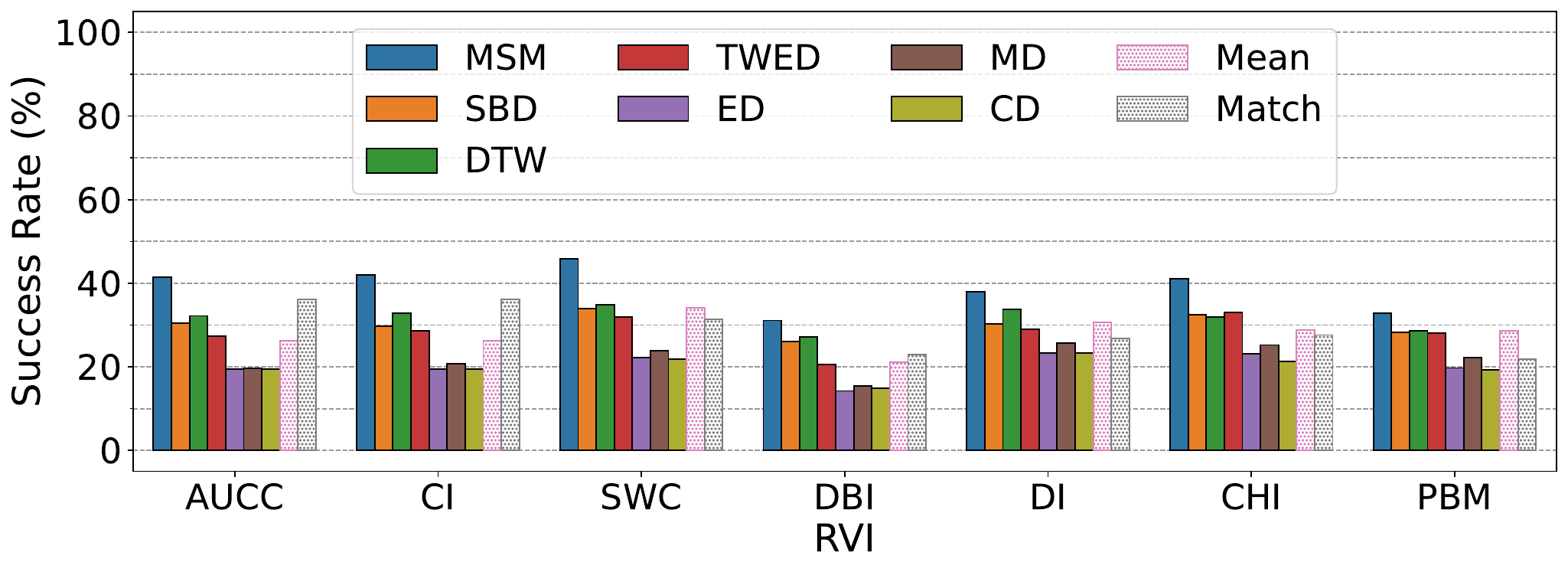}
        \caption{UCR (over SPs with max dataset ARI threshold of 0.6)}
        \label{Fig:CoincidencePlots-UCR-SP-Threshold-ami}
    \end{subfigure}
    \caption{The success rates for coincidence of optimal values for AMI and the different versions of each RVI. The first three rows show the results for each battery, and the left and right columns show the success rates for the $k$- and SP-selection tasks respectively. The final row has been produced for a subset of 41 datasets from the UCR archive where at least one of the dataset partitions had an ARI in excess of 0.6. Note that the legend on the right plot serves both plots in each row.}
    \label{Fig:CoincidencePlots-ami}
\end{figure}

\begin{figure}[!ht]
    \centering
    \includegraphics[width=0.95\textwidth]{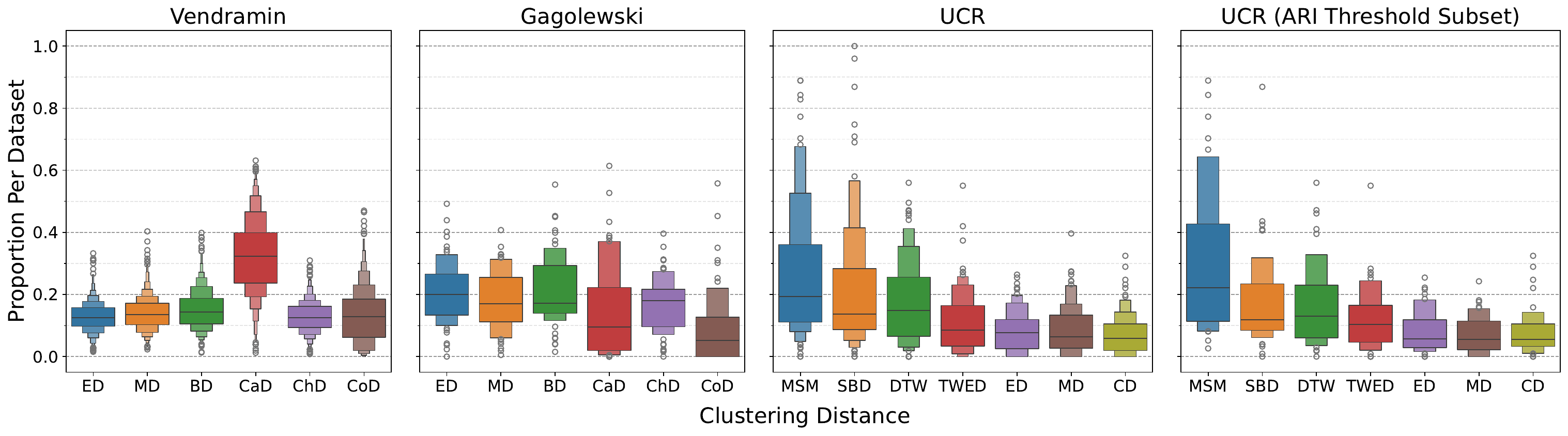}
    \caption{The proportion of times each SP generated the optimal partition according to the AMI for various combinations of clustering algorithm and $k$ for each of the datasets, i.e. which SP were found to be optimal according to the AMI when producing \Cref{Fig:CoincidencePlots-Vend-SP-ami,Fig:CoincidencePlots-Gag-SP-ami,Fig:CoincidencePlots-UCR-SP-ami,Fig:CoincidencePlots-UCR-SP-Threshold-ami}.}
    \label{Fig:Max-AMI-Distances}
\end{figure}

\newpage

\subsection*{Minimum EVI Requirements}

\begin{figure}[!h]
    \centering
    \begin{subfigure}{0.25\textwidth}
        \centering
        \includegraphics[height=4.2cm]{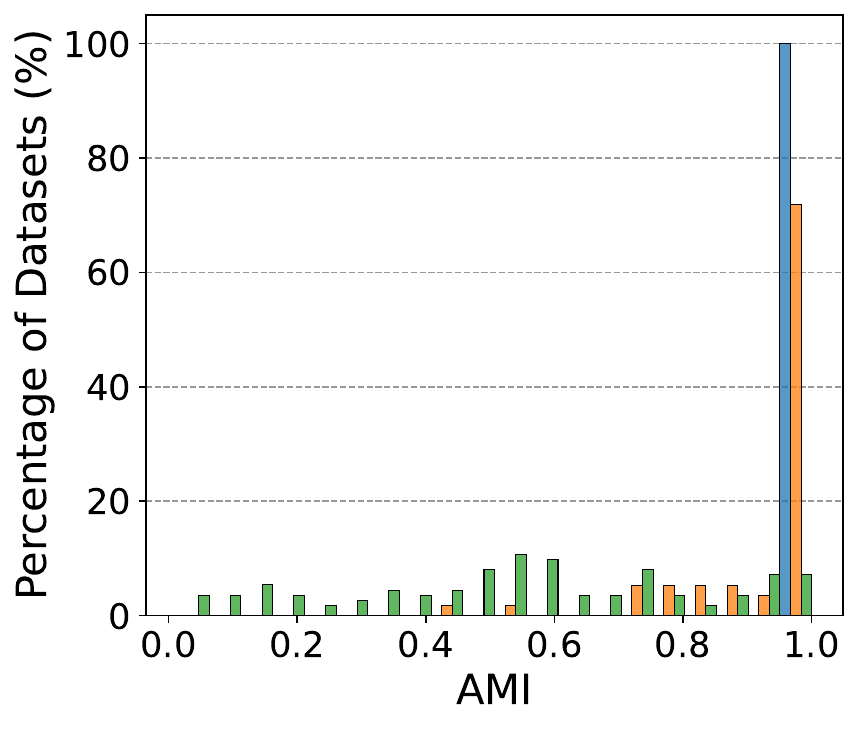}
        \caption{Maximum AMI}
        \label{Fig:AMI-Histograms-A}
    \end{subfigure}%
    \begin{subfigure}{0.75\textwidth}
        \centering
        \includegraphics[height=4.2cm]{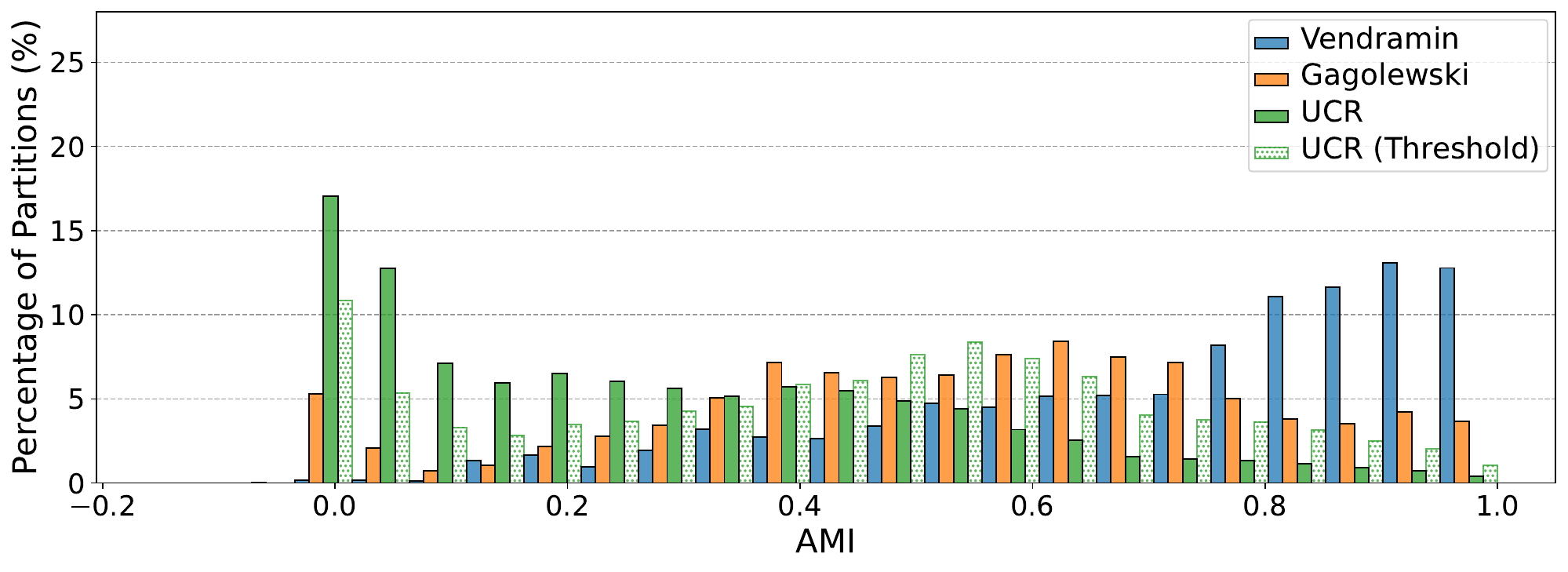}
        \caption{All AMI}
        \label{Fig:AMI-Histograms-B}
    \end{subfigure}
    \caption{Histograms of (a) the maximum AMI value per dataset, and (b) AMI values for every partition.}
    \label{Fig:AMI-Histograms}
\end{figure}

\subsection*{Correlation Analyses}
\begin{table}[!htb]
    \footnotesize
    \begin{adjustbox}{center}
    \begin{tabular}{ccccccccc} \toprule
         \multicolumn{1}{l}{} & \multicolumn{8}{c}{\textbf{Evaluation Schemes}} \\ \cmidrule(l){2-9}
          \textbf{RVI} & \textbf{ED} & \textbf{MD} & \textbf{BD} & \textbf{CaD} & \textbf{ChD} & \textbf{CoD} & \textbf{Mean} & \textbf{Match}\\ \midrule
        \multicolumn{9}{c}{\textcolor{gray}{\textbf{Selecting Optimal Partition With Varying $k$}}} \\ \cmidrule(l){1-9}
        AUCC & \mc{0.807}  & \mc{0.810}  & \mc{0.784}  & \mc{0.775}  & \mc{0.792}  & \mc{0.769}  & \mc{0.803}  & \mc{0.804}  \\
        CI & \mc{0.792}  & \mc{0.793}  & \mc{0.776}  & \mc{0.786}  & \mc{0.765}  & \mc{0.731}  & \mc{0.785}  & \mc{0.791}  \\
        SWC & \mc{0.669}  & \mc{0.668}  & \mc{0.680}  & \mc{0.706}  & \mc{0.675}  & \mc{0.688}  & \mc{0.689}  & \mc{0.655}  \\
        DBI & \mc{0.447}  & \mc{0.428}  & \mc{0.439}  & \mc{0.479}  & \mc{0.471}  & \mc{0.361}  & \mc{0.430}  & \mc{0.446}  \\
        DI & \mc{0.153}  & \mc{0.165}  & \mc{0.160}  & \mc{0.053}  & \mc{0.137}  & \mc{0.177}  & \mc{0.167}  & \mc{0.139}  \\
        CHI & \mc{0.801}  & \mc{0.804}  & \mc{0.807}  & \mc{0.756}  & \mc{0.800}  & \mc{0.648}  & \mc{0.708}  & \mc{0.798}  \\
        PBM & \mc{0.805}  & \mc{0.806}  & \mc{0.809}  & \mc{0.686}  & \mc{0.817}  & \mc{0.614}  & \mc{0.807}  & \mc{0.804}  \\ \cmidrule(l){1-9}
        \multicolumn{9}{c}{\textcolor{gray}{\textbf{Selecting Optimal Partition With Varying SP}}} \\ \cmidrule(l){1-9}
        AUCC & \mc{0.692}  & \mc{0.726}  & \mc{0.747}  & \mc{0.683}  & \mc{0.693}  & \mc{0.616}  & \mc{0.844}  & \mc{0.666}  \\
        CI & \mc{0.636}  & \mc{0.683}  & \mc{0.717}  & \mc{0.646}  & \mc{0.573}  & \mc{0.428}  & \mc{0.809}  & \mc{0.324}  \\
        SWC & \mc{0.625}  & \mc{0.637}  & \mc{0.606}  & \mc{0.655}  & \mc{0.535}  & \mc{0.337}  & \mc{0.635}  & \mc{0.402}  \\
        DBI & \mc{0.539}  & \mc{0.603}  & \mc{0.605}  & \mc{0.544}  & \mc{0.269}  & \mc{0.243}  & \mc{0.419}  & \mc{0.361}  \\
        DI & \mc{0.217}  & \mc{0.291}  & \mc{0.403}  & \mc{0.441}  & \mc{0.088}  & \mc{0.173}  & \mc{0.434}  & \mc{0.088}  \\
        CHI & \mc{0.069}  & \mc{0.099}  & \mc{0.060}  & \mc{0.470}  & \mc{0.056}  & \mc{-0.122}  & \mc{-0.122}  & \mc{0.077}  \\
        PBM & \mc{0.076}  & \mc{0.131}  & \mc{0.118}  & \mc{0.457}  & \mc{0.044}  & \mc{-0.063}  & \mc{0.110}  & \mc{-0.082}  \\
         \bottomrule
    \end{tabular}
    \end{adjustbox}
    \caption{Median Pearson correlations between the AMI and the different versions of each RVI for the $k$-selection task (top) and SP-selection task (bottom) on the Vendramin battery.}
    \label{Tab:Vend-AMI-Pearson-All}
\end{table}
\begin{table}[!htb]
    \footnotesize
    \begin{adjustbox}{center}
    \begin{tabular}{ccccccccc} \toprule
         \multicolumn{1}{l}{} & \multicolumn{8}{c}{\textbf{Evaluation Schemes}} \\ \cmidrule(l){2-9}
          \textbf{RVI} & \textbf{ED} & \textbf{MD} & \textbf{BD} & \textbf{CaD} & \textbf{ChD} & \textbf{CoD} & \textbf{Mean} & \textbf{Match} \\ \midrule
        \multicolumn{9}{c}{\textcolor{gray}{\textbf{Selecting Optimal Partition With Varying $k$}}} \\ \cmidrule(l){1-9}
        AUCC & \mc{0.604}  & \mc{0.603}  & \mc{0.499}  & \mc{0.520}  & \mc{0.550}  & \mc{0.423}  & \mc{0.555}  & \mc{0.547}  \\
        CI & \mc{0.577}  & \mc{0.615}  & \mc{0.372}  & \mc{0.472}  & \mc{0.565}  & \mc{0.418}  & \mc{0.519}  & \mc{0.497}  \\
        SWC & \mc{0.572}  & \mc{0.571}  & \mc{0.383}  & \mc{0.607}  & \mc{0.549}  & \mc{0.085}  & \mc{0.447}  & \mc{0.570}  \\
        DBI & \mc{0.353}  & \mc{0.346}  & \mc{0.092}  & \mc{0.151}  & \mc{0.268}  & \mc{-0.067}  & \mc{0.050}  & \mc{0.227}  \\
        DI & \mc{0.289}  & \mc{0.298}  & \mc{0.287}  & \mc{-0.043}  & \mc{0.313}  & \mc{-0.233}  & \mc{0.297}  & \mc{0.277}  \\
        CHI & \mc{0.501}  & \mc{0.455}  & \mc{0.252}  & \mc{0.084}  & \mc{0.520}  & \mc{0.197}  & \mc{0.150}  & \mc{0.211}  \\
        PBM & \mc{0.471}  & \mc{0.482}  & \mc{0.160}  & \mc{-0.192}  & \mc{0.469}  & \mc{0.121}  & \mc{0.146}  & \mc{0.277}  \\ \cmidrule(l){1-9} 
        \multicolumn{9}{c}{\textcolor{gray}{\textbf{Selecting Optimal Partition With Varying SP}}} \\ \cmidrule(l){1-9}
        AUCC & \mc{0.746}  & \mc{0.753}  & \mc{0.391}  & \mc{0.244}  & \mc{0.650}  & \mc{-0.330}  & \mc{0.577}  & \mc{0.169}  \\
        CI & \mc{0.755}  & \mc{0.758}  & \mc{0.267}  & \mc{0.273}  & \mc{0.692}  & \mc{-0.285}  & \mc{0.613}  & \mc{0.093}  \\
        SWC & \mc{0.728}  & \mc{0.750}  & \mc{0.564}  & \mc{0.382}  & \mc{0.686}  & \mc{-0.341}  & \mc{0.511}  & \mc{-0.179}  \\
        DBI & \mc{0.690}  & \mc{0.683}  & \mc{0.546}  & \mc{0.412}  & \mc{0.644}  & \mc{-0.191}  & \mc{-0.134}  & \mc{-0.206}  \\
        DI & \mc{0.561}  & \mc{0.582}  & \mc{0.129}  & \mc{0.235}  & \mc{0.560}  & \mc{-0.532}  & \mc{0.515}  & \mc{0.409}  \\
        CHI & \mc{0.577}  & \mc{0.570}  & \mc{-0.009}  & \mc{0.533}  & \mc{0.551}  & \mc{-0.483}  & \mc{-0.439}  & \mc{-0.503}  \\
        PBM & \mc{0.496}  & \mc{0.484}  & \mc{0.076}  & \mc{0.533}  & \mc{0.477}  & \mc{-0.482}  & \mc{-0.189}  & \mc{-0.238}  \\
         \bottomrule
    \end{tabular}
    \end{adjustbox}
    \caption{Median Pearson correlations between the AMI and the different versions of each RVI for the $k$-selection task (top) and SP-selection task (bottom) on the Gagolewski battery.}
    \label{Tab:Gag-AMI-Pearson-All}
\end{table}

\begin{table}[!htb]
    \footnotesize
    \begin{adjustbox}{center}
    \begin{tabular}{cccccccccc} \toprule
         \multicolumn{1}{l}{} & \multicolumn{9}{c}{\textbf{Evaluation Schemes}} \\ \cmidrule(l){2-10}
          \textbf{RVI} & \textbf{MSM} & \textbf{SBD} & \textbf{DTW} & \textbf{TWED} & \textbf{ED} & \textbf{MD} & \textbf{CD} & \textbf{Mean} & \textbf{Match} \\ \midrule
        \multicolumn{10}{c}{\textcolor{gray}{\textbf{Selecting Optimal Partition With Varying $k$}}} \\ \cmidrule(l){1-10}
        AUCC & \mc{0.560}  & \mc{0.465}  & \mc{0.538}  & \mc{0.560}  & \mc{0.505}  & \mc{0.521}  & \mc{0.505}  & \mc{0.555}  & \mc{0.485}  \\
        CI & \mc{0.555}  & \mc{0.520}  & \mc{0.675}  & \mc{0.383}  & \mc{0.345}  & \mc{0.450}  & \mc{0.430}  & \mc{0.595}  & \mc{0.532}  \\
        SWC & \mc{-0.250}  & \mc{-0.328}  & \mc{-0.358}  & \mc{-0.234}  & \mc{-0.354}  & \mc{-0.321}  & \mc{-0.385}  & \mc{-0.353}  & \mc{-0.127}  \\
        DBI & \mc{-0.210}  & \mc{-0.182}  & \mc{-0.257}  & \mc{-0.205}  & \mc{-0.231}  & \mc{-0.239}  & \mc{-0.243}  & \mc{-0.262}  & \mc{-0.166}  \\
        DI & \mc{-0.138}  & \mc{-0.254}  & \mc{-0.135}  & \mc{-0.223}  & \mc{-0.269}  & \mc{-0.169}  & \mc{-0.268}  & \mc{-0.274}  & \mc{-0.110}  \\
        CHI & \mc{-0.293}  & \mc{-0.044}  & \mc{-0.076}  & \mc{-0.296}  & \mc{-0.251}  & \mc{-0.341}  & \mc{-0.158}  & \mc{-0.206}  & \mc{-0.298}  \\
        PBM & \mc{-0.634}  & \mc{-0.447}  & \mc{-0.210}  & \mc{-0.653}  & \mc{-0.688}  & \mc{-0.656}  & \mc{-0.433}  & \mc{-0.554}  & \mc{-0.532}  \\ \cmidrule(l){1-10}
        \multicolumn{10}{c}{\textcolor{gray}{\textbf{Selecting Optimal Partition With Varying SP}}} \\ \cmidrule(l){1-10}
        AUCC & \mc{0.464}  & \mc{0.355}  & \mc{0.169}  & \mc{0.143}  & \mc{-0.230}  & \mc{0.012}  & \mc{-0.230}  & \mc{0.176}  & \mc{0.222}  \\
        CI & \mc{0.514}  & \mc{0.366}  & \mc{0.258}  & \mc{0.191}  & \mc{-0.162}  & \mc{0.017}  & \mc{-0.188}  & \mc{0.168}  & \mc{0.249}  \\
        SWC & \mc{0.527}  & \mc{0.348}  & \mc{0.293}  & \mc{0.212}  & \mc{-0.241}  & \mc{0.037}  & \mc{-0.250}  & \mc{0.189}  & \mc{0.212}  \\
        DBI & \mc{0.219}  & \mc{0.154}  & \mc{0.096}  & \mc{-0.024}  & \mc{-0.290}  & \mc{-0.208}  & \mc{-0.215}  & \mc{-0.000}  & \mc{0.017}  \\
        DI & \mc{0.215}  & \mc{0.154}  & \mc{0.090}  & \mc{-0.006}  & \mc{-0.246}  & \mc{-0.090}  & \mc{-0.227}  & \mc{-0.021}  & \mc{-0.019}  \\
        CHI & \mc{0.360}  & \mc{0.222}  & \mc{0.202}  & \mc{0.205}  & \mc{-0.119}  & \mc{0.101}  & \mc{-0.175}  & \mc{0.135}  & \mc{0.177}  \\
        PBM & \mc{0.239}  & \mc{0.174}  & \mc{0.097}  & \mc{0.123}  & \mc{-0.117}  & \mc{-0.024}  & \mc{-0.159}  & \mc{0.061}  & \mc{0.114}  \\
         \bottomrule
    \end{tabular}
    \end{adjustbox}
    \caption{Median Pearson correlations between the AMI and the different versions of each RVI for the $k$-selection task (top) and SP-selection task (bottom) on the UCR battery.}
    \label{Tab:UCR-AMI-Pearson-All}
\end{table}

\newpage

\begin{figure}[p]
    \centering
    \begin{subfigure}[t]{\textwidth}
        \centering
        \includegraphics[width=0.90\textwidth]{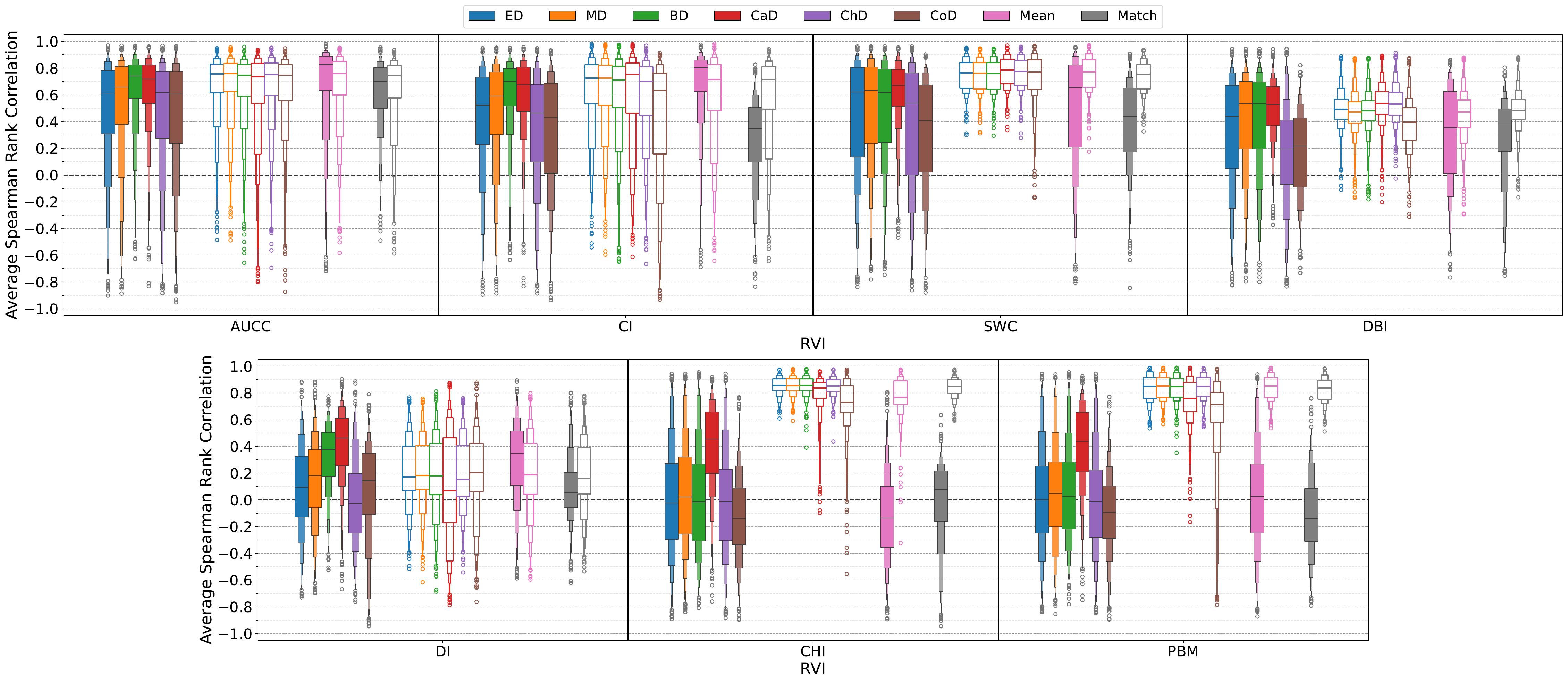}
        \caption{Vendramin}
        \label{Fig:CorrPlots-Vendramin}
    \end{subfigure}
    ~
    \begin{subfigure}[t]{\textwidth}
        \centering
        \includegraphics[width=0.90\textwidth]{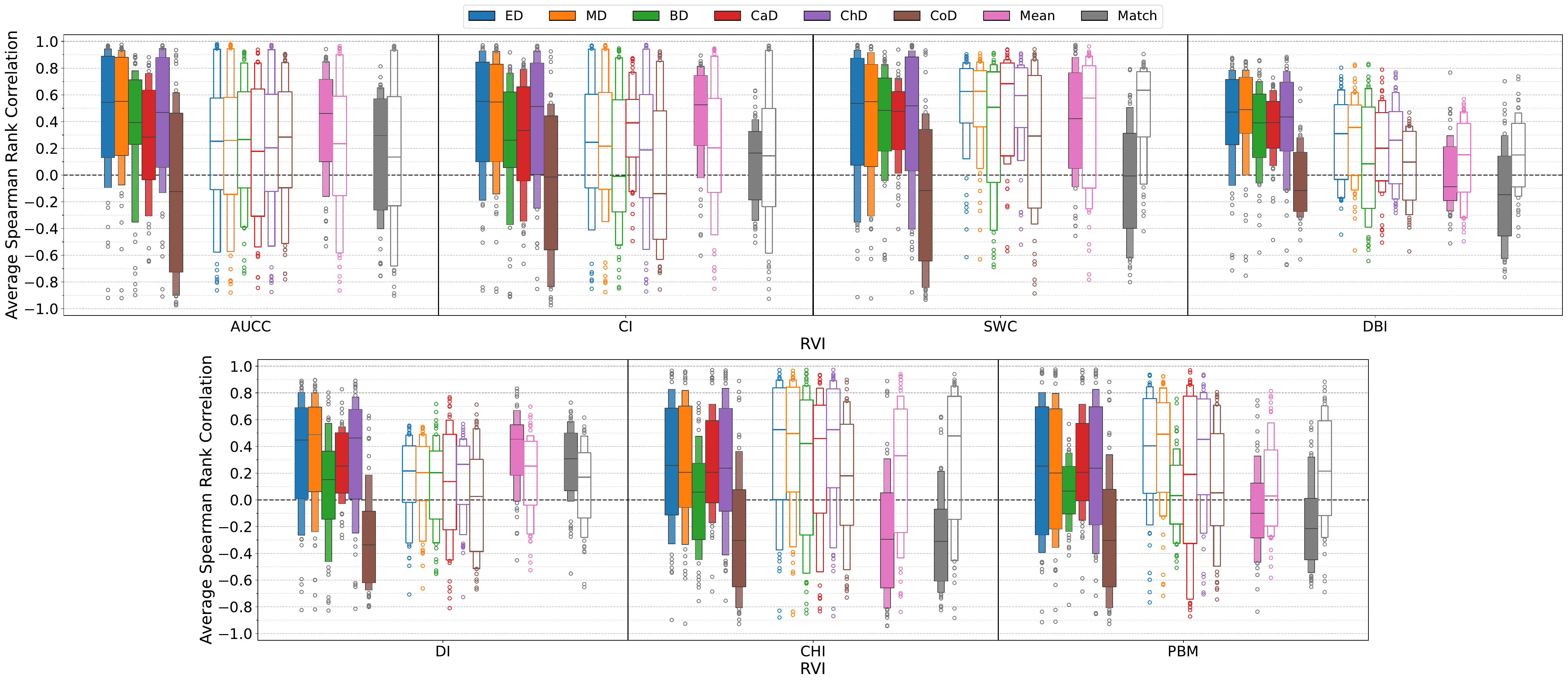}
        \caption{Gagolewski}
        \label{Fig:CorrPlots-Gagolewski}
    \end{subfigure}
    ~
    \begin{subfigure}[t]{\textwidth}
        \centering
        \includegraphics[width=0.90\textwidth]{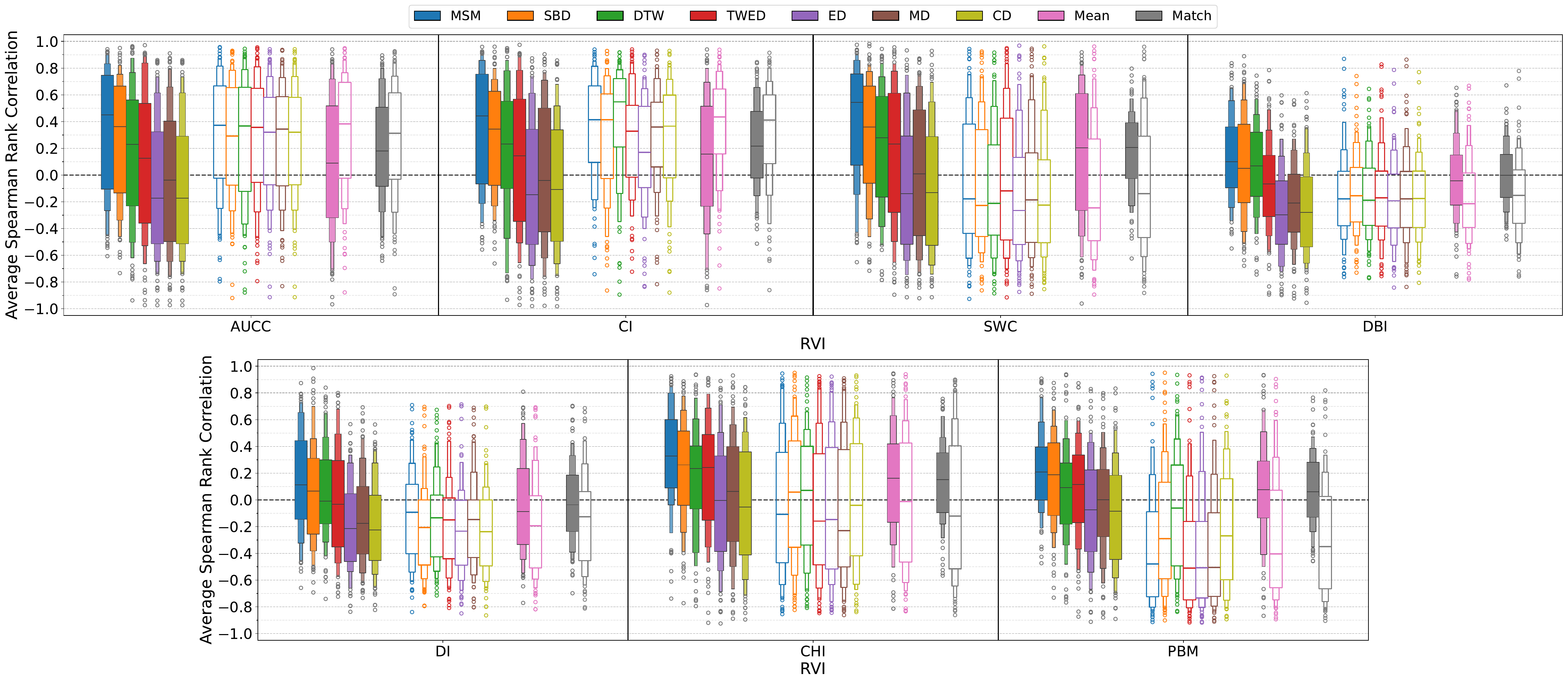}
        \caption{UCR}
        \label{Fig:CorrPlots-UCR}
    \end{subfigure}
    \caption{Enhanced boxplots of the distributions of median Pearson correlations between the ARI and various versions of RVIs, grouped by battery. The boxenplots are presented with no fill for the $k$-selection task, and filled for the SP-selection task.}
    \label{Fig:CorrPlots}
\end{figure}

\begin{figure}[p]
    \centering
    \begin{subfigure}[t]{\textwidth}
        \centering
        \includegraphics[width=0.90\textwidth]{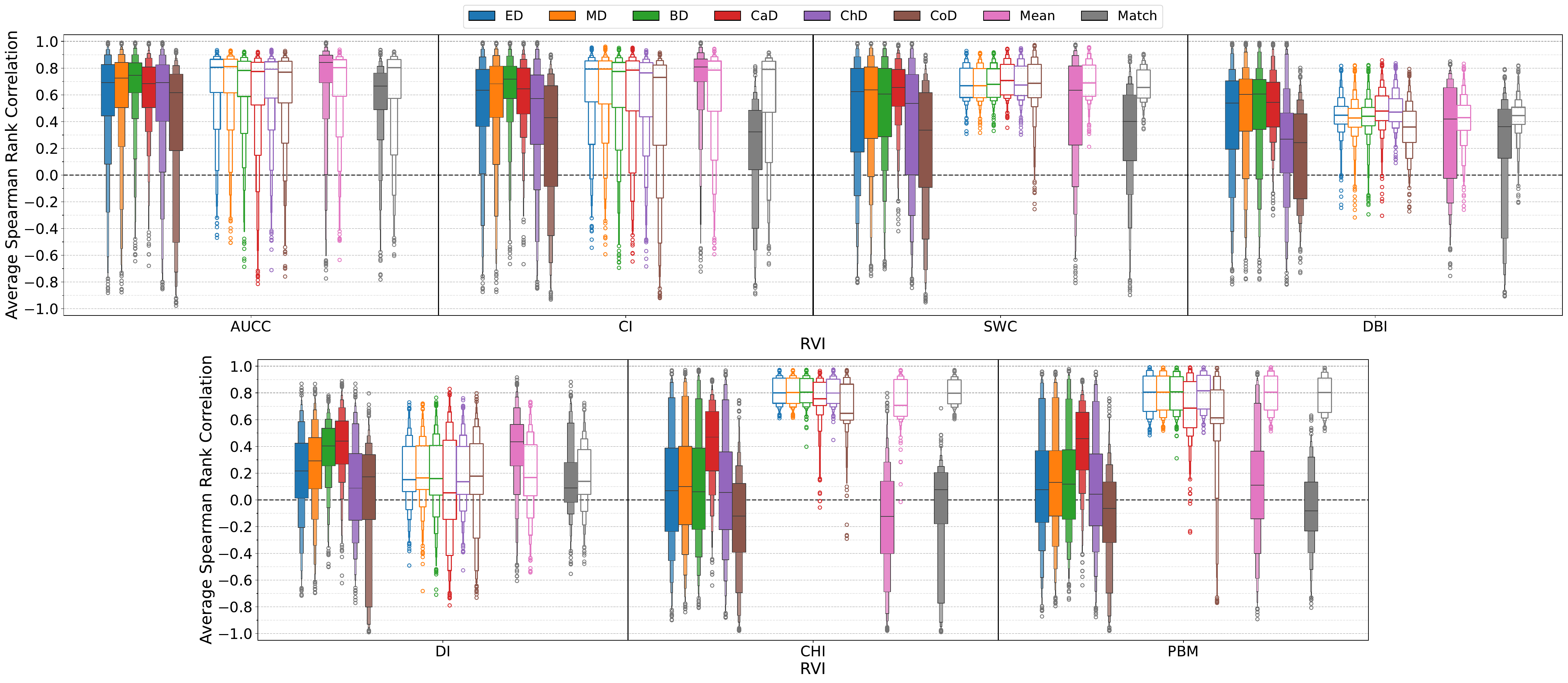}
        \caption{Vendramin}
        \label{Fig:CorrPlots-Vendramin-ami}
    \end{subfigure}
    ~
    \begin{subfigure}[t]{\textwidth}
        \centering
        \includegraphics[width=0.90\textwidth]{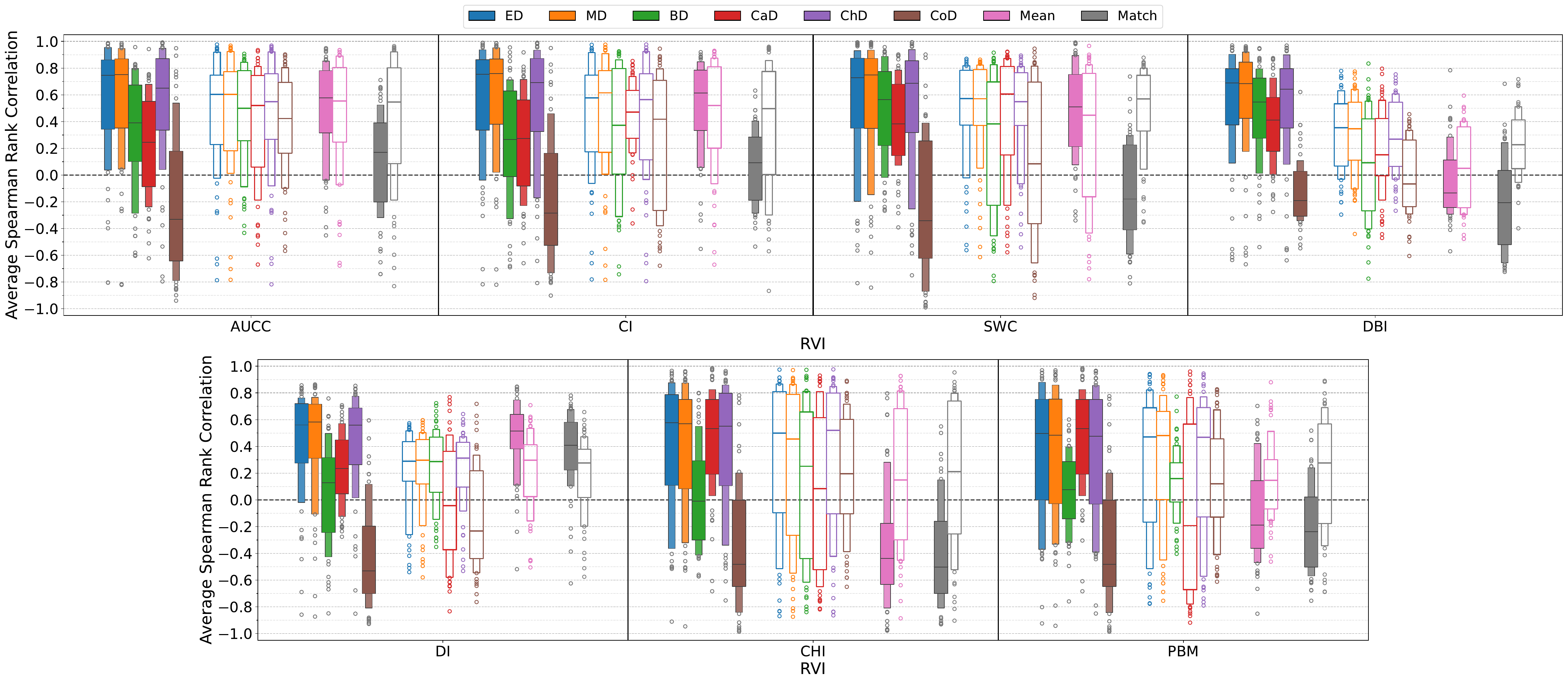}
        \caption{Gagolewski}
        \label{Fig:CorrPlots-Gagolewski-ami}
    \end{subfigure}
    ~
    \begin{subfigure}[t]{\textwidth}
        \centering
        \includegraphics[width=0.90\textwidth]{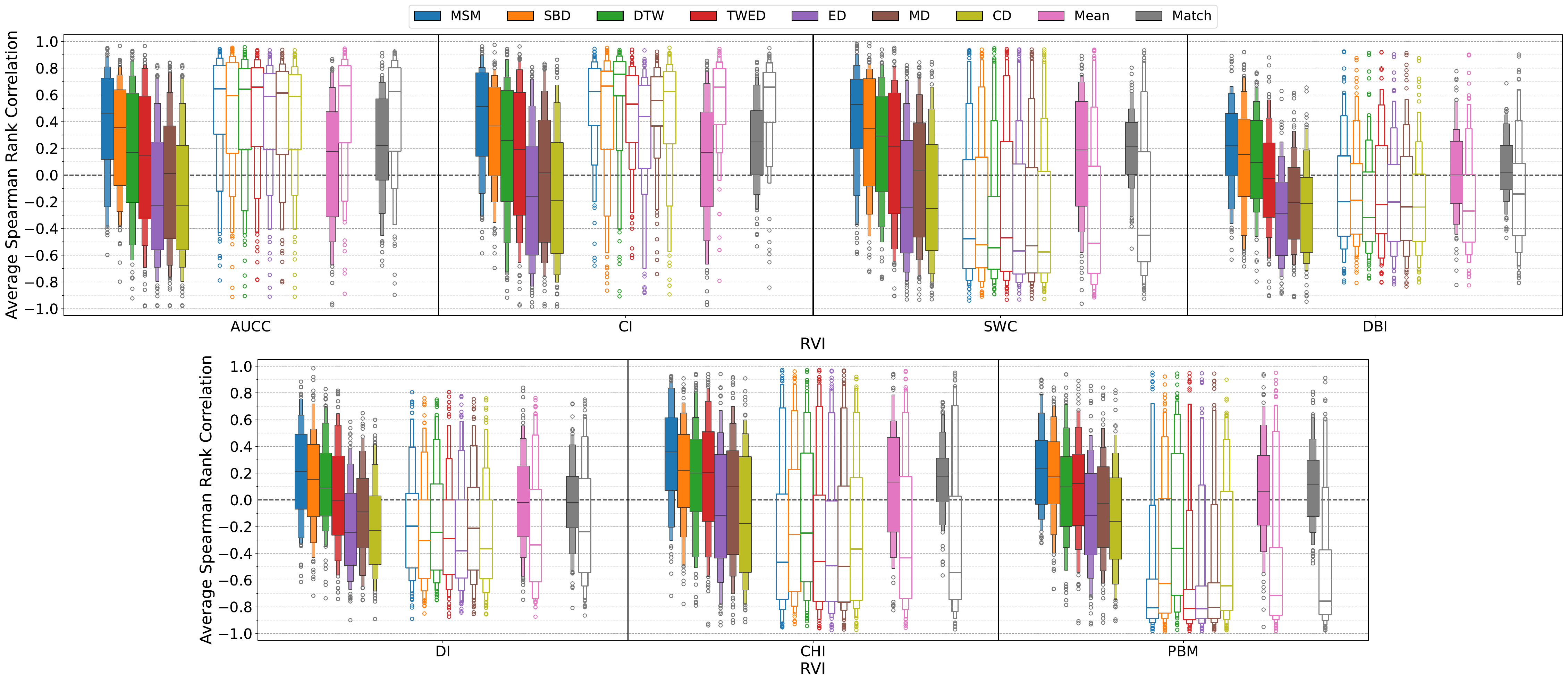}
        \caption{UCR}
        \label{Fig:CorrPlots-UCR-ami}
    \end{subfigure}
    \caption{Enhanced boxplots of the distributions of median Pearson correlations between the AMI and various versions of RVIs, grouped by battery. The boxenplots are presented with no fill for the $k$-selection task, and filled for the SP-selection task.}
    \label{Fig:CorrPlots-ami}
\end{figure}



\end{document}